\documentclass[10pt,journal]{IEEEtran}

\usepackage{amsmath,amsfonts,amssymb,bm} 
\usepackage{mathrsfs,pifont}          
\usepackage{tikz,overpic}      
\usepackage{array,booktabs,multirow,makecell} 
\usepackage[table]{xcolor}              
\usepackage{arydshln}                   
\usepackage{ragged2e}        
\usepackage{algorithmic}    
\usepackage{enumerate}                  
\usepackage[compress]{cite}           
\usepackage[colorlinks,linkcolor=red,anchorcolor=blue,citecolor=green]{hyperref} 

\graphicspath{{./figs/}}

\definecolor{light-gray}{gray}{0.9}
\definecolor{regressive}{RGB}{95,146,41}
\definecolor{generative}{RGB}{192,0,0}
\newcommand{\etal}{\textit{et al.}}
\newcommand{\revision}[1]{{\color{black}#1}}
\newcommand{\resubmit}[1]{{#1}}
\newcommand{\eg}{\textit{e.g.}} 
\newcommand*\circled[1]{\tikz[baseline=(char.base)]{%
    \node[shape=circle,fill,inner sep=.5pt](char){%
    \textcolor{white}{\footnotesize #1}};}}

\begin{document}

\title{Deep Learning Empowered Super-Resolution: A Comprehensive Survey and Future Prospects}
\author{Le~Zhang,~\IEEEmembership{Member,~IEEE},~Ao~Li,~Qibin~Hou,~\IEEEmembership{Member,~IEEE},~Ce~Zhu,~\IEEEmembership{Fellow,~IEEE},\\and~Yonina~C.~Eldar,~\IEEEmembership{Fellow,~IEEE}

\thanks{This work was supported by National Key R\&D Program of China (No.2024YFE0100700). (\textit{Corresponding author: Ce Zhu.})}
\thanks{Le Zhang,~Ao Li, and Ce Zhu are with the School of Information and Communication Engineering, University of Electronic Science and Technology of China, Chengdu, Sichuan 610054, China (e-mail: lezhang@uestc.edu.cn; aoli@std.uestc.edu.cn; eczhu@uestc.edu.cn).}
\thanks{Qibin Hou is with Tianjin Key Laboratory of Visual Computing and Intelligent Perception, School of Computer Science, Nankai University, Tianjin 300071, China (e-mail: houqb@nankai.edu.cn).}
\thanks{Yonina C. Eldar is with the Faculty of Mathematics and Computer Science, Weizmann Institute of Science, Rehovot 7610001, Israel (e-mail: yonina.eldar@weizmann.ac.il).}
}

\markboth{Journal of \LaTeX\ Class Files,~Vol.~14, No.~8, August~2021}%
{Shell \MakeLowercase{\textit{et al.}}: A Sample Article Using IEEEtran.cls for IEEE Journals}


\maketitle

\begin{abstract}
Super-resolution (SR) has garnered significant attention within the computer vision community, driven by advances in deep learning (DL) techniques and the growing demand for high-quality visual applications. With the expansion of this field, numerous surveys have emerged. Most existing surveys focus on specific domains, lacking a comprehensive overview of this field. Here, we present an in-depth review of diverse SR methods, encompassing single image super-resolution (SISR), video super-resolution (VSR), stereo super-resolution (SSR), and light field super-resolution (LFSR). We extensively cover over 150 SISR methods, nearly 70 VSR approaches, and approximately 30 techniques for SSR and LFSR. We analyze methodologies, datasets, evaluation protocols, empirical results, and complexity. In addition, we conducted a taxonomy based on each backbone structure according to the diverse purposes. We also explore valuable yet under-studied open issues in the field. We believe that this work will serve as a valuable resource and offer guidance to researchers in this domain. To facilitate access to related work, we created a dedicated repository available at \url{https://github.com/AVC2-UESTC/Holistic-Super-Resolution-Review}.
\end{abstract}


\section{Introduction}
\IEEEPARstart{S}{uper-Resolution} (SR) represents a highly promising field within computer vision, aiming to reconstruct high-resolution (HR) images from their low-resolution (LR) counterparts. \revision{Its applications span across various domains, including surveillance video~\cite{liu2022video,mudunuri2015low}, medical image enhancement~\cite{qiu2023medical,greenspan2009super}, reconstruction of old images~\cite{liang2021swinir, Ledig2017}, efficient image transmission~\cite{Zhang2017}, and specialized areas such as super-resolution microscopy~\cite{ben2024self,dardikman2020learned,solomon2019sparcom} and ultrasound~\cite{van2021super}. Recent studies also highlight advancements in single-image SR and model-based AI methods for imaging~\cite{tirer2021deep,monga2021algorithm}.} However, the reconstruction process encounters difficulties because of the inherent ambiguity arising from multiple HR patches that can degrade to the same LR patch, thereby making the problem ill-posed.

To address this problem, several traditional methods have been proposed~\cite{yang2010image,chang2004super,timofte2013anchored}. Yang~\etal~\cite{yang2010image} introduced a sparse coding-based approach to obtain high-resolution feature representations and reconstruct arbitrary images. Chang~\etal~\cite{chang2004super} utilized neighbor embedding technique with a locally linear embedding technique to reconstruct HR images. Timofte~\etal~\cite{timofte2013anchored} proposed an anchored neighborhood regression method to improve SR quality. However, these traditional methods exhibit three primary drawbacks. First, they necessitate manual design of the mapping function from LR to HR space, which is prone to sub-optimal performance. Second, they entail high computational costs, rendering them impractical in many real-time applications. \revision{Finally, the quality of the reconstructed super-resolution (SR) images, as indicated by metrics such as lower PSNR~\cite{wang2004image} and higher LPIPS~\cite{zhang2018unreasonable}, often does not meet the high expectations required for exceptional clarity and detail.}

In contrast, deep learning (DL) methods have shown remarkable potential in SR by harnessing the expressive power of neural networks. \revision{DL-based approaches automatically learn a mapping function between LR and HR images through weight adjustments using gradient descent and back-propagation~\cite{rumelhart1986learning}. These methods effectively overcome the manual design requirement associated with traditional approaches, addressing the first drawback~\cite{dong2014learning,li2019feedback,zhang2021edge}. Deep learning techniques benefit from efficient structural designs~\cite{kong2021classsr,bhardwaj2022collapsible,wang2021exploring,zhan2021achieving} and often require only a single pass to generate super-resolved images, enabling faster processing. This efficiently mitigates the second major drawback of traditional approaches. Furthermore, DL-based methods excel in extracting complex patterns~\cite{liang2021swinir,zheng2022cross,zhang2018image,Xia2022} that bridge the gap between LR and HR space, leading to promising advancements in both quantitative and qualitative aspects, thereby tackling the third drawback of traditional techniques.}

\begin{figure*}
    \centering
    \includegraphics[width=\linewidth]{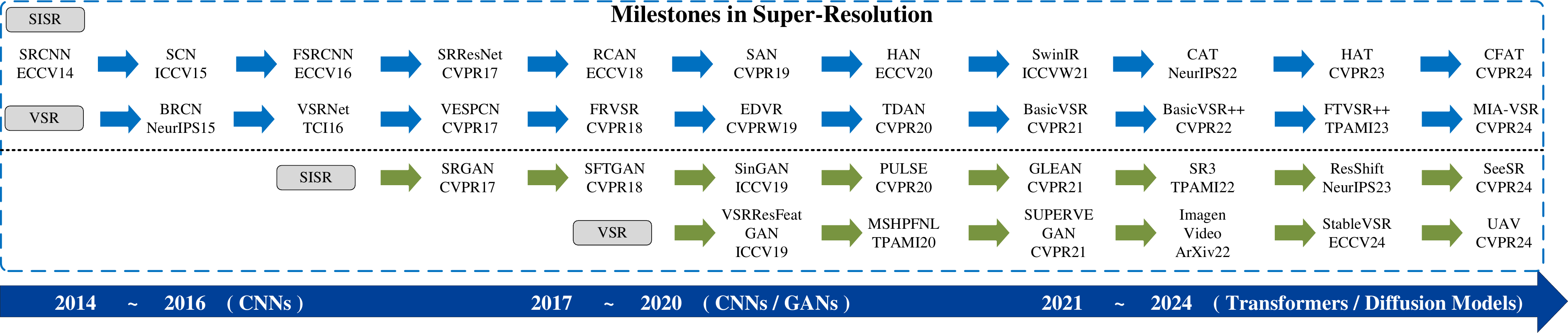}
    \caption{\revision{Deep-learning based super-resolution has witnessed remarkable advancements in recent years. In 2014, the integration of CNN-based techniques brought about a revolution in super-resolution, surpassing traditional methods. Subsequently, GAN-based methods were introduced to enhance perceptual quality in 2017. More recently, transformer-based models have emerged, demonstrating promising performance in super-resolution tasks. Super-resolution approaches can generally be categorized into two main branches: generative models (indicated in dark green arrows) and regression-based models (indicated in blue arrows). Generative models, such as GANs~\cite{Ledig2017,wang2018recovering,Shaham2019} and Diffusion models~\cite{saharia2022image,yue2023resshift,rota2023enhancing}, focus on generating perceptually realistic images, while regression models aim for pixel-wise accuracy. Within this framework, single image super-resolution initially dominated the field, with significant advancements in resolution and detail generation. Later, video super-resolution emerged, leveraging temporal information to further enhance video frame quality.}}      
    \label{fig:timeline}
\end{figure*}

 Motivated by the advantages mentioned above, substantial efforts have been devoted to DL-based SR approaches, as depicted in Fig.~\ref{fig:timeline}. Pioneered by SRCNN~\cite{dong2014learning}, subsequent research has explored deeper networks for single image super-resolution (SISR)~\cite{zhang2018image, tong2017image, umer2020deep, liang2021swinir}, introduced spatio-temporal sub-pixel convolution networks for video super-resolution (VSR)~\cite{Caballero2017}, employed parallax prior learning for stereo super-resolution (SSR)~\cite{jeon2018enhancing}, and utilized convolutional neural networks (CNNs) for light field super-resolution (LFSR)~\cite{yoon2017light}.
With the evolution of SR technology, numerous surveys have been published to provide valuable guidance to researchers in this domain. \resubmit{To better illustrate the landscape of existing SR surveys and highlight the specific contributions of our work, we present a comparative overview in Table \ref{tab:survey_results}. From this table, it is evident that while numerous surveys exist, they often focus on a single modality (\eg, Image SR as seen in Yang~\etal~\cite{Yang2019tmm} and Wang~\etal~\cite{wang_survey_tpami_20}, or Video SR as in Liu~\etal~\cite{Liu2022survey}), a specific technique (\eg, Diffusion Models for Image SR by Moser~\etal~\cite{moser2024diffusion}), or a particular application domain (\eg, Remote Sensing SR in Wang~\etal~\cite{WANG_survey_2022}). Furthermore, many cover literature up to earlier timeframes. Our survey distinguishes itself by offering a uniquely comprehensive and contemporary perspective. Specifically, we are the first to systematically review and unify four major super-resolution modalities—Image SR, Video SR, Stereo SR, and Light Field SR—under a consistent backbone-based taxonomy (\textit{i.e.}, CNN, Transformer, GAN, and Diffusion models). This approach not only provides a holistic understanding of the architectural evolution and methodological trends across these diverse SR tasks but also covers the most recent advancements up to early 2025, thereby addressing a significant gap in the existing literature for a truly integrative and up-to-date resource.}

\begin{figure*}
    \centering
    \includegraphics[width=0.95\linewidth]{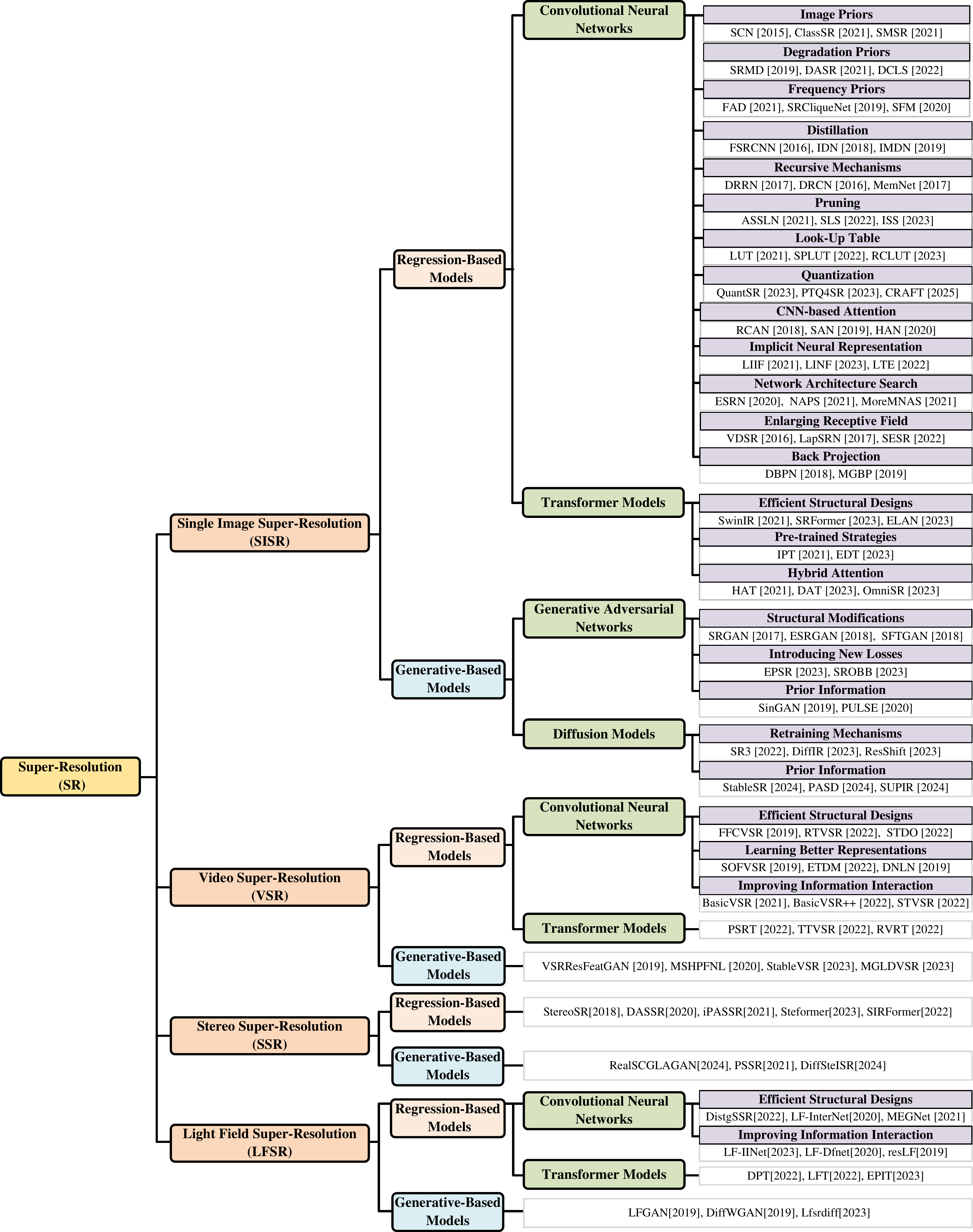}
    \caption{\resubmit{Taxonomy of super-resolution techniques in this survey. The figure presents a hierarchical categorization of super-resolution methods, initially divided into four major branches: single image super-resolution (SISR), video super-resolution (VSR), stereo super-resolution (SSR), and light field super-resolution (LFSR). Each branch is further divided into two primary pathways: regressive models (shown in yellow) and generative models (shown in blue). Within these pathways, methods are then categorized based on the network backbone, including convolutional neural networks (CNNs), transformer models, generative adversarial networks (GANs), and diffusion models. Additionally, each model class is further subdivided based on distinct design purposes such as efficient structural designs, hybrid attention mechanisms, and retraining techniques.}}
    \label{fig:sr_catalog}
\end{figure*}

\revision{Although existing surveys have significantly contributed to addressing various super-resolution challenges, a comprehensive review that spans all related domains is still lacking. To fill this gap, we present an in-depth review of various super-resolution (SR) problems. These include single image super-resolution (SISR), which exclusively utilizes intra-frame information to enhance the resolution of individual static images. In contrast, other SR methods such as video super-resolution (VSR), stereo super-resolution (SSR), and light field super-resolution (LFSR) employ various types of inter-frame information, including temporal and angular data, to improve image quality and resolution. As shown in Fig.~\ref{fig:sr_catalog}, we first categorize SR methods within these domains based on two primary approaches: generative models and regression models. Generative models are further divided into GAN-based~\cite{goodfellow2020generative} and diffusion-based methods~\cite{ho2020denoising,nikankin2022sinfusion}, while regression models are classified according to their network backbone, either CNN-based~\cite{lecun1998gradient} or transformer-based~\cite{vaswani2017attention, dosovitskiy2020image}. For each backbone type, we also perform a detailed classification based on their structural characteristics, providing a comprehensive overview of the landscape. 

CNNs excel at capturing local features and delivering strong performance, leveraging their inherent characteristics of translation invariance and locality. Transformers, on the other hand, have gained prominence in visual tasks by effectively capturing long-range spatial dependencies, enabling larger perceptual fields, and yielding improved performance. Commonly used loss functions in SR, such as $\mathcal{L}_1$ or $\mathcal{L}_2$, tend to encourage the network to minimize pixel-wise errors, which, in the context of an ill-posed problem, can lead to overly smooth results and suboptimal perceptual quality. To address this, GAN-based methods employ perceptual loss to enhance perceptual quality and provide a closer approximation to the human visual system (HVS)~\cite{Ledig2017,soh2019natural}}.
\begin{table*}[!t]
  \caption{\resubmit{Comparison of Existing Super-Resolution Survey Articles.}}
  \centering
  \resizebox{\linewidth}{!}{
  \setlength{\tabcolsep}{3pt}{
  \begin{tabular}{ccccc}
     \toprule[1pt]
    \textbf{Survey} & \textbf{Venue} & \textbf{Primary Scope} & \textbf{Timeframe} & \textbf{Description}\\
    \midrule
    Yue~\etal~\cite{YUE_survey_2016} (2016) & IEEE Signal Processing & Image SR & Up to 2015 & Reviews regularized frameworks for multi/single-frame SR; discusses priors and optimization. \\
    
    Yang~\etal~\cite{Yang2019tmm} (2019) & IEEE T-MM & Image SR & Up to 2018 & Details CNN architectures and optimization strategies for SISR. \\
    
    Wang~\etal~\cite{wang_survey_tpami_20} (2021) & IEEE T-PAMI & Image SR & Up to 2019 & Broadly surveys DL methods (CNN/GAN), learning paradigms, metrics, and applications for image SR.\\
    
    Anwar~\etal~\cite{Anwar2020} (2020) & ACM Computing Surveys & Image SR & Up to 2019 & Taxonomizes DL architectures for SISR; evaluates models and discusses real-world SR challenges.\\

    Li~\etal~\cite{li2020survey} (2020) & IET Image Processing & Image SR & Up to 2019 & Surveys classical and DL (CNN/GAN) SISR; discusses loss functions, architectures and metrics.\\

    Li~\etal~\cite{li2021review} (2021) & Irbm & Image SR (Medical Imaging) & Up to 2021 & Reviews DL for Medical Image SR; covers architectures, loss functions and modality applications.\\
    
    Liu~\etal~\cite{Liu2022survey} (2022) & Artificial Intelligence Review & Video SR & Up to 2021 & Systematically categorizes VSR methods by inter-frame information usage; discusses metrics and challenges.\\
    
    Chen~\etal~\cite{CHEN_survey_2022} (2022) & Information Fusion & Image SR (Real-World SISR) & Up to 2021 & Categorizes Real-World SISR methods; discusses datasets, metrics, and challenges.\\
    
    Wang~\etal~\cite{WANG_survey_2022} (2022) & Earth-Science Reviews & Image SR (Remote Sensing) & Up to 2022 & Reviews DL for Remote Sensing SR; categorizes methods (CNN/GAN/Attention) and discusses data.\\
    
    Bashir~\etal~\cite{bashir2021comprehensive} (2022) & PeerJ Computer Science & Image SR & Up to 2021 & Reviews DL for SISR by learning paradigms; details metrics, datasets and applications.\\
    
    Wang~\etal~\cite{wang2022review} (2022) & Remote sensing & Image SR (Remote Sensing) & Up to 2022 & Reviews DL for Remote Sensing SR; discusses architectures, learning strategies and challenges.\\

    Liu~\etal~\cite{liu2022super} (2022) &Annual Review of Biophysics & Super-Resolution Microscopy (SRM) & Up to 2022 & Explores SMLM and MINFLUX for cell biology; discusses principles, advancements and data analysis.\\
    
    Lepcha~\etal~\cite{LEPCHA_survey_2023} (2023) & Information Fusion & Image SR & Up to 2022 & Surveys traditional and DL (CNN/GAN/Transformer) SR; discusses metrics and applications.\\

    Prakash~\etal~\cite{prakash2023super} (2023) & Adv. Med. Imaging Detect. Diagn. & Super-Resolution Microscopy (SRM) & Up to 2022 & Overviews SRM techniques (STED/SIM/SMLM); discusses principles, applications and tools.\\
    
    Baniya~\etal~\cite{baniya2024survey} (2024) & IEEE T-ETCI & Video SR & Up to 2023 & Surveys VSR via key components and taxonomy; covers applications, challenges, and trends.\\
    
    Moser~\etal~\cite{moser2024diffusion} (2024) & IEEE T-NNLS & Image SR (Diffusion Models) & Up to 2024 & Reviews Diffusion Models for Image SR; covers theory, applications, and future challenges.\\
    
    Li~\etal~\cite{li2024systematic} (2024) & ACM Computing Surveys & Image SR & Up to 2023 & Systematically reviews DL for SISR by simulation, real-world, and domain-specific categories; notes challenges.\\
    
    \rowcolor{light-gray}Ours & N/A & Image/Video/Stereo/Light Field SR & Up to 2025 & Comprehensively reviews Image/Video/Stereo/LF SR; offers backbone-based taxonomy and discusses open issues.\\
    
     \bottomrule[1.5pt]
  \end{tabular}}
  }
  \label{tab:survey_results}
\end{table*}

The remaining sections of this paper are organized as follows: In Section \ref{sec:definition}, we revisit the definition of the SR task, explore evaluation metrics, elaborate on different domains, and discuss corresponding optimization objectives and network backbones. Subsequently, Section \ref{sec:sisr} provides a comprehensive review of SISR methods, while Section \ref{sec:vsr} focuses on VSR techniques. We then delve into SSR approaches in Section \ref{sec:ssr}, followed by an in-depth review of LFSR methods in Section \ref{sec:lfsr}. Section \ref{future} addresses existing challenges and outlines potential future directions in the SR field. Finally, in Section \ref{conclusion}, we present concluding remarks.

\section{Preliminaries}
\label{sec:definition}
\subsection{\textbf{Super-resolution Definition}}
\revision{In general, we define super-resolution (SR) in the context of 2D natural images. We assume that the HR image $X_{hr}$ and the LR image $X_{lr}$ obey the following degradation process:
\begin{equation}
X_{lr} = f_{de}(X_{hr}, \Theta) \downarrow_{_s} + \hspace{0.2cm}\epsilon,
\label{eq:lr_image}
\end{equation}
where $f_{de}(\cdot)$ represents the degradation process, $\Theta$ refers to the parameters of the degradation kernel, $\downarrow_{_s}$ indicates a down-sampling operation, and the variable $\epsilon$ denotes additive white Gaussian noise (AWGN) with a standard deviation of $\sigma$ (noise level). 

There are two main approaches to model this reconstruction process:

\textbf{Mapping-based SR Model:} In this regression approach, the objective is to learn a mapping function $\mathcal{F}$ by minimizing the distance between the predicted HR image $\mathcal{F}(X_{lr}, \theta)$ and the ground truth HR image $X_{hr}$. This can be formalized as: 
\begin{equation} 
\theta^* = \arg \min_{\theta} \sum_{i=1}^{N} E(\mathcal{F}(X_{lr}^i, \theta), X_{hr}^i), 
\label{eq} 
\end{equation} 
where $E(\cdot)$ denotes a distance function (\eg, mean squared error), and $N$ is the number of training samples $(X_{lr}^i, X_{hr}^i)$.

\textbf{Degradation Matching SR Model:}
This approach focuses on reconstructing the high-resolution (HR) image $\hat{X}_{hr}$ such that it best matches the given low-resolution (LR) image $X_{lr}$ after undergoing a specific degradation process. The relationship can be expressed as: 
\begin{equation} 
\hat{X}_{hr} = \arg\min_{X_{hr}} ||X_{lr} - f_{de}(X_{hr}, \Theta)||, 
\end{equation} 
where $f_{de}(\cdot)$ is the degradation function as defined in Eq.~\ref{eq:lr_image}. Recent advancements in this model involve leveraging generative models to explore the latent space for HR images that mimic the degraded versions of the LR input~\cite{Menon2020,bora2017compressed,gottschling2020troublesome}.

For each specific sample $(X_{lr}^i,X_{hr}^i)$, $i\in{1,\cdots,N}$, the details are defined according to the specific domains:
\begin{enumerate}
    \item In SISR, only intra-view information is used, where the LR image $X_{lr}^i\in\mathbb{R}^{H \times W\times C}$ is equivalent to the HR image $X_{hr}^i\in \mathbb{R}^{sH \times sW \times C}$ degraded by a magnification factor $s$. Here, $H$ and $W$ represent the height and width of the LR image, and $C$ is the number of color channels.
    \item For SSR, cross-view information can be utilized to provide more disparity correlation. The dimension is defined as: $X_{lr}^i\in \mathbb{R}^{M\times H \times W \times C}$ and $X_{hr}^i\in \mathbb{R}^{M\times sH \times sW \times C}$, where $M$ indicates different perceptive views.
    \item  VSR introduces temporal information, and the dimension is defined as: $X_{lr}^i\in \mathbb{R}^{T\times H \times W \times C}$ and $X_{hr}^i\in \mathbb{R}^{T\times sH \times sW \times C}$, where $T$ indicates the time serial index. 
    \item LFSR incorporates angular information, enabling the model to utilize multiple views from different angles for generating higher quality super-resolved images. This is exemplified by a light field camera that captures scenes from varied perspectives, encoding these as angular dimensions within the dataset. Specifically, the dimensions of an LFSR sample are defined as: \(X_{lr}^i \in \mathbb{R}^{H \times W \times C \times U \times V}\) and \(X_{hr}^i \in \mathbb{R}^{sH \times sW \times C \times rU \times rV}\), where \(U\) and \(V\) represent the angular dimensions and \(r\) signifies the angular magnification factor.

\end{enumerate}
}

In practical scenarios, simulating all degradation processes is challenging. Therefore, to simplify the SR task, the bicubic kernel and Gaussian blur kernel are commonly used to simulate the degradation process. Unless explicitly stated, the reviewed methods in this survey are designed under the condition of these two degradation approach.
        
\revision{\subsection{\textbf{Metrics}}
\textbf{Distortion Metrics:} In SR, two commonly used metrics for evaluating distortion performance are the peak signal-to-noise ratio (PSNR) and the structural similarity index measure (SSIM)~\cite{wang2004image}.

PSNR evaluates the quality of the reconstructed SR image $X_{sr}$ by comparing it to the ground truth HR image $X_{hr}$, and it is calculated as:
\begin{equation}
\text{PSNR} = 10 \cdot \log_{10} \left(\frac{\text{MAX}^2}{\text{MSE}}\right),
\label{eq:psnr}
\end{equation}
where MAX represents the maximum possible image value. Choosing the correct value for MAX is critical, especially in imaging fields like MRI~\cite{de2022deep,lyu2018super,andrew2021super}, where images may be stored in floating-point format. In such cases, using a fixed value like 255 may not be appropriate, and alternative normalization strategies should be considered. The mean square error (MSE) is defined as:
\begin{equation}
\text{MSE} = \frac{1}{CHW} \sum_{k=0}^{C-1} \sum_{i=0}^{H-1} \sum_{j=0}^{W-1} \left(X_{sr}(k,i,j) - X_{hr}(k,i,j)\right)^2,
\label{eq:mse}
\end{equation}
where $C$, $H$, and $W$ represent the channel, height, and width of the evaluated images, respectively.

SSIM offers a more perceptually motivated evaluation by modeling distortion through brightness, contrast, and structure changes. SSIM is computed on small windows, typically $8\times8$ or $11\times11$ pixels, of the image. The SSIM index between two image patches 
$X_{sr}$ and $X_{hr}$ is calculated as::
\begin{equation}
\text{SSIM}(X_{sr}, X_{hr}) = \frac{(2\mu_{X_{sr}}\mu_{X_{hr}}+c_1)(2\sigma_{X_{sr}X_{hr}}+c_2)}{(\mu_{X_{sr}}^2+\mu_{X_{hr}}^2+c_1)(\sigma_{X_{sr}}^2+\sigma_{X_{hr}}^2+c_2)},
\label{eq:ssim}
\end{equation}
\noindent where $\mu_{X_{sr}}$, $\sigma_{X_{sr}}^2$, $\mu_{X_{hr}}$, and $\sigma_{X_{hr}}^2$ represent the mean and variance of $X_{sr}$ and $X_{hr}$, respectively. The $\sigma_{X_{sr}X_{hr}}$ indicates the covariance between $X_{sr}$ and $X_{hr}$, and $c_1$ and $c_2$ are constant values used for numerical stability.

\textbf{Perceptual Metrics:} Beyond traditional distortion metrics, perceptual quality is often better captured by advanced methods such as the Frechet Inception Distance (FID)\cite{heusel2017gans} and the Learned Perceptual Image Patch Similarity (LPIPS)\cite{zhang2018unreasonable}. These metrics evaluate the similarity of high-level features rather than focusing purely on pixel differences. 

FID measures the similarity between the distributions of real and generated images in a feature space and is defined as:
\begin{equation}
\text{FID} = ||\mu_r - \mu_g||^2 + \text{Tr}(\Sigma_r + \Sigma_g - 2(\Sigma_r\Sigma_g)^{1/2}),
\end{equation}
where $(\mu_r, \Sigma_r)$ and $(\mu_g, \Sigma_g)$ are the means and covariances of the feature vectors from real and generated images, respectively. FID is particularly useful for assessing how well the generated images mimic the overall distribution of real images, making it valuable in tasks involving natural image generation. In practice, a lower FID score indicates better perceptual similarity between real and generated images.

LPIPS evaluates the similarity between deep feature representations of two images and is computed as:
\begin{equation}
\text{LPIPS}(x, y) = \sum_l \frac{1}{H_l W_l} \sum_{h,w} ||w_l\odot(\phi_l(x)_{hw} - \phi_l(y)_{hw})||_2^2,
\end{equation}
where $\phi_l$ denotes the deep features from layer $l$, and $H_l$ and $W_l$ are the height and width of the feature map at layer $l$. Note that using $w_l=1$ for all $l$ is equavalent to computing a cosine distance. In practice, LPIPS is widely adopted in tasks that prioritize perceptual similarity, as it leverages deep neural network activations to compare how similarly the networks perceive two images.

\textbf{Complexity:} In general, model complexity is indicated by two metrics: parameters and multi-add operations. The number of learnable parameters in a model reflects its size, while the number of multiplication and addition operations approximates the inference time on hardware platforms. However, since most works only report the number of parameters, we focus on parameters as the complexity metric in this survey.}
\subsection{\textbf{Domains Elaboration and Corresponding Optimization Objective}}
\subsubsection{\textbf{Single Image Super-Resolution}}
\
\newline
\indent Deep learning-based single image super-resolution (SISR) techniques leverage information from a single input image to generate a high-resolution output. \revision{These methods primarily focus on extracting intra-view features, and a larger receptive field often leads to higher-quality results~\cite{Gu2021,monga2021algorithm,tirer2021deep}.}

\textbf{Loss Functions:} Regarding the choice of loss function, pixel-wise losses are commonly used in SISR. The most widely adopted loss function is the $\mathcal{L}_1$ loss:
\begin{equation}
\mathcal{L}_1 = \frac{1}{CHW}\sum_{k=0}^{C-1}\sum_{i=0}^{H-1}\sum_{j=0}^{W-1}\lVert X_{sr}(k,i,j) - X_{hr}(k,i,j)\rVert_1,
\label{eq:l1}
\end{equation}
where $H$ and $W$ represent the height and width of the evaluated images, $C$ denotes the number of channels, and $X_{sr}$ and $X_{hr}$ represent the super-resolved and high-resolution images, respectively.

To address sensitivity to outliers, some approaches employ a smooth approximation of the $\mathcal{L}_1$ loss, known as the Charbonnier loss~\cite{charbonnier1994two}. It can be expressed as:
\begin{equation}
\mathcal{L}_{Char.} \!=\! \frac{1}{CHW}\!\!\sum_{k=0}^{C-1}\sum_{i=0}^{H-1}\sum_{j=0}^{W-1}\!\!\sqrt{(X_{sr}(k,i,j) \!\!-\!\! X_{hr}(k,i,j))^2 \!+\! \epsilon^2},
\label{eq:char}
\end{equation}
\revision{where $\epsilon$ is a small constant to ensure the function remains differentiable at zero. This constant helps to avoid the non-differentiability issue of the $\mathcal{L}_1$ loss at zero, providing a robust, smooth loss function that is less sensitive to outliers than the squared error loss.} Alternatively, some methods employ the $\mathcal{L}_2$ loss as their training objective:
\begin{equation}
\mathcal{L}_2 = \frac{1}{CHW}\sum_{k=0}^{C-1}\sum_{i=0}^{H-1}\sum_{j=0}^{W-1}(X_{sr}(k,i,j) - X_{hr}(k,i,j))^2.
\label{eq:l2}
\end{equation}
Since minimizing the above losses commonly leads to smooth results, GAN-based models introduced perceptual loss to obtain better quality super-resolved images:
\begin{equation}
\mathcal{L}_{Perc.} = \sum_{i}\lVert\varphi_{_i}(X_{sr}) - \varphi_{_i}(X_{hr})\rVert_2^2,
\label{eq:perc}
\end{equation}
where $\varphi(\cdot)$ indicates the high-level feature extractor, commonly adopting the VGG network~\cite{simonyan2014very}. Consequently, $\varphi_{_i}(X_{sr})$ and $\varphi_{_i}(X_{hr})$ represent the feature maps of the SR and HR images, which are extracted from the $i$-th layer of $\varphi(\cdot)$. Additionally, adversarial loss is introduced to train the GAN-based model, defined as:
\begin{equation}
\mathcal{L}_{Adv.} = -logD(G(X_{lr})),
\label{eq:adv.}
\end{equation}
where the discriminator is trained to minimize:
\begin{equation}
\mathcal{L}_{Dis.} = -log(D(X_{hr}))-log(1-D(G(X_{lr}))).
\label{eq:discriminator.}
\end{equation}
Here, $G(\cdot)$ represents the generator, and $D(G(\cdot))$ represents the probability of the discriminator over the training samples.

\subsubsection{\textbf{Video Super-Resolution}}
\
\newline
\indent Video Super-Resolution (VSR) extends the concept of SISR to video sequences, leveraging the temporal coherence between consecutive frames. By exploiting the temporal information, VSR methods produce sharper and more temporally consistent HR video frames. VSR techniques involve multiple stages, including motion estimation, frame alignment, and temporal fusion. Motion estimation estimates the motion between consecutive frames to compensate for inter-frame misalignment. Frame alignment techniques align the low-resolution frames to a common reference frame, enabling effective fusion and aggregation of information across frames. Temporal fusion methods combine the aligned frames to generate a high-resolution video sequence.

\textbf{Loss Functions:} In addition to the commonly used loss functions described in SISR, some works utilize optical flow to estimate the motion information and apply motion compensation loss, which can be formulated as follows:
\begin{equation}
\mathcal{L}_{mc} = \sum_{i=-T}^T\lVert X_{lr}^i - \Tilde{X}^{0\rightarrow{i}}_{lr}\rVert_1 + \alpha\lVert\Delta F_{i\rightarrow0}\rVert_1,
\label{eq:mc.}
\end{equation}
where $T$ is the size of the temporal span, $X_{lr}^i$ represents the $i$-th frame, and $\Tilde{X}_{lr}^{0\rightarrow i}$ indicates the backward warped $X_{lr}^0$ according to the estimated flow $F_{i\rightarrow 0}$. Additionally, $\Delta F_{i\rightarrow 0}$ is the total variation term. This motion compensation loss helps enhance the temporal coherence of the super-resolved video frames, leading to improved video quality and stability.

\subsubsection{\textbf{Stereo Super-Resolution}}
\
\newline
\indent Stereo Super-Resolution (SSR) aims to enhance the resolution and quality of depth or disparity maps obtained from multiple LR stereo image pairs. By exploiting the inherent correlation between stereo images, SSR generates HR disparity maps that accurately represent the scene geometry. The SSR process involves three main stages: disparity estimation, disparity refinement, and disparity fusion. Disparity estimation techniques estimate the disparity map from each LR stereo image pair, while disparity refinement methods enhance the disparity maps by exploiting inter-view and intra-view constraints. Disparity fusion techniques combine disparity maps from multiple viewpoints to generate a HR disparity map.

\textbf{Loss Functions:} In addition to the commonly used $\mathcal{L}1$ loss, other loss functions are employed in SSR. Considering the left-right consistency which can be formulated as:
\begin{equation}
\left\{
  \begin{array}{ll}
    X_{lr}^{left}\!\!\!&=\mathbf{M}_{right\rightarrow left}\otimes X_{lr}^{right} \\
    X_{lr}^{right}\!\!\!&=\mathbf{M}_{left\rightarrow right}\otimes X_{lr}^{left} \\
  \end{array}
\right.,
\label{eq:cons.}
\end{equation}
where $\mathbf{M}_{right\rightarrow left}$ and $\mathbf{M}_{left\rightarrow right}$ represent the two parallax-attention maps generated by PAM~\cite{wang2019learning} and $\otimes$ denotes batch-wise matrix multiplication. The photometric loss is defined as:
\begin{equation}
  \begin{aligned}
    \mathcal{L}_{pho.} &= \sum_{p\in\mathbf{V}_{left\rightarrow right}}\!\!\!\!\!\!\!\lVert X_{lr}^{left}(p)-(\mathbf{M}_{right\rightarrow left} \otimes X_{lr}^{right}(p))\rVert_1 \\
      &+ \sum_{p\in\mathbf{V}_{right\rightarrow left}}\!\!\!\!\!\!\!\lVert X_{lr}^{right}(p)-(\mathbf{M}_{left\rightarrow right} \otimes X_{lr}^{left}(p))\rVert_1
  \end{aligned},
\label{eq:pho.}
\end{equation}
where $p$ represents the pixel with a valid mask value, and $\mathbf{V}$ is defined as:
\begin{equation}
\mathbf{V}_{left\rightarrow right}(i,j) \!\!=\!\! \left\{
  \begin{array}{ll}
    1, & \!\!\!\text{if }\sum_{k\in[1,W]}\mathbf{M}_{left\rightarrow right}(i,j,k) \!>\!\tau \\
    0, & \!\!\!\text{otherwise} \\
  \end{array}
\right.,
\label{eq:valid.}
\end{equation}
where $\tau$ is a threshold and $W$ is the width of stereo images. $\mathbf{M}_{left\rightarrow right}(i,j,k)$ indicates the contribution of pixel $(i,j)$ in the left image to pixel $(i,k)$ in the right image.

The smoothness loss is also utilized often and can be formulated as:
\begin{equation}
  \begin{aligned}
    \mathcal{L}_{smo.} = \sum_{\mathbf{M}}\sum_{i,j,k}&(\lVert \mathbf{M}(i,j,k)-\mathbf{M}(i+1,j,k)\rVert_1 \quad + \\
      &\lVert \mathbf{M}(i,j,k)-\mathbf{M}(i,j+1,k)\rVert_1)
  \end{aligned},
\label{eq:smo.}
\end{equation}
where $\mathbf{M}\in{\mathbf{M}_{left\rightarrow right}, \mathbf{M}_{right\rightarrow left}}$. The cycle loss is also commonly used in SSR:
\begin{equation}
  \begin{aligned}
    \mathcal{L}_{cyc.} &= \sum_{p\in\mathbf{V}_{left\rightarrow right}}\lVert\mathbf{M}_{left\rightarrow right\rightarrow left}(p)-I(p)\rVert_1\\
      &+\sum_{p\in\mathbf{V}_{right\rightarrow left}}\lVert\mathbf{M}_{right\rightarrow left\rightarrow right}(p)-I(p)\rVert_1
  \end{aligned},
\label{eq:cyc.}
\end{equation}
where $I\in\mathbb{R}^{H\times W \times W}$ is a stack of $H$ identity matrices. $\mathbf{M}_{right\rightarrow left\rightarrow right}$ and $\mathbf{M}_{right\rightarrow left\rightarrow right}$ can be defined as:
\begin{equation}
\left\{
  \begin{array}{ll}
    \mathbf{M}_{right\rightarrow left\rightarrow right}&=\mathbf{M}_{right\rightarrow left}\otimes \mathbf{M}_{left\rightarrow right} \\
    \mathbf{M}_{right\rightarrow left\rightarrow right}&=\mathbf{M}_{left\rightarrow right}\otimes \mathbf{M}_{right\rightarrow left} \\
  \end{array}
\right..
\label{eq:cyc_atten.}
\end{equation}

\subsubsection{\textbf{Light Field Super-Resolution}}
\
\newline
\revision{\indent Light Field Super-Resolution (LFSR) focuses on enhancing the spatial resolution and quality of light field data, which captures both spatial and angular information of a scene. LFSR typically involves two main steps: sub-aperture image SR and angular upsampling. Sub-aperture image SR enhances the resolution of each sub-aperture image within the light field. Angular upsampling increases the angular resolution of the light field by generating new views between existing views. By harnessing this angular information, LFSR enhances computational models' capability to perceive depth and texture in a manner akin to human vision, making it particularly valuable in applications such as depth estimation~\cite{chen2023take,sheng2023lfnat}, semantic segmentation~\cite{sheng2022urbanlf,cong2023combining}, and saliency detection~\cite{sheng2016relative}.}

\textbf{Loss Functions:} In general, LFSR methods utilize the $\mathcal{L}_1$ loss as the main term of total loss. Considering the structural consistency, some works introduced the epipolar-plane image (EPI) gradient loss which can be formulated as
\begin{equation}
  \begin{aligned}
    \mathcal{L}_{e} &= \lVert\Delta_x\Hat{E}_{y,v}-\Delta_xE_{y,v}\rVert_1+\lVert\Delta_u\Hat{E}_{y,v}-\Delta_uE_{y,v}\rVert_1\\
      &+ \lVert\Delta_y\Hat{E}_{x,u}-\Delta_yE_{x,u}\rVert_1+\lVert\Delta_v\Hat{E}_{x,u}-\Delta_vE_{x,u}\rVert_1
  \end{aligned},
\label{eq:epi}
\end{equation}
where $\Hat{E}_{y,v}$ and $\Hat{E}_{x,u}$ indicate the EPIs of the super-resolved LF images, and $E_{y,v}$ and $E_{x,u}$ represent the EPIs of the groud-truth LF images. In additon, $(x,y)$ and $(u,v)$ denote the spatial plane and angular plane, and the gradients are calculated along both spatial and angular dimensions with both horizontal and vertical EPIs. 
\subsection{\textbf{Network backbone}}
\subsubsection{\textbf{Convolutional Neural Network}}
\
\newline
\indent \revision{Convolutional neural network (CNN)~\cite{he2017mask, girshick2014rich, liu2016ssd, redmon2016you,shin2017pixel, Fu2019tmm, liu2020crnet} is a deep learning model widely used for image analysis and computer vision tasks.} It consists of multiple layers, each performing specific operations to extract and learn features from input images. The operations in a CNN can be described as follows:

\textbf{Convolution Operation:} The convolution operation applies a filter to the input image using element-wise multiplication and summation. It can be represented as:
\begin{equation}
Conv(i,j)=\sum_{s=-a}^{a}\sum_{t=-b}^{b}w(s,t)x(i-s,j-t),
\label{eq:conv.}
\end{equation}
where $(i,j)$ is the coordinates of the input image, $w$ is the filter (also known as the kernel), and the size of the filter is $(2a+1)\times(2b+1)$.

\textbf{Activation Function:} After the convolution operation, an activation function is applied element-wise to introduce non-linearity. \revision{One commonly used activation function is the rectified linear unit (ReLU)~\cite{glorot2011deep}, defined as:
\begin{equation}
\text{ReLU}(x)=\max(0,x).
\label{eq:act.}
\end{equation}
This function sets negative values to zero and keeps positive values unchanged.}

\textbf{Pooling Operation:} Pooling reduces the spatial dimensions of the feature maps obtained from the convolutional layers. For example, max pooling selects the maximum value within a local region:
\begin{equation}
MaxPooling(x)=\max_{k\in S}x(k),
\label{eq:pool.}
\end{equation}
where $S$ indicates the region of calculation.

\textbf{Fully Connected Layers:} Fully connected layers are responsible for the final classification or regression tasks. Each neuron in these layers is connected to all neurons in the previous layer. The output of a fully connected layer is computed as:
\begin{equation}
FC(x)=\sigma(Wx+b),
\label{eq:fc}
\end{equation}
where $x$ is the input vector, $W$ is the weight matrix, $b$ is the bias vector, and $\sigma$ represents the activation function.

\subsubsection{\textbf{Transformer Network}}
\
\newline
\indent \revision{The transformer network~\cite{vaswani2017attention} is a highly effective architecture widely applied to various computer vision tasks~\cite{liu2021swin,carion2020end,zhu2020deformable,wang2021max,dosovitskiy2020image}, owing to its ability to capture long-range dependencies.} The main components are described as follows:

\textbf{Input Embedding:} The input image is divided into patches, which are linearly projected into higher-dimensional embeddings. Let $\mathbf{X}=[x_1,x_2,\dots,x_N]$ represent the input sequence of embeddings, where $x_i$ denotes the embedding for the $i$-th patch.

\revision{\textbf{Positional Encoding:} To incorporate positional information, a positional encoding matrix $\mathbf{P}\in \mathbb{R}^{N\times d}$ is added to the input embeddings~\cite{su2024roformer,vaswani2017attention,liu2021swin,liu2020learning,luo2022your}, where $N$ is the sequence length, and $d$ is the dimension of the embeddings.}

\textbf{Self-Attention:} For the embedding sequence $\mathbf{X}$, the self-attention mechanism computes the attention weights $\mathbf{W}$ using the following definition:
\begin{equation}
\mathbf{Q}=\mathbf{X}\mathbf{W}_Q, \quad \mathbf{K}=\mathbf{X}\mathbf{W}_K, \quad \mathbf{V}=\mathbf{X}\mathbf{W}_V,
\label{eq:qkv.}
\end{equation}
where $\mathbf{Q}\in\mathbb{R}^{N\times d_k}$, $\mathbf{K}\in\mathbb{R}^{N\times d_k}$, and $\mathbf{V}\in\mathbb{R}^{N\times d_k}$ are linear projections of $\mathbf{X}$ into query, key, and value vectors, respectively. Here, $\mathbf{W}_Q$, $\mathbf{W}_K$, and $\mathbf{W}_V$ are learnable weight parameters, and $d_k$ is the dimension of the key vectors. The self-attention operation can be formulated as:
\begin{equation}
\mathbf{W}=Softmax\left(\frac{\mathbf{Q}\mathbf{K}^T}{\sqrt{d_k}}\right)\mathbf{V}.
\label{eq:attn.}
\end{equation}

\textbf{Feed-Forward Networks:} After the self-attention step, a position-wise feed-forward network is applied to each embedding independently. It consists of two linear transformations followed by a non-linear activation function:
\begin{equation}
FFN=\text{ReLU}(xW_1+b_1)W_2+b_2,
\label{eq:ffn}
\end{equation}
where $x$ represents the input features, $W_1$ and $W_2$ are learnable weight parameters, and $b_1$ and $b_2$ are bias terms.

\subsubsection{\textbf{Generative Adversarial Network}}
\
\newline
\indent A Generative adversarial network (GAN)~\cite{goodfellow2020generative} is a powerful deep learning model that comprises two key components: a generator $G$ and a discriminator $D$. GANs are commonly used for generating synthetic data that closely resembles real data. The main modules of GAN can be described as follows:

\textbf{Generator:} The generator takes random noise as input and produces synthetic data samples. It can be represented as a function $G$ that maps a noise vector $z$ to a generated sample $x$:
\begin{equation}
x = G(z).
\label{eq:gen.}
\end{equation}
In some applications, instead of random noise $z$, a LR image $X_{lr}$ is used as input to generate the super-resolved image $X_{sr}$.

\textbf{Discriminator:} The discriminator is used to differentiate between real and generated data samples. It is represented by a function $D$ that outputs a probability indicating the likelihood that the input sample is real:
\begin{equation}
D(x) = P(\text{real}|x).
\label{eq:disc.}
\end{equation}
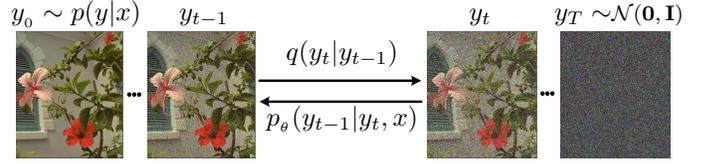
\begin{figure}[!t]
    \centering
    \begin{overpic}[scale=.41]{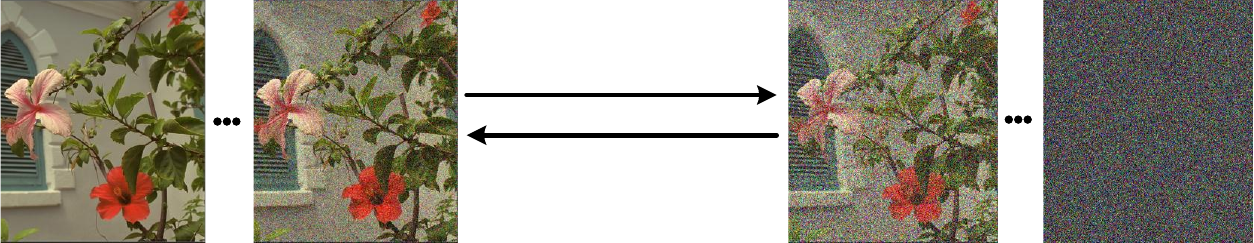}
        \put(-0.7, 21){$y_{_0}\sim p(y|x)$}
        \put(25, 21){$y_{t-1}$}
        \put(69.3, 21){$y_{t}$}
        \put(82.3, 21){$y_{T}\sim$\footnotesize$\mathcal{N}(\mathbf{0,I})$}
        \put(41.3, 15){$q(y_t|y_{t-1})$}
        \put(38.3, 5){$p_{_\theta}(y_{t-1}|y_{t},x)$}
    \end{overpic}
    \caption{\resubmit{In the forward diffusion process $q$ (left to right), Gaussian noise is incrementally added to the target image. Conversely, in the reverse inference process $p$ (right to left), the target image is iteratively denoised under the guidance of a source image $x$.}}     
    \label{fig:diffusion}
\end{figure}
\subsubsection{\textbf{Diffusion Models}}
\
\newline
\indent \resubmit{Diffusion models~\cite{yang2023diffusion,moser2024diffusion} refer to a class of generative models that simulate a diffusion process, where data is transformed into noise and then denoised to recover the original distribution. These models can be divided into three main categories: (1) Denoising Diffusion Probabilistic Models (DDPMs), (2) Score-based Generative Models, and (3) Variational Diffusion Models. In this section, we focus on Denoising Diffusion Probabilistic Models (DDPMs), a prominent approach within this class~\cite{ho2020denoising}. The forward diffusion process gradually transforms data into Gaussian noise, and the reverse process models the denoising to recover the original data distribution, as shown in Fig.~\ref{fig:diffusion}.}

\revision{\textbf{Forward Process:} This process gradually corrupts a data sample $\mathbf{x}_0$ into a sample $\mathbf{x}_T$ that is close to Gaussian noise, using a series of small Gaussian transitions:
\begin{equation}
q(\mathbf{x}_t \mid \mathbf{x}_{t-1}) = \mathcal{N}(\mathbf{x}_t; \sqrt{1-\beta_t} \mathbf{x}_{t-1}, \beta_t \mathbf{I}),
\end{equation}
where the hyper-parameters $\beta_{1:T}$, subject to $0<\beta_t<1$, which indicate the noise level added at each step, and $\mathbf{I}$ is the identity matrix. Each step of the diffusion process generates a noisy data $\mathbf{x}_t$, and the entire diffusion process forms a Markov chain
\begin{equation}
q(\mathbf{x}_{1:T} \mid \mathbf{x}_0) = \prod_{t=1}^T q(\mathbf{x}_t \mid \mathbf{x}_{t-1}).
\end{equation}
A key characteristic of the diffusion process is that we can directly sample $\mathbf{x}_t$ at any step $t$ based on the original data $\mathbf{x}_0$:
\begin{equation}
    \mathbf{x}_t \sim q(\mathbf{x}_t \mid \mathbf{x}_0).
\end{equation}
Here, we define $\alpha_t = 1 - \beta_t$ and $\overline{\alpha_t} = \prod_{i=1}^t \alpha_i$. By employing the reparameterization trick, we have
\begin{equation}
\mathbf{x}_t = \sqrt{\overline{\alpha}_t} \mathbf{x}_0 + \sqrt{1 - \overline{\alpha}_t} \boldsymbol{\epsilon}
\end{equation}

\textbf{Reverse Process:} The reverse process is defined by learning to reverse the forward diffusion process. Starting from $\mathbf{x}_T$ assumed to be drawn from a normal distribution, the process is modeled as:
\begin{equation}
p(\mathbf{x}_{t-1} \mid \mathbf{x}_t) = \mathcal{N}(\mathbf{x}_{t-1}; \mu_{\theta}(\mathbf{x}_t, t), \Sigma_{\theta}(\mathbf{x}_t, t)).
\end{equation}
The denoising network predicts the mean $\mu(\mathbf{x}_t, t)$ and covariance $\Sigma(\mathbf{x}_t, t)$ in this process. Importantly, the mean and covariance parameters are crucial for accurately simulating the reverse diffusion process and have significant impacts on the quality of the generated samples. In the super-resolution domain, the conditional formulation $ p_\theta (\mathbf{x}_{t-1} \mid \mathbf{x}_t, \mathbf{z}) $, conditioned on $ \mathbf{z} $ (\eg, a low-resolution image), uses $ \mu_\theta (\mathbf{x}_t, \mathbf{z}, t) $ and $ \Sigma_\theta (\mathbf{x}_t, \mathbf{z}, t) $ instead.

\textbf{Optimization Objective:} If we consider the intermediate variables as latent variables, then the diffusion model is actually a latent variable model containing $T$ latent variables. It may be considered a special case of Hierarchical VAEs~\cite{luo2022understanding}. We can then use variational inference to derive the variational lower bound (VLB, also known as ELBO) as the maximization objective, which leads to:

\begin{equation}
\begin{aligned}
\log p_{\theta}(\mathbf{x}_0) &= \log \int p_{\theta}(\mathbf{x}_0:\mathbf{x}_T) \, d\mathbf{x}_{1:T} \\
&= \log \int \frac{p_{\theta}(\mathbf{x}_0:\mathbf{x}_T) q(\mathbf{x}_{1:T} \mid \mathbf{x}_0)}{q(\mathbf{x}_{1:T} \mid \mathbf{x}_0)} \, d\mathbf{x}_{1:T} \\
&\geq \mathbb{E}_{q(\mathbf{x}_{1:T} \mid \mathbf{x}_0)} \left[ \log \frac{p_{\theta}(\mathbf{x}_0:\mathbf{x}_T)}{q(\mathbf{x}_{1:T} \mid \mathbf{x}_0)} \right].
\end{aligned}
\end{equation}
The last step employs Jensen's inequality. For network training, the training objective is to minimize the negative VLB:
\begin{equation}
\begin{aligned}
L &= \mathbb{E}_{q(\mathbf{x}_{1:T} \mid \mathbf{x}_0)} \left[ -\log \frac{p_\theta (\mathbf{x}_0:\mathbf{x}_T)}{q(\mathbf{x}_{1:T} \mid \mathbf{x}_0)} \right]\\ &= \mathbb{E}_{q(\mathbf{x}_{1:T} \mid \mathbf{x}_0)} \left[ \log \frac{q(\mathbf{x}_{1:T} \mid \mathbf{x}_0)}{p_\theta (\mathbf{x}_0:\mathbf{x}_T)} \right].
\end{aligned}
\end{equation}
}

\subsection{\textbf{Learning Strategies}}
Learning strategies in super-resolution aim to leverage various information sources to enhance image resolution. Supervised and unsupervised approaches form the two primary categories of learning strategies, each addressing the super-resolution task through different means.
\subsubsection{\textbf{Supervised Super-resolution}}
\
\newline
\indent In supervised super-resolution, the training process involves paired LR-HR image examples. These pairs serve as input-output pairs for training deep neural networks~\cite{liang2021swinir,li2023feature,wang2018esrgan}. The network learns to capture the relationship between LR and HR images, using HR images as the target. The learning process is guided by minimizing the distance between the super-resolved image and the HR image. The loss function typically encompasses pixel-wise differences, such as $\mathcal{L}_1$~\cite{liang2021swinir,li2023feature} or perceptual losses~\cite{wang2018esrgan} that consider feature representations. 
\subsubsection{\textbf{Unsupervised Super-resolution}}
\
\newline
\indent Unsupervised super-resolution~\cite{wei2021unsupervised, wang2021unsupervised, Wang2021, shocher2018zero, soh2020meta, yuan2018unsupervised} aims to enhance image resolution without the need for explicit HR supervision during training. DASR~\cite{wang2021unsupervised} estimates degradation information using a trainable encoder in the latent feature space, with the degradation encoder trained in an unsupervised manner through contrastive learning. ZSSR~\cite{shocher2018zero} employs a self-supervised approach, where the network utilizes the internal recurrence of information within a single image. The unsupervised framework typically incorporates perceptual losses, adversarial losses, or cycle-consistency losses~\cite{yuan2018unsupervised} to further improve the quality of the super-resolved images. 

\section{Single Image Super-resolution}
\label{sec:sisr}

\subsection{\textbf{Regressive Models}}
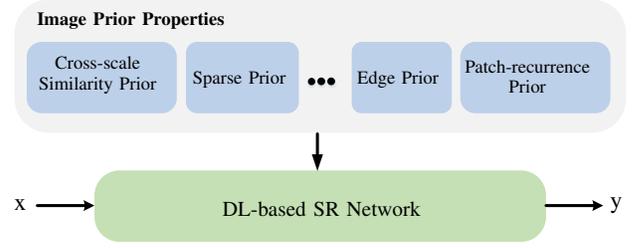
\begin{figure}
    \centering
    \begin{overpic}[scale=.38]{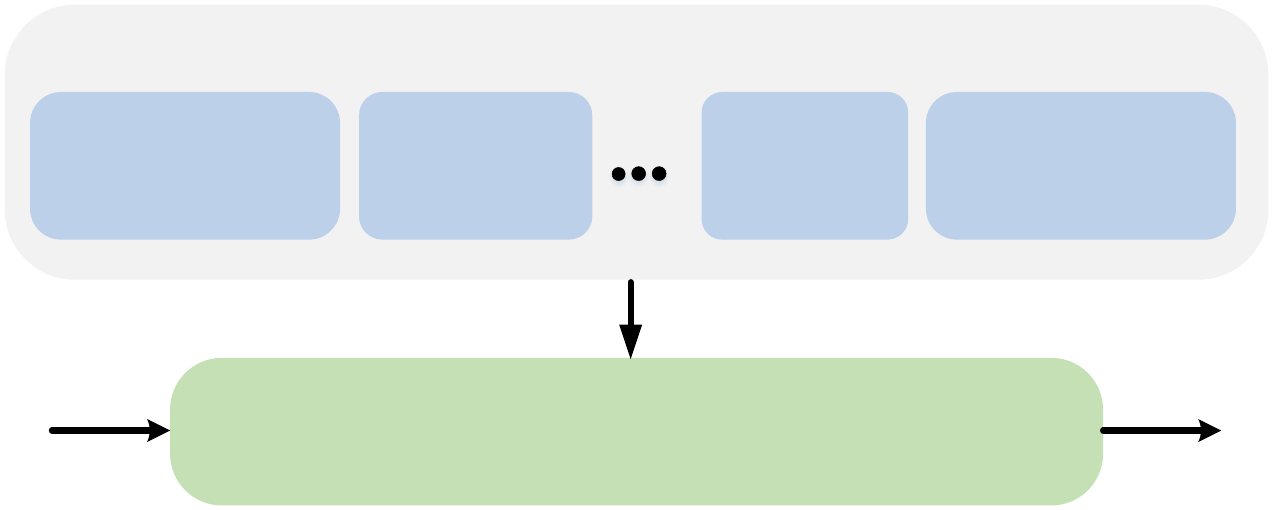}
        \put(33, 4.6){\footnotesize~DL-based SR Network}
        \put(3.3, 25){\scriptsize~\shortstack{Cross-scale \\Similarity Prior}}
        \put(28.3, 26){\scriptsize~Sparse Prior}
        \put(55, 26){\scriptsize~Edge Prior}
        \put(72.5, 24.5){\scriptsize~\shortstack{Patch-recurrence \\Prior}}
        \put(-1, 5.5){\small~x}
        \put(95.8, 5.8){\small~y}
        \put(3, 35.6){\scriptsize~\textbf{Image Prior Properties}}
    \end{overpic}
    \caption{Incorporating various image priors to inform the design of the SR network.}       
    \label{fig:priors}
\end{figure}
\subsubsection{\textbf{CNN-based SISR}} We categorize CNN-based methods into several more specific subcategories to organize these approaches in a structured way.

\indent\textbf{Image Priors:} Incorporating domain-specific prior knowledge is a common approach to improve the efficiency of SISR models~\cite{mei2020image,kong2021classsr, Nazeri2019,Cheng2020} as shown in Fig.~\ref{fig:priors}. Wang~\etal~\cite{wang2015deep} proposed a network that combined sparse coding with expert knowledge to enhance the training process. Mei~\etal~introduced a cross-scale non-local attention mechanism based on a recurrent network, considering cross-scale image correlation prior~\cite{mei2020image}. Kong~\etal~presented a unified framework that combined classification and SR, employing a classification module to classify sub-images into different classes based on restoration difficulties computed by PSNR, and then performed SR separately for each class~\cite{kong2021classsr}. Wang~\etal~explored sparsity in image SR and developed a sparse mask network that learns sparse masks to prune redundant computation~\cite{wang2021exploring}. Gu~\etal~identified the indispensability of edge processing in SR and proposed an efficient SR approach by dividing the SR process into edge and flat parts~\cite{Gu2022eccv}. Nazeri~\etal~integrated edge priors into their model and formulated a loss function that combines image content and structure terms~\cite{Nazeri2019}. Jiang~\etal~improved weight and data initialization and modified batch normalization (BN) for binary convolution networks by analyzing feature distributions~\cite{Jiang2021}. Shocher~\etal~exploited the internal recurrence of information within a single image and trained their model with both LR and counterpart downsampling versions of the input image~\cite{shocher2018zero}. Cheng~\etal~leveraged depth information to learn the internal degradation of the image, treating distant patches as LR and short-distance patches as HR, and constructing a self-supervised model~\cite{Cheng2020}. Park~\etal~trained their model via meta-learning and fine-tuned it at the test stage, allowing the model to adapt to the specific test image using its patch-recurrence property~\cite{Park2020}. Gu~\etal~performed attribution analysis of SR networks to understand which inputs have the greatest influence on SR performance, providing insights for designing SR networks and interpreting low-level vision deep models~\cite{Gu2021}. Tian~\etal~introduced a bit mapping strategy in~\cite{tian2023cabm} that resorts the edge information to build the look-up table of the bit selector, resulting in an efficient network.
\begin{figure}
    \centering
    \begin{overpic}[scale=.6]{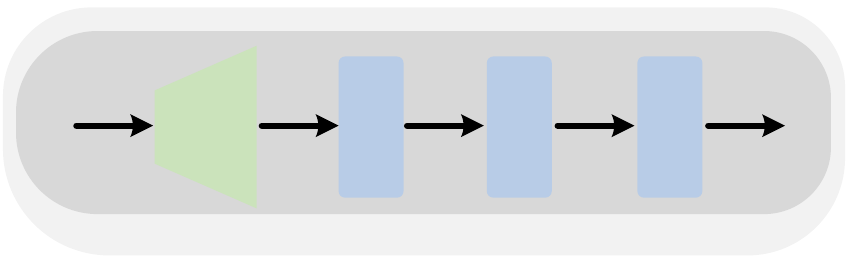}
        \put(22.8, 14.3){\scriptsize~\rotatebox{90}{UP}}
        \put(41.5, 13){\scriptsize~\rotatebox{90}{Conv}}
        \put(59, 13){\scriptsize~\rotatebox{90}{Conv}}
        \put(76.8, 13){\scriptsize~\rotatebox{90}{Conv}}
        \put(4.8, 15.5){\small~x}
        \put(92.2, 16){\small~y}
        \put(41.2, 3){\scriptsize~SRCNN~\cite{dong2014learning}}
    \end{overpic}
    \caption{The SRCNN~\cite{dong2014learning} framework with only three convolutional layers. This pioneering work was the first to successfully introduce deep learning into SISR.}       
    \label{fig:srcnn}
\end{figure}
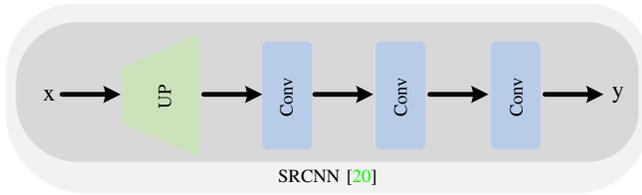

\textbf{Lightweight Blocks:} \revision{In efforts to optimize efficiency, one prevalent strategy is the adoption of lightweight blocks that boost performance without compromising computational economy.}

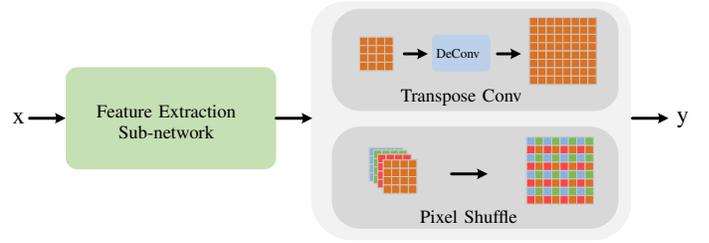
\begin{figure}[!t]
    \centering
    \begin{overpic}[scale=.37]{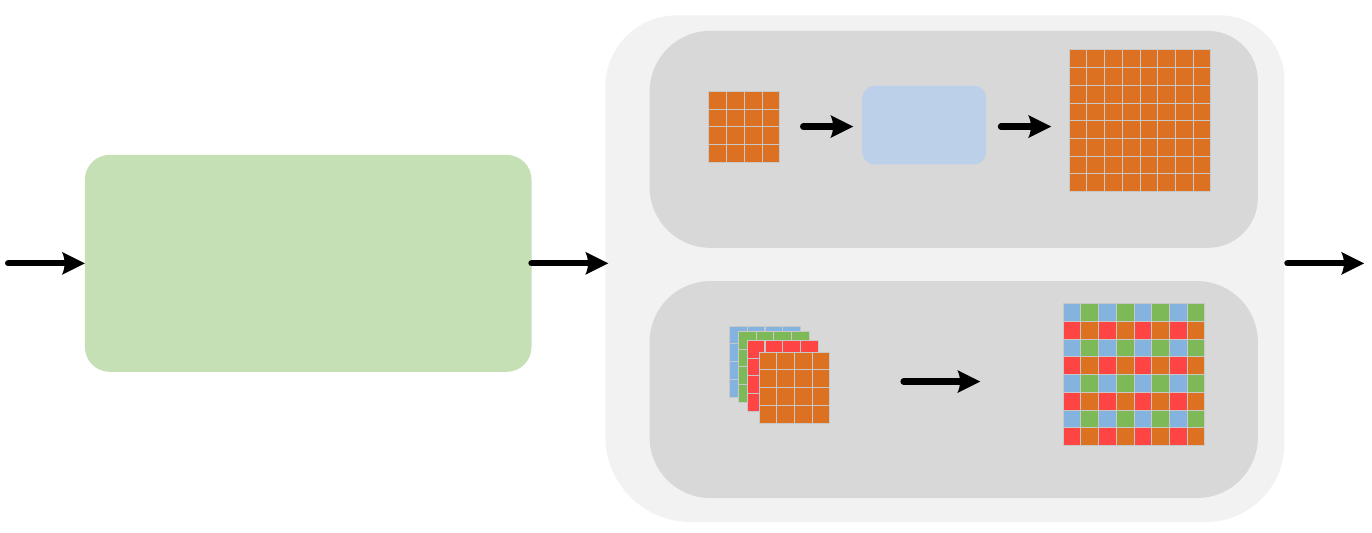}
        \put(10, 16.5){\scriptsize~\shortstack{Feature Extraction \\Sub-network}}
        \put(57, 22.3){\scriptsize~Transpose Conv}
        \put(-3.3, 18.8){\small~x}
        \put(99.4, 19){\small~y}
        \put(60, 3.5){\scriptsize~Pixel Shuffle}
        \put(62.8, 29){\tiny~DeConv}
    \end{overpic}
    \caption{Two efficient methods for reconstructing HR images.} 
    \label{fig:lightweight}
\end{figure}

A common early approach, as shown in Fig.~\ref{fig:srcnn}, is to upsample the LR image to the same size as the HR image and then pass it through SRCNN\cite{dong2014learning}, which consists of three convolutional layers, to generate the super-resolved image. Building on this, FSRCNN~\cite{fsrcnn2016} improved both model performance and computational efficiency by replacing the up-sampling step in the input stage with transposed convolution~\cite{dumoulin2016guide} at the final stage. Similarly, Shi~\etal~\cite{Shi2016} proposed efficient sub-pixel shuffling to enhance the reconstruction of HR images, as illustrated in Fig.~\ref{fig:lightweight}.

Another line of work focuses on splitting the SR process into stages to further optimize performance. Fan~\etal~\cite{Fan2017} introduced a lightweight conv-block, separating the process into LR and HR stages to extract multi-scale features. Zhang~\etal~\cite{zhang2019nonlinear} followed with an efficient ensemble block, designed to mimic ensemble deep networks, which demonstrated improved results. In a similar vein, Luo~\etal~\cite{luo2020latticenet} developed the lattice block, which uses two butterfly structures to integrate residual blocks.

Recent advancements have also targeted reducing the computational complexity of common SR operations. Chao~\etal~\cite{chao2023equivalent} transformed several time-consuming operations, such as clipping, repeating, and concatenating, into more efficient convolutional operations, leading to a highly efficient SR model. Lin~\etal~\cite{lin2023memory} proposed a scalable SR model that is memory-efficient and can handle multiple scales by dynamically switching masks based on the lottery tickets hypothesis.

Additionally, the integration of distillation techniques has emerged as an effective way to build lightweight models with enhanced feature extraction. Hui~\etal~\cite{Hui2018} proposed a compact network composed primarily of information distillation blocks, combining enhancement and compression units to capture both long- and short-range features. This concept was further extended by Hui~\etal~\cite{Hui2019}, who introduced selective fusion with multi-distillation blocks to optimize the model. Meanwhile, Lee~\etal~\cite{Lee2020} leveraged HR information as privileged data within a knowledge distillation network, significantly boosting model efficiency.

RLFN~\cite{kong2022residual} advanced efficient SISR by utilizing a residual local feature learning method with three convolutional layers, achieving a balance between performance and inference time. This approach also incorporated a contrastive loss and a multi-stage warm-start training strategy, further enhancing runtime efficiency. Finally, Gao~\etal~\cite{Gao2022} proposed diverse feature distillation blocks to improve performance while maintaining low computational complexity. They also introduced a self-calibration feature fuse block for adaptive information combination.

\begin{figure}[!t]
    \centering
    \begin{overpic}[scale=.4]{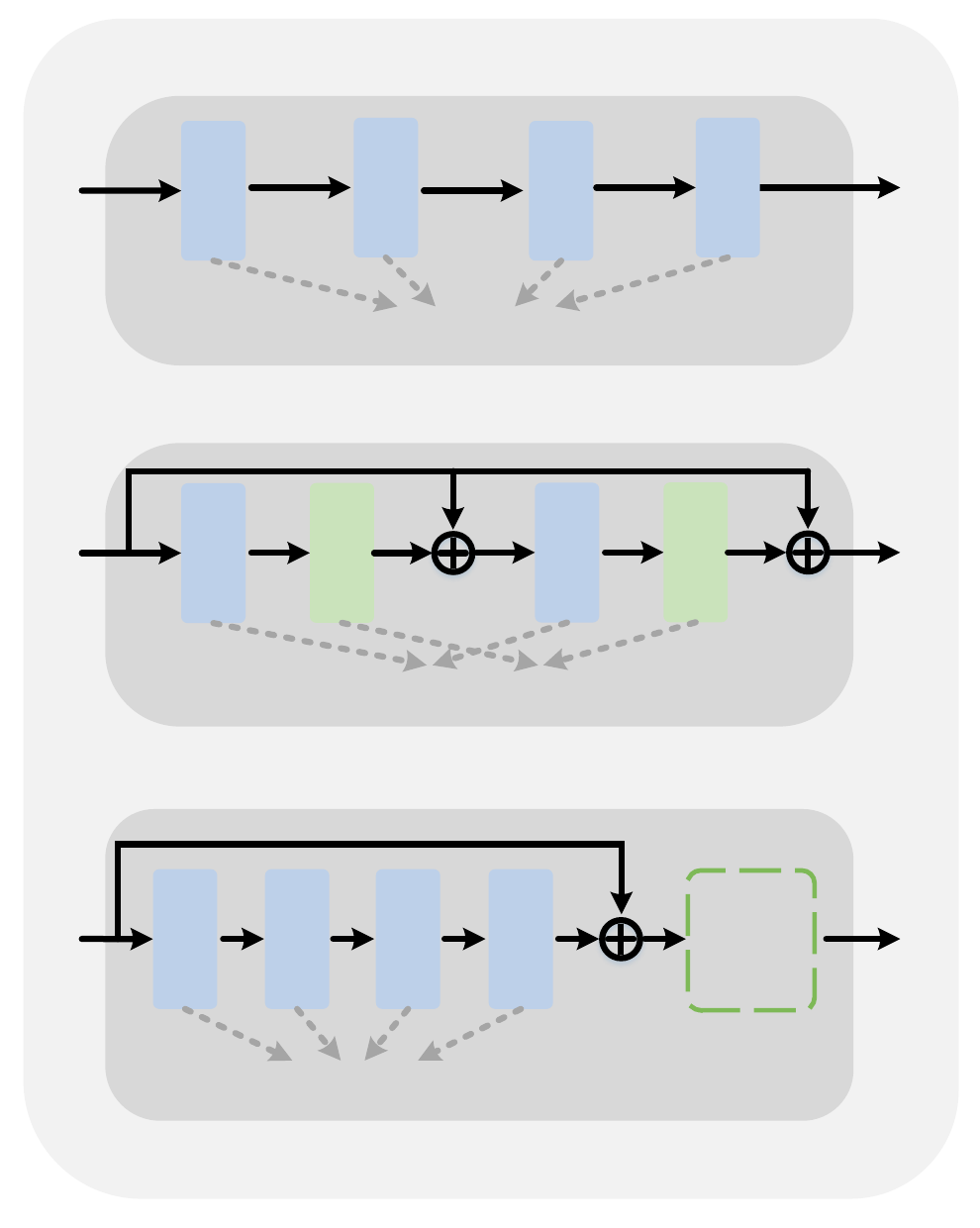}
        \put(57.5, 20.6){\scriptsize~\shortstack{Gate\\Unit}}
        \put(9, 95){\scriptsize~\textbf{Recursive Mechanisms}}
        \put(3, 84){\small~x}
        \put(73, 84.6){\small~y}
        \put(15.5, 81.6){\scriptsize~\rotatebox{90}{Conv}}
        \put(29.5, 81.6){\scriptsize~\rotatebox{90}{Conv}}
        \put(43.5, 81.6){\scriptsize~\rotatebox{90}{Conv}}
        \put(57, 81.6){\scriptsize~\rotatebox{90}{Conv}}
        \put(28, 72){\scriptsize~Shared weights}
        \put(33, 66.5){\scriptsize~DRCN~\cite{kim2016deeply}}

        \put(3, 54){\small~x}
        \put(73, 54.6){\small~y}
        \put(15.2, 51.2){\scriptsize~\rotatebox{90}{Conv}}
        \put(26, 51.2){\scriptsize~\rotatebox{90}{Conv}}
        \put(44.1, 51.2){\scriptsize~\rotatebox{90}{Conv}}
        \put(54.6, 51.2){\scriptsize~\rotatebox{90}{Conv}}
        \put(28, 42){\scriptsize~Shared weights}
        \put(33, 36.5){\scriptsize~DRRN~\cite{tai2017image}}

        \put(3, 22.5){\small~x}
        \put(73, 23){\small~y}
        \put(13.3, 19.4){\scriptsize~\rotatebox{90}{Conv}}
        \put(22.2, 19.4){\scriptsize~\rotatebox{90}{Conv}}
        \put(31.3, 19.4){\scriptsize~\rotatebox{90}{Conv}}
        \put(40.5, 19.4){\scriptsize~\rotatebox{90}{Conv}}
        \put(19, 10){\scriptsize~Shared weights}
        \put(33, 4.4){\scriptsize~MemNet~\cite{tai2017memnet}}
    \end{overpic}
    \caption{Utilizing a recursive mechanism to guide efficient structural designs. The symbol $\bm{\bigoplus}$ represents element-wise addition.} 
    \label{fig:recursive}
\end{figure}

\textbf{Recursive Mechanisms:} Recursive layers have been employed to reduce complexity and increase the receptive field as shown in Fig.~\ref{fig:recursive}. Kim~\etal~\cite{kim2016deeply} introduced a recursive layer to enlarge the receptive field and save computational resources. This recursive concept was further applied in DRRN~\cite{tai2017image} by adding residual recursive blocks to reduce the number of parameters and facilitate backpropagation. Tai~\etal~\cite{tai2017memnet} introduced a memory block consisting of a recursive unit and a gate unit to explicitly capture persistent memory through adaptive learning. Han~\cite{han2018image} formulated the SR network as a recurrent neural network (RNN)~\cite{liao2016bridging} and proposed a network that allows both LR and HR features to jointly contribute to learning the mappings.

\begin{figure}[!t]
    \centering
    \begin{overpic}[scale=.5]{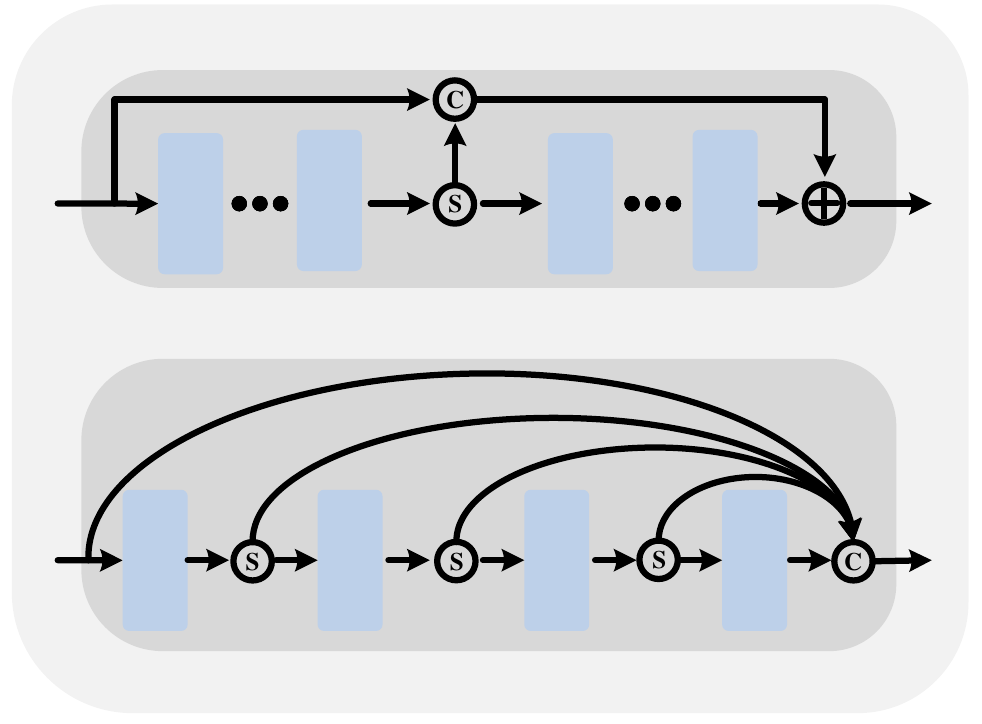}
        \put(8.5, 69){\scriptsize~\textbf{Distillation Blocks}}
        \put(42, 40){\scriptsize~IDN~\cite{Hui2018}}
        \put(1, 51.8){\small~x}
        \put(95, 52.5){\small~y}
        \put(17.5, 49.5){\scriptsize~\rotatebox{90}{Conv}}
        \put(31.2, 49.5){\scriptsize~\rotatebox{90}{Conv}}
        \put(57, 49.5){\scriptsize~\rotatebox{90}{Conv}}
        \put(71.8, 49.5){\scriptsize~\rotatebox{90}{Conv}}
        \put(41, 3){\scriptsize~IMDN~\cite{Hui2019}}
        \put(1, 15.5){\small~x}
        \put(95, 16){\small~y}
        \put(13.5, 12.6){\scriptsize~\rotatebox{90}{Conv}}
        \put(33.8, 12.6){\scriptsize~\rotatebox{90}{Conv}}
        \put(54.8, 12.6){\scriptsize~\rotatebox{90}{Conv}}
        \put(74.8, 12.6){\scriptsize~\rotatebox{90}{Conv}}
    \end{overpic}
    \caption{Distillation blocks. The symbols \textbf{\textcircled{\scriptsize~\hspace{-0.25em}S}} and \textbf{\textcircled{\scriptsize~\hspace{-0.3em}C}} indicate the split and concatenate operations.} 
    \label{fig:distillation}
\end{figure}
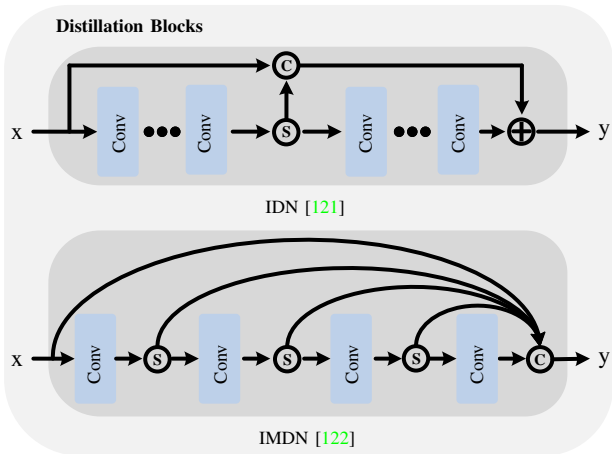

\textbf{Pruning:} Pruning techniques aim to reduce complexity by eliminating unnecessary parameters. Zhang~\etal~\cite{Zhang2021} proposed aligned structured sparsity learning to constrain the filter pruning process. Oh~\etal~\cite{oh2022attentive} developed a pruning method that determines the pruning ratio for N:M structured sparsity at each layer, optimizing the trade-off between efficiency and restoration accuracy. Wang~\etal~\cite{wang2023iterative} perform network pruning starting from random initialization and introduce a flexible thresholding technique to preserve sparse network trainability and achieve improved performance.

\textbf{Look-Up Table:} Jo~\etal~\cite{jo2021practical} introduced a precomputed look-up table (LUT) that retrieves HR output values for query LR input pixels. Building upon this work, Yang~\etal~\cite{yang2022seplut} analyzed the limitations of LUT and proposed a separable image-adaptive LUT that addresses these limitations. Ma~\etal~\cite{ma2022learning} proposed a series-parallel LUT to increase the receptive fields. RCLUT~\cite{liu2023reconstructed} divides the features into channel and spatial dimensions, exploring their respective relationships. This approach significantly increases the receptive field while keeping complexity low. Additionally, a convolutional block is introduced to enhance the interaction of information.

\textbf{Network Architecture Search:} Network architecture search (NAS) has been utilized to explore efficient model architectures. Song~\etal~\cite{song2020efficient} applied NAS to develop an efficient residual dense block. Chen~\etal~\cite{chen2020} employed NAS and decomposed upsampling operators to obtain a larger search space. Zhan~\etal~\cite{zhan2021achieving} utilized NAS to study effective pruning strategies for reducing model complexity. MoreMNAS~\cite{chu2021fast} method leverages neural architecture search with a multi-objective approach, combining evolutionary computation and reinforcement learning to automatically balance restoration capacity and model simplicity, resulting in models that outperform state-of-the-art methods in terms of FLOPS efficiency.

\textbf{Implicit Neural Representation:} Drawing inspiration from recent advancements in 3D reconstruction utilizing implicit neural representation~\cite{jiang2020local}, Chen~\etal~\cite{chen2021learning} introduced the concept of local implicit image function. This function predicts RGB values at specified coordinates by incorporating image coordinates and 2D deep features in the vicinity of those coordinates. Building upon this work, Lee~\etal~\cite{lee2022local} proposed the local texture estimator, which serves as a dominant-frequency estimator for natural images. This estimator enables implicit functions to capture intricate details while ensuring the continuous reconstruction of images. Yao~\etal~\cite{yao2023local} combined the implicit representation with normalizing flow and developed a visually pleasing arbitrary-scale model.

\textbf{Diverse Degradations:} To address the challenges posed by multiple and spatially variant degradations, Zhang~\etal~\cite{zhang2018learning} introduced the use of blur kernels and noise levels as inputs. Expanding on this idea, Zhang~\etal~\cite{Zhang2019} extended the applicability of bicubic degradation by employing a variable splitting technique to handle arbitrary blur kernels. IKC~\cite{gu2019blind} addresses blind super-resolution by iteratively correcting blur kernel estimates to reduce artifacts, and utilizes spatial feature transform (SFT) layers to effectively manage multiple unknown blur kernels, leading to improved SR performance. DAN~\cite{huang2020unfolding} addresses the blind super-resolution problem by using an alternating optimization approach, integrating Restorer and Estimator modules in a single end-to-end trainable network, where the blur kernel estimation and SR image restoration mutually enhance each other for improved performance. Kim~\etal~\cite{KIM2021} proposed a two-branch network that learns spatially variant information and adaptively fuses degradation kernel information. DASR~\cite{wang2021unsupervised} addresses blind super-resolution by learning abstract degradation representations instead of relying on explicit degradation estimation, enabling flexible adaptation to various degradations and achieving state-of-the-art performance. Hui~\etal~\cite{Hui2021} integrated reinforcement learning~\cite{lillicrap2015continuous} into SR to guide kernel estimation using a non-differentiable objective function. Xu~\etal~\cite{Xu2020} considered multiple degrading effects and incorporated dynamic convolution to handle various types of degradations. Additionally, Luo~\etal~\cite{luo2022deep} transferred the blurring process from the LR space and introduced a dynamic deep linear filter to generate appropriate deblurring kernels for producing clean features.
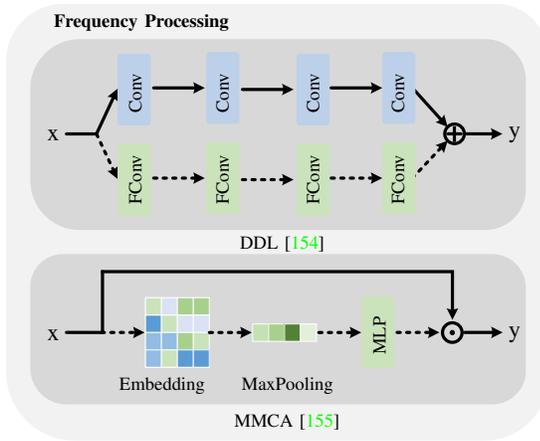
\begin{figure}
    \centering
    \begin{overpic}[scale=.4]{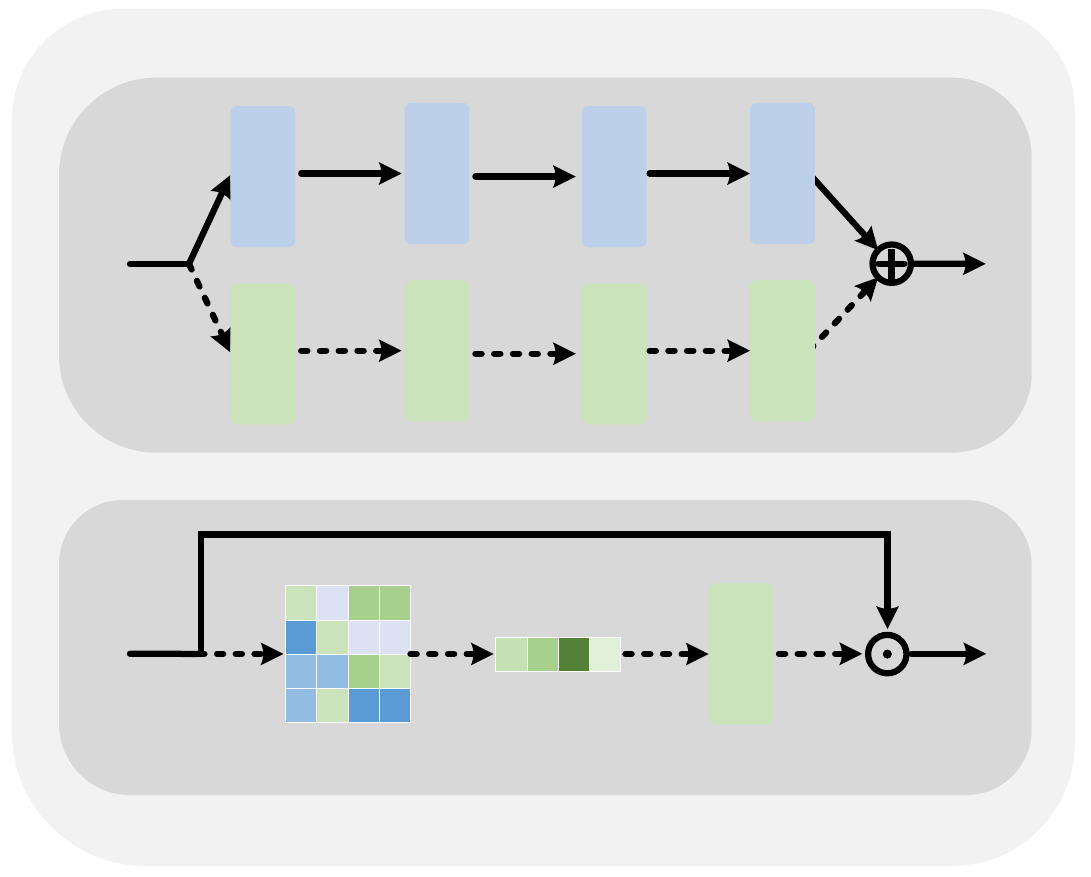}
        \put(8.5, 75.5){\scriptsize~\textbf{Frequency Processing}}
        \put(42, 35.5){\scriptsize~DDL~\cite{liu2023spectral}}
        \put(7, 55){\small~x}
        \put(90, 55.5){\small~y}
        \put(22, 60){\scriptsize~\rotatebox{90}{Conv}}
        \put(37.7, 60){\scriptsize~\rotatebox{90}{Conv}}
        \put(54, 60){\scriptsize~\rotatebox{90}{Conv}}
        \put(69.3, 60){\scriptsize~\rotatebox{90}{Conv}}
        \put(41, 3.5){\scriptsize~MMCA~\cite{magid2021dynamic}}
        \put(7, 19){\small~x}
        \put(90, 19.5){\small~y}
        \put(22, 43.5){\scriptsize~\rotatebox{90}{FConv}}
        \put(37.7, 43.5){\scriptsize~\rotatebox{90}{FConv}}
        \put(54, 43.5){\scriptsize~\rotatebox{90}{FConv}}
        \put(69.3, 43.5){\scriptsize~\rotatebox{90}{FConv}}
        \put(20.3, 10){\scriptsize~Embedding}
        \put(42.3, 10){\scriptsize~MaxPooling}
        \put(66, 16.5){\scriptsize~\rotatebox{90}{MLP}}
    \end{overpic}
    \caption{Frequency processing. The dashed line indicates the process in the frequency domain. FConv represents the complex convolutional operation. The symbol $\bm{\bigoplus}$ and $\bm{\bigodot}$ indicate element-wise multiplication and element-wise multiplication.} 
    \label{fig:frequency}
\end{figure}

\textbf{Frequency Processing:} Frequency information has proven valuable for effective representation learning in various approaches as shown in Fig.~\ref{fig:frequency}. Helou~\etal~\cite{ElHelou2020} introduced stochastic masking of high-frequency bands, enhancing the capability of restoring high frequencies. Xie~\etal~\cite{xie2021learning} explored a frequency-aware dynamic network that partitions the input based on its coefficients in the discrete cosine transform domain. Qiu~\etal~\cite{qiu2019embedded} addressed different frequencies with complexity modules and designed a block residual module to restore information from easy-to-recover to hard-to-recover frequencies. Zhong~\etal~\cite{Zhong2019} focused on SR in the wavelet domain, utilizing sub-networks to predict sub-band coefficients and applying inverse discrete wavelet transform for HR image reconstruction. Magid~\etal~\cite{magid2021dynamic} introduced a dynamic high-pass filtering module to preserve high-frequency information and a multi-spectral channel attention module to recalibrate responses across different frequencies. Liu~\etal~\cite{liu2023spectral} explored the spectral Bayesian uncertainty and introduced complex layers to perform dual-domain learning. They introduced a frequency loss defined as 
\begin{equation}
\mathcal{L}_{freq.} = \frac{1}{HW}\sum_{i=0}^{H-1}\sum_{j=0}^{W-1}\lVert \Phi(X_{sr})(i,j) - \Phi(X_{hr})(i,j)\rVert_1,
\label{eq:freq.}
\end{equation}
where $\Phi(\cdot)$ represents the Fourier transformation, and $H$ and $W$ indicates the height and the width of frequency map.
\begin{figure}
    \centering
    \begin{overpic}[scale=.4]{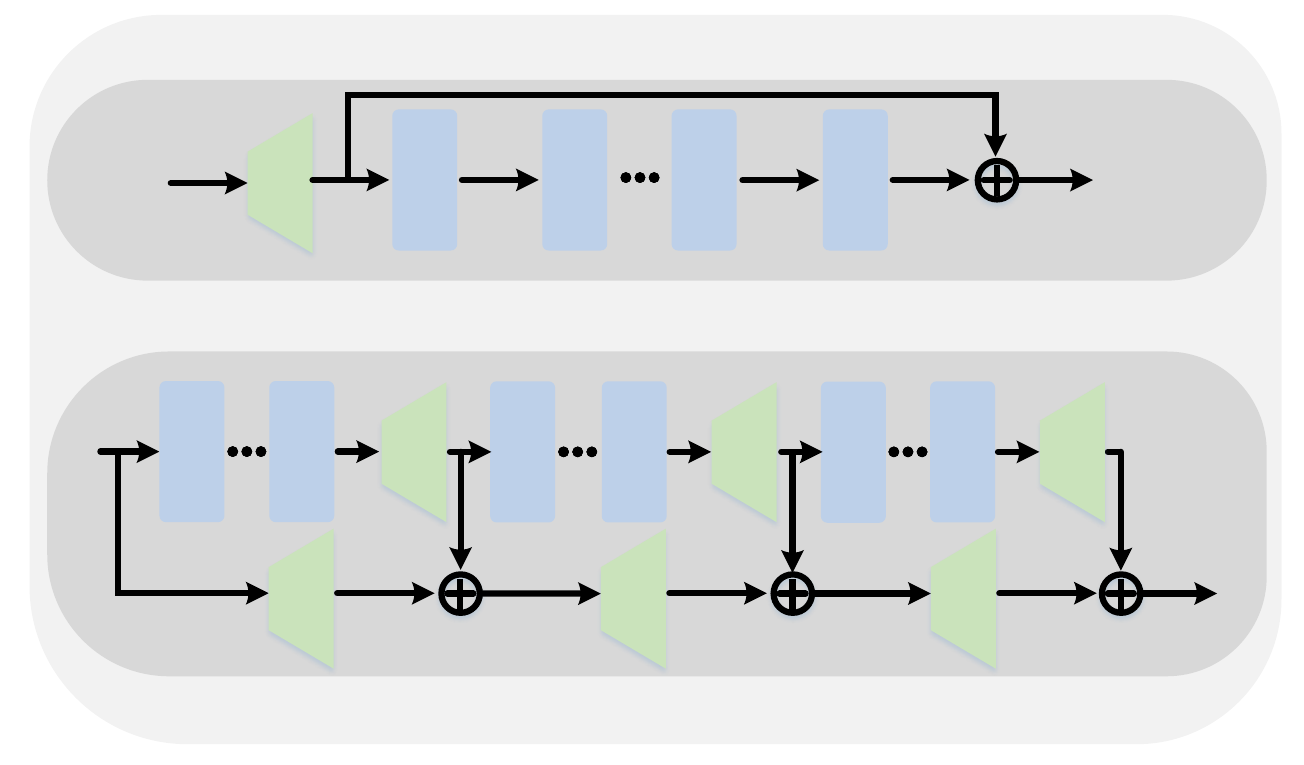}
        \put(8.5, 54){\scriptsize~\textbf{\revision{Enlarging Receptive Field}}}
        \put(42, 33.8){\scriptsize~VDSR~\cite{kim2016accurate}}
        \put(9.1, 43.5){\small~x}
        \put(83.8, 44){\small~y}
        \put(20, 42.5){\scriptsize~\rotatebox{90}{UP}}
        \put(30.5, 41){\scriptsize~\rotatebox{90}{Conv}}
        \put(42.2, 41){\scriptsize~\rotatebox{90}{Conv}}
        \put(52.2, 41){\scriptsize~\rotatebox{90}{Conv}}
        \put(63.5, 41){\scriptsize~\rotatebox{90}{Conv}}
        \put(41, 3.2){\scriptsize~LapSRN~\cite{lai2017deep}}
        
        \put(4, 23){\small~x}
        \put(93, 12.5){\small~y}
        \put(30, 22.2){\scriptsize~\rotatebox{90}{UP}}
        \put(13, 21){\scriptsize~\rotatebox{90}{Conv}}
        \put(21.5, 21){\scriptsize~\rotatebox{90}{Conv}}
        \put(38.5, 21){\scriptsize~\rotatebox{90}{Conv}}
        \put(46.7, 21){\scriptsize~\rotatebox{90}{Conv}}
        \put(55.5, 22.2){\scriptsize~\rotatebox{90}{UP}}
        \put(64, 21){\scriptsize~\rotatebox{90}{Conv}}
        \put(71.5, 21){\scriptsize~\rotatebox{90}{Conv}}
        \put(80.5, 22.2){\scriptsize~\rotatebox{90}{UP}}
        
        \put(21.5, 10.8){\scriptsize~\rotatebox{90}{UP}}
        \put(47, 10.8){\scriptsize~\rotatebox{90}{UP}}
        \put(72.5, 10.8){\scriptsize~\rotatebox{90}{UP}}
    \end{overpic}
    \caption{\revision{VDSR~\cite{kim2016accurate} stacked more than 20 layers to enlarge the receptive field, while LapSRN~\cite{lai2017deep} introduced a hierarchical structure to capture diverse scale features. The symbol $\bm{\bigoplus}$ indicates element-wise addition.}} 
    \label{fig:enlarge}
\end{figure}
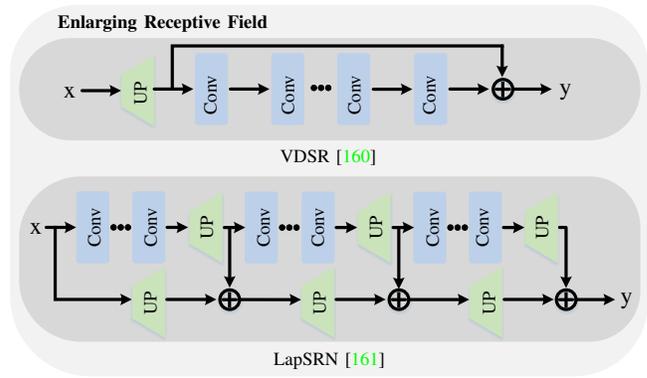

\textbf{\revision{Enlarging Receptive Field:}} \revision{Stacking more blocks and utilizing a hierarchy mechanism are introduced into SISR to increase the receptive field as shown in Fig.~\ref{fig:enlarge}.} \revision{Kim~\etal~\cite{kim2016accurate} proposed the use of very deep networks with 20 cascaded layers to obtain a larger receptive field.} Building upon this idea, Lim~\etal~\cite{Lim2017} introduced an enhanced deep super-resolution network, which achieved significant performance improvements by removing the BN~\cite{ioffe2015batch} layer in the residual block and adding more convolution layers. Lai~\etal~\cite{lai2017deep} developed the laplacian pyramid super-resolution network to progressively reconstruct the sub-band residuals of HR images. Fan~\etal~\cite{fan2020scale} focused on scale-invariant modeling and proposed scale-wise convolution. They also built a feature pyramid representation through progressive bilinear downscaling to improve restoration performance. Bhardwaj~\etal~\cite{bhardwaj2022collapsible} introduced the over-parameterization mechanism~\cite{arora2018optimization,ding2019acnet} to enlarge the network and achieved promising results.

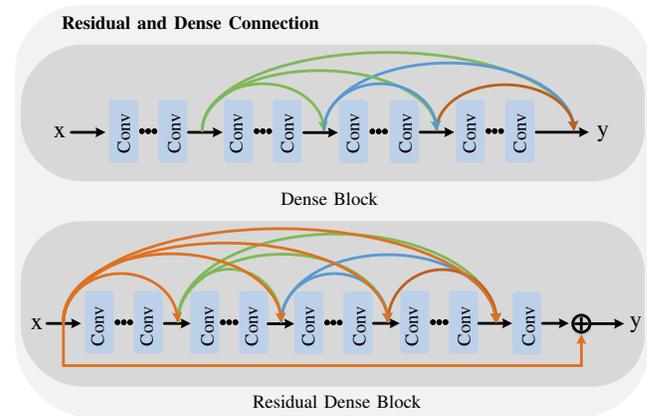
\begin{figure}
    \centering
    \begin{overpic}[scale=.35]{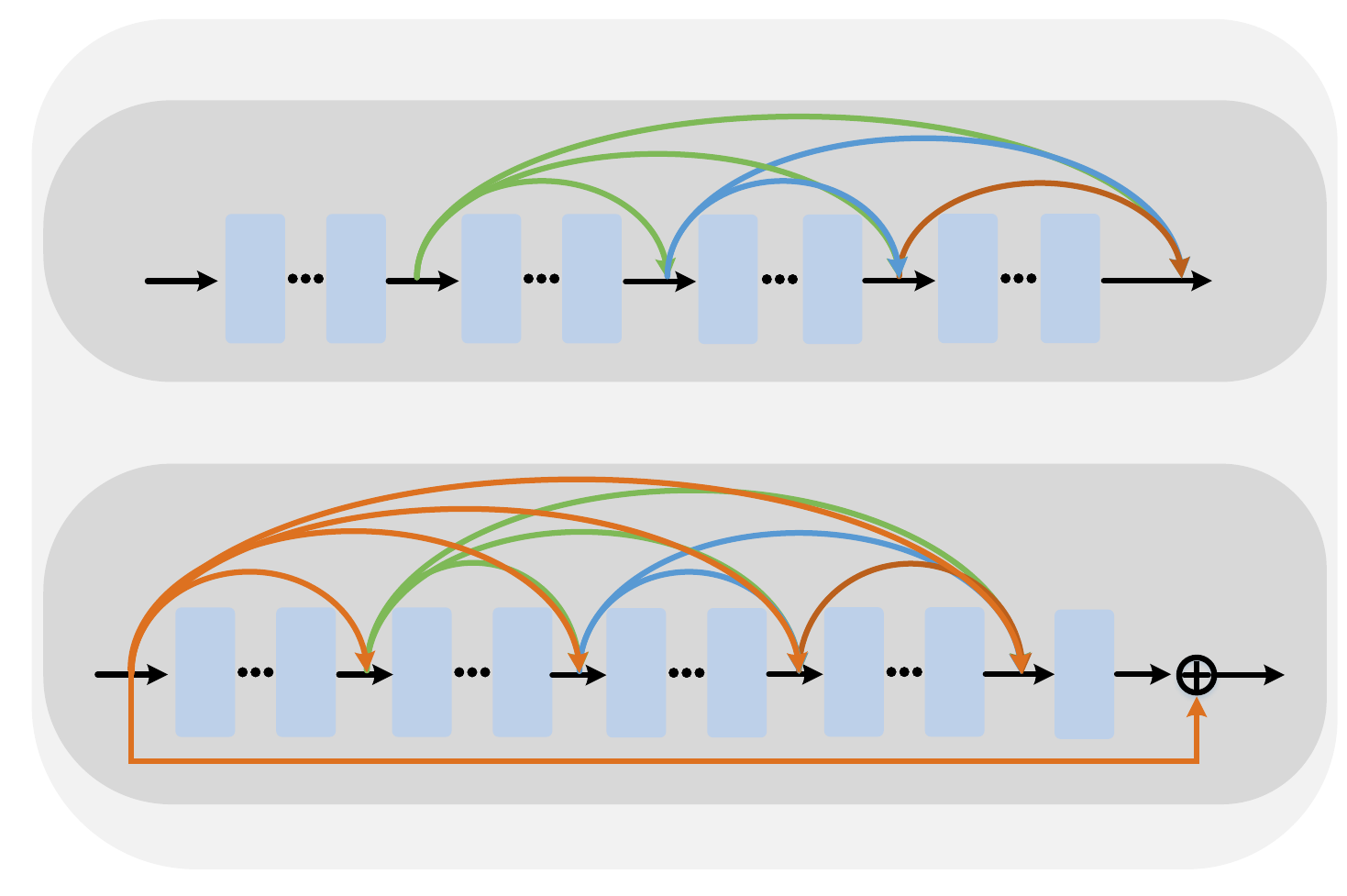}
        \put(8.5, 61.5){\scriptsize~\textbf{Residual and Dense Connection}}
        \put(42, 34.5){\scriptsize~Dense Block}
        \put(7, 44.8){\small~x}
        \put(90, 45){\small~y}
        \put(17, 42.5){\scriptsize~\rotatebox{90}{Conv}}
        \put(24.7, 42.5){\scriptsize~\rotatebox{90}{Conv}}
        \put(34.2, 42.5){\scriptsize~\rotatebox{90}{Conv}}
        \put(42, 42.5){\scriptsize~\rotatebox{90}{Conv}}
        \put(52, 42.5){\scriptsize~\rotatebox{90}{Conv}}
        \put(60, 42.5){\scriptsize~\rotatebox{90}{Conv}}
        \put(69.8, 42.5){\scriptsize~\rotatebox{90}{Conv}}
        \put(77.5, 42.5){\scriptsize~\rotatebox{90}{Conv}}
        \put(37.7, 3.5){\scriptsize~Residual Dense Block}
        \put(3.5, 15.5){\small~x}
        \put(95, 15.8){\small~y}
        \put(13, 13){\scriptsize~\rotatebox{90}{Conv}}
        \put(20.5, 13){\scriptsize~\rotatebox{90}{Conv}}
        \put(29.8, 13){\scriptsize~\rotatebox{90}{Conv}}
        \put(36.5, 13){\scriptsize~\rotatebox{90}{Conv}}
        \put(45.2, 13){\scriptsize~\rotatebox{90}{Conv}}
        \put(53, 13){\scriptsize~\rotatebox{90}{Conv}}
        \put(61, 13){\scriptsize~\rotatebox{90}{Conv}}
        \put(69, 13){\scriptsize~\rotatebox{90}{Conv}}
        \put(78.8, 13){\scriptsize~\rotatebox{90}{Conv}}
    \end{overpic}
    \caption{Discrepancy between dense connection and residual dense connection. The symbol $\bm{\bigoplus}$ indicates element-wise addition.}
    \label{fig:residual_dense}
\end{figure}

\begin{figure}
    \centering
    \begin{overpic}[scale=.35]{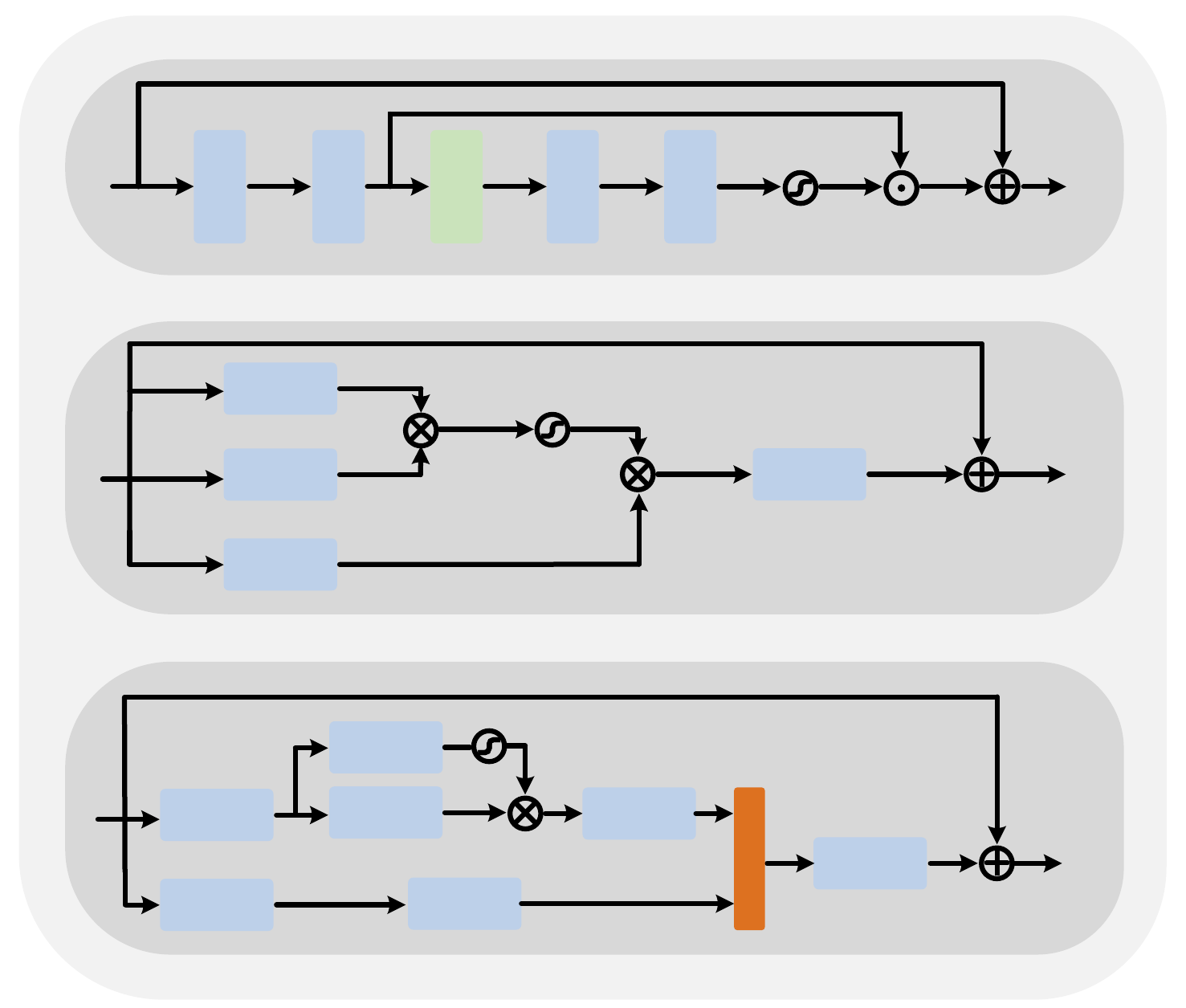}
        \put(8.5, 81){\scriptsize~\textbf{Attention Mechanisms}}
        \put(39, 59.2){\scriptsize~Channel Attention~\cite{zhang2018image}}
        \put(6, 68.5){\small~x}
        \put(89.5, 68.8){\small~y}
        \put(16.6, 66){\scriptsize~\rotatebox{90}{Conv}}
        \put(27.2, 66){\scriptsize~\rotatebox{90}{Conv}}
        \put(36.2, 64.8){\fontsize{6.5pt}{14pt}\selectfont~~\rotatebox{90}{Pooling}}
        \put(45.8, 66){\scriptsize~\rotatebox{90}{Conv}}
        \put(56, 66){\scriptsize~\rotatebox{90}{Conv}}
        
        \put(37, 30.5){\scriptsize~Non-local Attention~\cite{zhang2019residual}}
        \put(5, 44){\small~x}
        \put(89.5, 44.5){\small~y}
        \put(19.5, 52){\scriptsize~Conv}
        \put(19.5, 44){\scriptsize~Conv}
        \put(19.5, 36.5){\scriptsize~Conv}
        \put(64.5, 44){\scriptsize~Conv}

        \put(37, 2){\scriptsize~Pixel Attention~\cite{Zhao2020}}
        \put(4.5, 15){\small~x}
        \put(89.6, 11.8){\small~y}
        \put(13.5, 15.5){\scriptsize~Conv}
        \put(14, 8){\scriptsize~Conv}
        \put(28.5, 21){\scriptsize~Conv}
        \put(28.5, 15.5){\scriptsize~Conv}
        \put(35.2, 8){\scriptsize~Conv}
        \put(50, 16){\scriptsize~Conv}
        \put(69.5, 11.5){\scriptsize~Conv}
        \put(61.4, 7.7){\fontsize{6.5pt}{14pt}\selectfont~\rotatebox{90}{Concate}}
    \end{overpic}
    \caption{Illustration of channel attention, spatial attention, and pixel attention. The symbols $\bm{\bigoplus}$, $\bm{\bigotimes}$, and $\bm{\bigodot}$ indicate element-wise addition, matrix multiplication, and element-wise multiplication, respectively.}
    \label{fig:attention}
\end{figure}
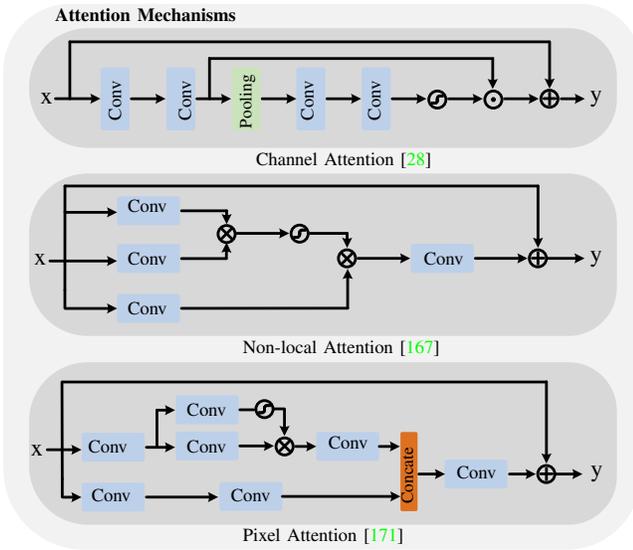

\textbf{Attention mechanisms:} Attention mechanisms have been widely used to improve information interaction as shown in Fig.~\ref{fig:attention}. Zhang~\etal~\cite{zhang2018image} proposed the residual channel attention networks to assign diverse weights to each channel and enhance performance. Zhang~\etal~\cite{zhang2019residual} designed local and non-local attention blocks to capture long-range dependencies between pixels and focus more on challenging parts. Zamir~\etal~\cite{Zamir2020} introduced multi-scale information interaction and attention mechanisms to combine contextual information and preserve HR spatial details. Dai~\etal~\cite{dai2019second} introduced a second-order attention network for more powerful feature expression and feature correlation learning. Niu~\etal~\cite{niu2020single} proposed a holistic attention network consisting of a layer attention module and a channel-spatial attention module to model interdependencies among layers, channels, and positions. Zhao~\etal~\cite{Zhao2020} proposed pixel attention to obtain 3D attention maps, achieving promising results with low parameter cost. Mei~\etal~\cite{mei2021image} combined the non-local attention mechanism with sparse representation and rectified non-local attention with spherical locally sensitive hashing to reduce computational cost. 

\resubmit{Unlike the traditional attention mechanisms that typically focus on capturing relationships between pixels, Transformer-based self-attention mechanisms capture dependencies between patches in the input. This difference makes Transformer attention more efficient, as it operates over a higher-level representation, enabling it to handle long-range dependencies with fewer computations compared to pixel-based attention. Additionally, Transformer models integrate feed-forward networks (FFNs) to further enhance the feature extraction process, which leads to better performance.}

\textbf{Back Projection:} Haris~\etal~\cite{Haris2018} proposed deep back-projection networks, which utilized iterative up- and down-sampling layers to provide an error feedback mechanism for projection errors at each stage. Michelini~\etal~\cite{Michelini2019} extended the iterative back–projections (IBP)~\cite{IraniM1991} to multi-level IBP and demonstrated its effectiveness comparable to classic IBP. Li~\etal~\cite{li2019feedback} explored the feedback mechanism present in HVS and proposed a super-resolution feedback network to refine low-level representations using high-level information.

\textbf{Structural Improvements:} Several structural improvements have been proposed to enhance the representation learning capabilities of SISR models. \revision{Building on previous work by Shocher~\etal~\cite{shocher2018zero}, which employs zero-shot learning methods, MZSR~\cite{soh2020meta} integrates meta-transfer learning with zero-shot super-resolution, allowing for rapid adaptation to specific image conditions with minimal gradient updates, and effectively utilizing both external and internal information for flexible and efficient super-resolution.} Guo~\etal~\cite{guo2020closed} introduced a dual regression constraint and devised a closed-loop framework to reduce the space of possible functions. Xiao~\etal~\cite{xiao2020invertible} developed the invertible rescaling net, which employed deliberately designed objectives to produce visually-pleasing LR images. Hu~\etal~\cite{hu2019meta} introduced the meta-upscale module, which dynamically predicts the weights of the upscale filter, enabling arbitrary scale factor super-resolution with a single network. Wang~\etal~\cite{wang2021learning} incorporated scale-aware feature blocks and upsampling layers to learn scale-arbitrary SR based on fixed-scale networks. Maeda~\etal~\cite{Maeda2022eccv} proposed a deep dictionary learned from the network to handle diverse degradations not present in the training set.

\indent\textbf{Residual and Dense Connection:} Several methods have explored the use of residual and dense connections to improve information interaction as shown in Fig.~\ref{fig:residual_dense}. Lai~\etal~\cite{Lai2019} introduced local skip connections, parameter sharing, and multi-scale training strategies to maintain information interaction. Tong~\etal~\cite{tong2017image} added dense connections to the network and demonstrated their effectiveness in achieving good performance on low-level tasks. Zhang~\etal~\cite{Zhang2018} combined residuals and dense connections to obtain hierarchical features and achieved favorable performance. Ahn~\etal~\cite{Ahn2018} designed an architecture that cascades more residual blocks and dense connections. Li~\etal~\cite{Li2020} introduced a set of predefined filters and used a CNN to learn the coefficients, enabling a linear combination of these filters to produce the final results. Liu~\etal~\cite{liu2020residual} proposed a residual feature aggregation framework to fully extract hierarchical features on the residual branches.

\subsubsection{\textbf{Transformer-based SISR}}
Due to their ability to model long-range dependencies, Transformers have achieved superior super-resolution results compared to CNNs in super-resolution tasks. However, the high-dimensional matrix operations in Transformers make these methods significantly more complex than CNN-based approaches. Here, we categorize Transformer-based methods into three subclasses based on reducing complexity and enhancing performance, which we will introduce accordingly.
\begin{figure}[!t]
    \centering
    \begin{overpic}[scale=.4]{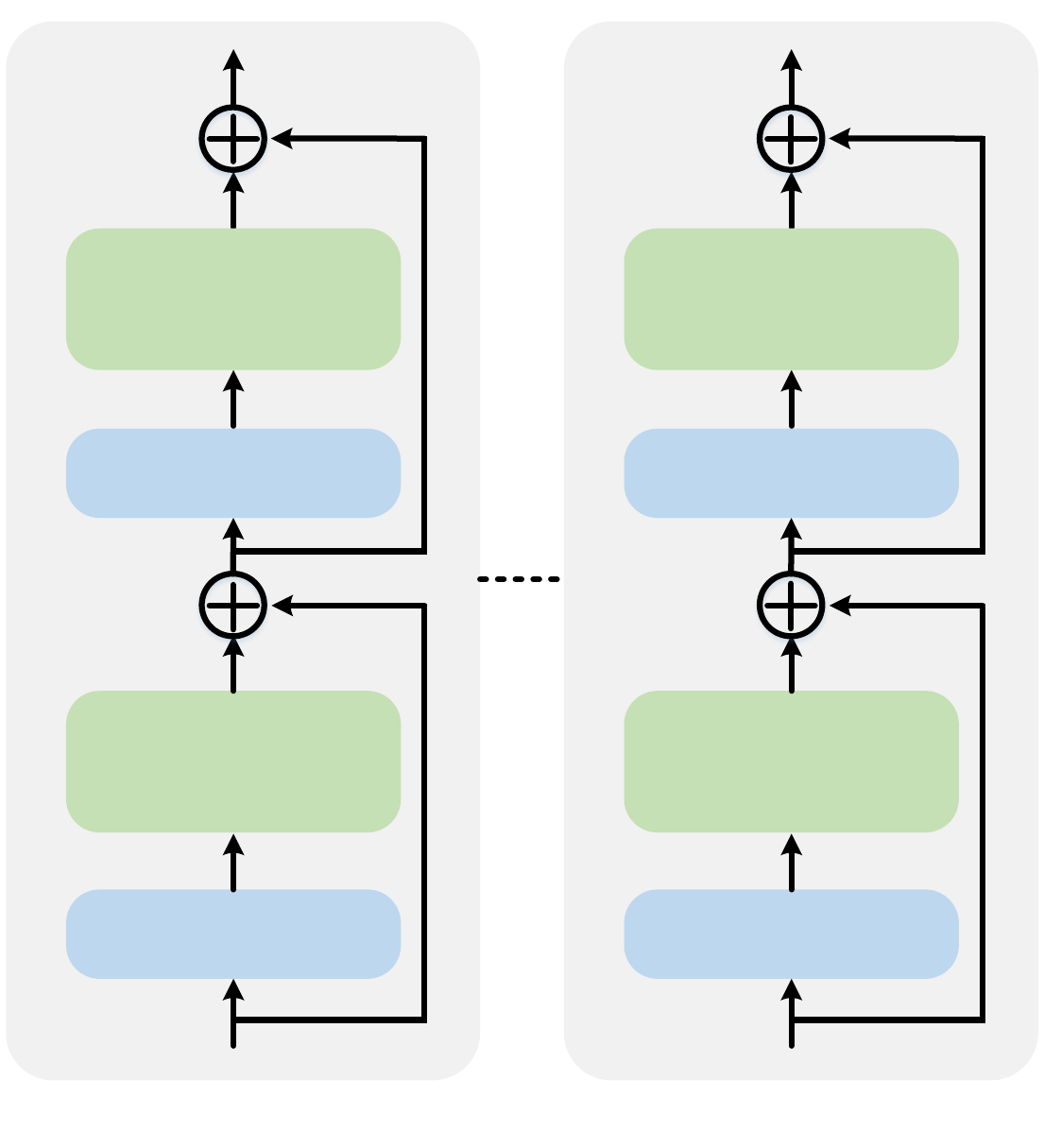}
        \put(8.8, 1.5){\scriptsize~\textbf{Swin Block-R}}
        \put(11, 71.5){\scriptsize~\shortstack{Point-wise \\ Feed-forward}}
        \put(11, 57.5){\scriptsize~Layer Norm}
        \put(9, 31.2){\scriptsize~\shortstack{Window-based \\ Attention}}
        \put(11, 17.5){\scriptsize~Layer Norm}

        \put(58.8, 1.5){\scriptsize~\textbf{Swin Block-S}}
        \put(59.5, 71.5){\scriptsize~\shortstack{Point-wise \\ Feed-forward}}
        \put(61, 57.5){\scriptsize~Layer Norm}
        \put(58.8, 31.2){\scriptsize~\shortstack{Shift-window \\ Attention}}
        \put(61, 17.5){\scriptsize~Layer Norm}
        
    \end{overpic}
    \caption{The framework of swin transformer. -R indicates the regular window attention and -S indicates the shift window attention.}
    \label{fig:swinT}
\end{figure}
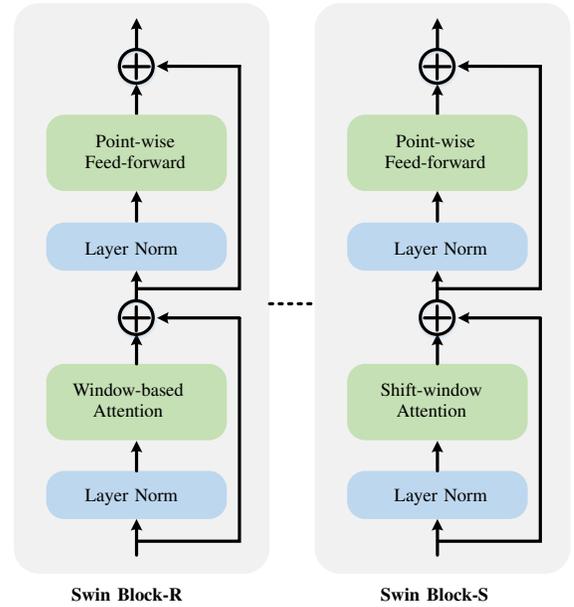

\textbf{Efficient Structural Designs:} Liang~\etal~\cite{liang2021swinir} presented a strong baseline model called SwinIR for image restoration, based on the swin transformer~\cite{liu2021swin}. The swin transformer architecture is depicted in Fig. \ref{fig:swinT}. To reduce computational complexity, the authors designed a window-based attention mechanism and incorporated shift operations to enlarge perceptual fields. Choi~\etal~\cite{choi2023n} introduced the N-Gram mechanism into the image domain to enlarge the perceptive fields. Meanwhile, SRFormer~\cite{zhou2023srformer} evaluated that larger window size can lead to better performance and proposed a permutation mechanism to enlarge the window size while maintaining efficient parameters scale. Gao~\etal~\cite{gao2022lightweight} introduced a lightweight bimodal network that integrates CNN and transformer structures for SISR, achieving a better trade-off between performance and complexity. Lu~\etal~\cite{lu2022transformer} proposed a lightweight transformer backbone that captures long-term dependencies between similar image patches using an efficient transformer and efficient multi-head attention mechanism. Zhang~\etal~\cite{zhang2022efficient} introduced shift convolution and group-wise multi-scale self-attention to reduce the complexity of transformer-based architectures. Li~\etal~\cite{li2023dlgsanet} proposed a dynamic convolution and a sparse self-attention mechanism to explore the most useful similarity values to reconstruct the images. SAFMN~\etal~\cite{sun2023spatially} utilized multi-scale strategy to explore global information and conducted a convolutional channel mixer to introduce local information.

\textbf{Pre-trained Strategies:} Chen~\etal~\cite{chen2021pre} proposed a pre-trained model called image processing transformer, demonstrating that pre-trained transformers significantly enhance performance for low-level tasks. Li~\etal~\cite{li2021efficient} conducted a comprehensive analysis of pre-training effects and proposed a versatile model capable of addressing various low-level tasks.

\textbf{Hybrid Attenion Blocks:} Yang~\etal~\cite{Yang2020} introduced a transformer structure that utilizes soft and hard attention maps to establish feature correspondences between LR and reference images. Chen~\etal~\cite{chen2022activating} proposed a hybrid attention transformer that combines channel attention and self-attention schemes. To mitigate the patch effect caused by window-based attention, the authors introduced an overlapping window partition mechanism. Wang~\etal~\cite{wang2023deep} considered the scale-equivariance characteristic and proposed an adaptive feature extractor module to incorporate the scale prior and a neural kriging upsampler to learn feature distance and similarity. Cao~\etal~\cite{cao2022ciaosr} proposed an attention-in-attention network to implicitly explore scale non-local information. Omni-SR~\cite{wang2023omni} leveraged the interaction between spatial and channel information and proposed a diverse transformer attention block to enlarge the effective receptive field. Chen~\etal~\cite{chen2023cascaded} integrated cross-scale attention and local frequency information and proposed a local implicit transformer to adapt to arbitrary SISR. DAT~\cite{chen2023dual} incorporates self-attention mechanisms across both spatial and channel dimensions and introduces a dual feature aggregation model. Simultaneously, it leverages convolution layers and adaptive interaction modules to facilitate interactions between local and global information. CRAFT~\cite{li2023feature} investigates the discrepancy in high-frequency dependencies between CNN and transformer structures and introduces the high-frequency enhancement module to capture high-frequency information. Simultaneously, it employs channel-wise attention to blend diverse frequency characteristics for the enhancement of global representations. CFAT~\cite{ray2024cfat} introduced a novel attention mechanism that combines triangular window-based attention with rectangular window and channel attention. This integration enables the model to capture richer contextual information, leading to improved performance.

\subsection{\textbf{Generative Models}}
\subsubsection{\textbf{GAN-based SISR}}
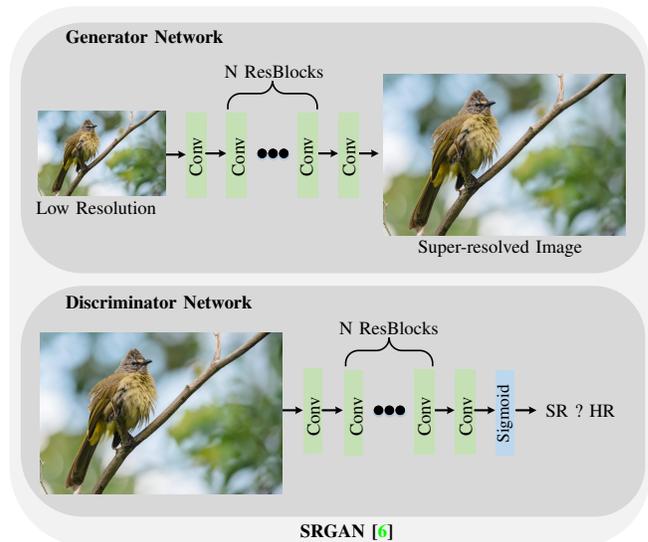
\begin{figure}
\centering
    \begin{overpic}[scale=.27]{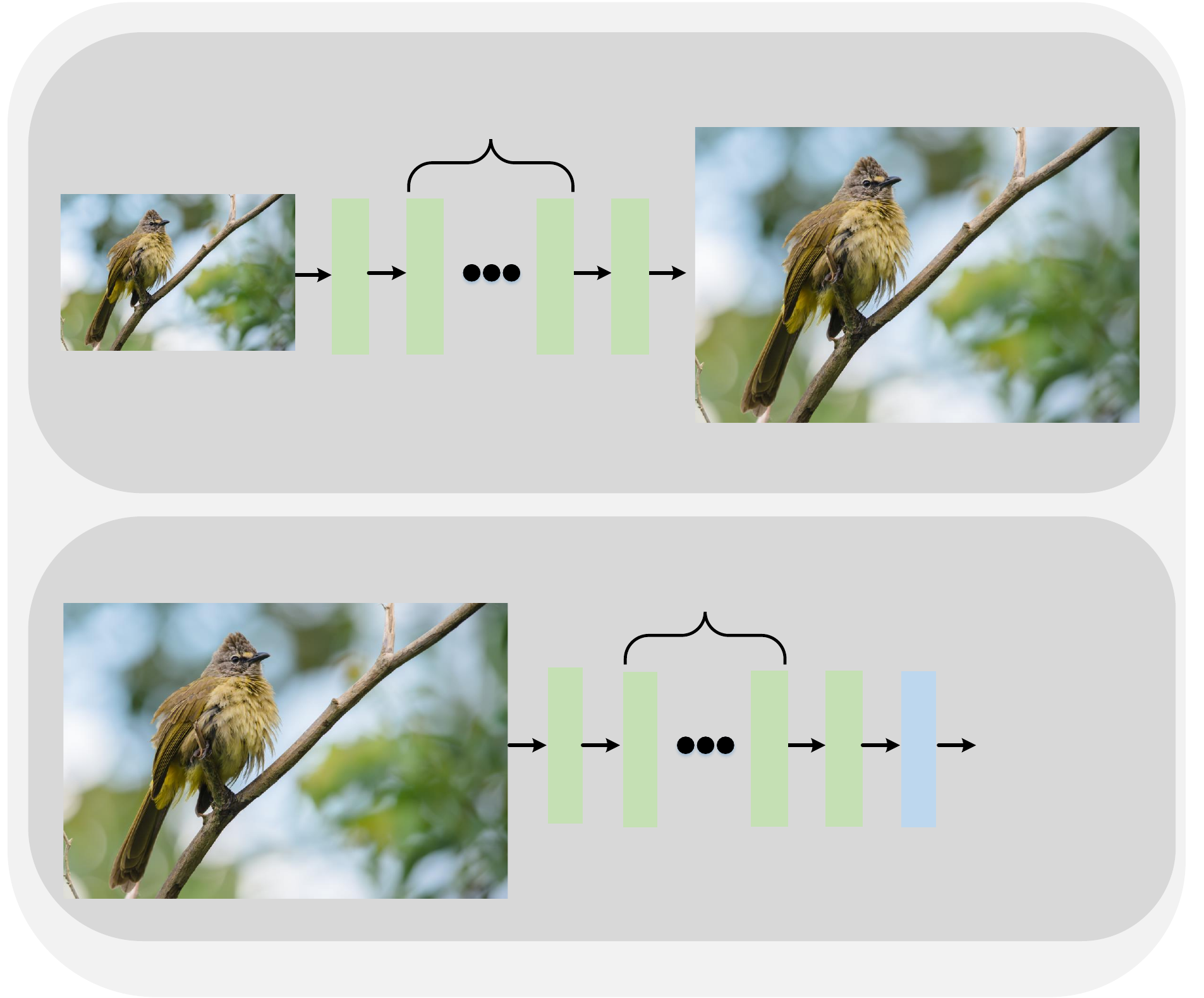}
        \put(8.2, 77){\scriptsize~\textbf{Generator Network}}
        \put(3.7, 51){\scriptsize~Low Resolution}
        \put(32.5, 71.8){\scriptsize~N ResBlocks}
        \put(62, 45){\scriptsize~Super-resolved Image}
        \put(27.1, 56.7){\scriptsize~\rotatebox{90}{Conv}}
        \put(33.6, 56.7){\scriptsize~\rotatebox{90}{Conv}}
        \put(44, 56.7){\scriptsize~\rotatebox{90}{Conv}}
        \put(50.5, 56.7){\scriptsize~\rotatebox{90}{Conv}}

        \put(50, 32.5){\scriptsize~N ResBlocks}
        \put(81.5, 20){\scriptsize~SR ? HR}
        \put(8.2, 36.5){\scriptsize~\textbf{Discriminator Network}}
        \put(44, 1.5){\scriptsize~\textbf{SRGAN~\cite{Ledig2017}}}
        \put(45.1, 17){\scriptsize~\rotatebox{90}{Conv}}
        \put(51.6, 17){\scriptsize~\rotatebox{90}{Conv}}
        \put(62, 17){\scriptsize~\rotatebox{90}{Conv}}
        \put(68.3, 17){\scriptsize~\rotatebox{90}{Conv}}
        \put(74.4, 16){\fontsize{6.5pt}{14pt}\selectfont~\rotatebox{90}{Sigmoid}}
    
    \end{overpic}
    \caption{The framework of SRGAN~\cite{Ledig2017}. It pioneered the utilization of an exceptionally deep ResNet~\cite{he2016deep} architecture, incorporating the concept of GANs, in order to create a perceptual loss function that faithfully emulates human perception and facilitates the attainment of photo-realistic SISR.}
\label{fig:srgan}
\end{figure}

\
\newline
\textbf{Structural Modifications:} To recover finer texture details, GAN-based SISR methods have been introduced with various structural enhancements. The pioneering approach, SRGAN~\cite{Ledig2017}, consists of a generator and a discriminator, as depicted in Fig.\ref{fig:srgan}. \revision{The generator super-resolves the LR image, while the discriminator learns to distinguish between the reconstructed image and real HR images.} Building on SRGAN, Sønderby\etal~\cite{Snderby2017} explored the connection between GANs and amortized variational inference, which aligns with principles of variational autoencoders.
To address the problem of high-frequency noise often introduced by GAN-based methods, Park~\etal~\cite{park2018srfeat} proposed an additional discriminator in the feature domain, improving the quality of reconstructed textures. ESRGAN~\cite{wang2018esrgan} further improved upon SRGAN by refining the architecture with Residual-in-Residual Dense Blocks, adopting relativistic GANs for relative realness prediction, and refining perceptual loss, leading to more realistic and visually pleasing textures.

Wang~\etal~\cite{wang2018recovering} made significant changes to intermediate layers, focusing on recovering textures that align with semantic categories. Building on perceptual enhancements, Shang~\etal~\cite{Shang2020} introduced the receptive field block~\cite{liu2018receptive} to produce super-resolved images with extreme perceptual sharpness. Zhang~\etal~\cite{zhang2019ranksrgan} proposed RankSRGAN, a model that optimizes the generator using a ranker, allowing for improvements in perceptual metrics, particularly addressing the challenges posed by non-differentiable perceptual metrics.

In contrast, Zhou~\etal~\cite{zhou2019kernel} focused on the convolution kernel mismatch between training and real-world camera blur. They introduced kernel modeling SR, which integrates blur kernel modeling directly into the training process to generate more accurate super-resolved images. To prevent structural distortions in SR, Ma~\etal~\cite{ma2020structure} proposed a structure-preserving GAN-based approach that maintains perceptual quality while minimizing unwanted distortions.

Another important advancement came from Maeda~\cite{maeda2020unpaired}, who developed an unpaired SR method that eliminates the need for paired training data by leveraging GANs. Similarly, Umer~\cite{umer2020deep} drew inspiration from CycleGAN~\cite{zhu2017unpairediccv}, creating SRResCycGAN, which performs end-to-end translation from LR to HR domains without requiring paired data. Lee~\etal~\cite{lee2020journey} used Neural Architecture Search (NAS) to design more efficient generator and discriminator structures, optimizing performance automatically.

Recent work has focused on further improving restoration quality under challenging conditions. Chan~\etal~\cite{Chan2021_cvpr} introduced a latent bank built from pre-trained GANs, allowing for enhanced restoration even in extreme upscaling scenarios. Zhang~\etal~\cite{zhang2021designing} proposed a robust model capable of handling diverse image degradations, including random shuffle blur, downsampling, and noise. Finally, Luo~\etal~\cite{luo2022learning} tackled the uncertainty in degradation by modeling the degradation process as a probabilistic distribution, enabling the system to learn the mapping from a prior random variable to the degradation process.

\textbf{Introducing New Losses:} New loss functions have been introduced to enhance the performance of GAN-based SISR methods. Sajjadi~\etal~\cite{sajjadi2017enhancenet} proposed automated texture synthesis combined with perceptual loss to produce more realistic images. Here,
\begin{equation}
\mathcal{L}_{Tex.} = \lVert G(\varphi(X_{lr}))-G(\varphi(X_{hr}))\rVert_2^2,
\label{eq:tex}
\end{equation}
where the $\varphi(\cdot)$ represents the feature activation from a given VGG layer. The $G(\cdot)$ indicates the Gram matrix, and the result is calculated based on the patch-wise level. Vasu~\etal~\cite{Vasu2018} combined distortion loss, perceptual loss, and adversarial loss to achieve less distortion and better perceptual quality. Rad~\etal~\cite{Rad2019} explored the limits of perceptual loss and proposed a loss function that considered different semantic terms. It divided the object into three parts which are background, boundary, and object and calculated the loss respectively. Mathematically, it can be formulated as
\begin{equation}
  \begin{aligned}
    \mathcal{L}_{Tar.} &= \alpha\cdot \mathcal{G}_e(X_{sr}\circ M_{OBB}^{boundary},X_{hr}\circ M_{OBB}^{boundary}) \\
      &+ \beta\cdot \mathcal{G}_b(X_{sr}\circ M_{OBB}^{background},X_{hr}\circ M_{OBB}^{background}) \\
      &+ \gamma\cdot \mathcal{G}_o
  \end{aligned},
\label{eq:tar.}
\end{equation}
where $\alpha$, $\beta$, and $\gamma$ indicate the weight of corresponding loss terms, and $\mathcal{G}_e$, $\mathcal{G}_b$, and $\mathcal{G}_o$ represent the functions to calculate the space distances between two images. $M_{OBB}^{background}$, $M_{OBB}^{boundary}$ mean the mask of each class of interest. $\circ$ denotes the point-wise multiplication. Kim~\etal~\cite{Kim2020aaai} combined SR and video frame interpolation, proposing a training strategy with multi-scale temporal loss specific to the task. Fuoli~\etal~\cite{Fuoli2021} introduced a loss function in the Fourier domain to recover missing high-frequency content and proposed a Fourier domain discriminator to better match the high-frequency distribution. Park~\etal~\cite{park2023perception} designed a predict module to estimate the optimal combination of diverse losses and achieve superior perceptual quality. To better distinguish genuine high-frequency details from artifacts in image super-resolution, WGSR~\cite{korkmaz2024training} proposes training GAN-based models with wavelet-domain losses—where the discriminator operates solely on high-frequency wavelet sub-bands, and the generator is optimized via wavelet fidelity losses.

\textbf{Introducing Prior Properties:} Prior properties have been introduced in GAN-based SISR methods. Bulat~\etal~\cite{bulat2018learn} enhances real-world image and face super-resolution by employing a two-stage GAN framework that first learns to degrade high-resolution images with unpaired data, then trains a low-to-high GAN with paired data, achieving significant improvements over prior methods. cinCGAN~\cite{yuan2018unsupervised} addresses unsupervised single image super-resolution in challenging scenarios with noise and blur by using a Cycle-in-Cycle network structure that combines GAN-based noise removal, up-sampling, and end-to-end fine-tuning to achieve results comparable to state-of-the-art supervised models. KernelGAN~\cite{bell2019blind} is an image-specific Internal-GAN that trains solely on the low-resolution (LR) test image at test time, learning its internal patch distribution. Its generator is trained to downscale the LR image such that its discriminator cannot differentiate between the patch distributions of the downscaled and original LR images. This method determines the correct image-specific SR-kernel and enhances Blind-SR performance when integrated with existing SR algorithms. Soh~\etal~\cite{soh2019natural} focused on domain prior properties and defined naturalness prior in the low-level domain, constraining the output image to the natural manifold to ensure high perceptual quality. Shaham~\etal~\cite{Shaham2019} introduced an unconditional generative model to learn the patch distribution from a single natural image. Instead of adding details from the LR image, Menon~\etal~\cite{Menon2020} explored the HR image manifold and searched for images whose downscale version resembled the LR image. Pan~\etal~\cite{pan2021exploiting} leveraged the image prior captured by a GAN to restore missing semantics. Wang~\etal~\cite{Wang2021} proposed a model to find indistinguishable feature maps and map them to the HR domain. Wei~\etal~\cite{wei2021unsupervised} improves unsupervised super-resolution by addressing domain gaps between synthetic and real low-resolution images through domain-gap aware training and domain-distance weighted supervision, resulting in more realistic and natural SR outputs. SeD~\cite{li2024sed} employs a pretrained model to extract fine-grained semantic priors, enhancing the discriminator's capability through cross-attention mechanisms.
\begin{table*}[!t]
    \caption{\revision{Performance comparison of different SISR models on four benchmarks for the $\times 4$ task. PSNR/SSIM on the Y channel are reported for each dataset. The \textbf{\textcolor{blue}{best}} and \underline{\textcolor{blue}{second best}} performances are indicated for models trained on DIV2K, and the \textbf{\textcolor{red}{best}} and \underline{\textcolor{red}{second best}} performances are indicated for models trained on DF2K. The results for each model are obtained from their respective original papers.}}
    \centering
    \footnotesize
    \setlength{\tabcolsep}{4pt}{
    \begin{tabular}{c|c|cccc|c|c|c}
        \toprule[1.8pt]
        Model & \makecell[c]{Params\\(K)} & \makecell[c]{Set5\\(PSNR/SSIM)} & \makecell[c]{Set14\\(PSNR/SSIM)} & \makecell[c]{BSD100\\(PSNR/SSIM)} & \makecell[c]{Urban100\\(PSNR/SSIM)} & Training Sets & Backbone & Venue\\
        \midrule
        \hline

    SRCNN~\cite{Lim2017} & 24 & 30.48/0.8628 & 27.50/0.7513 & 26.90/0.7103 & 24.52/0.7226 & T91+ImageNet & CNN & ECCV'14\\

    \rowcolor{light-gray}SCN~\cite{wang2015deep} & - & 30.86/0.8732 & 27.64/0.7578 & 27.03/0.7161 & - & T91 & CNN & ICCV'15\\

    FSRCNN~\cite{fsrcnn2016} & 12 & 30.70/0.8657 & 27.59/0.7535 & 26.96/0.7128 & 24.60/0.7258 & T91+General-100 & CNN & ECCV'16\\
    
    \rowcolor{light-gray}ESPCN~\cite{Shi2016} & 25 & 30.52/0.8697 & 27.42/0.7606 & 26.87/0.7216 & 24.39/0.7241 & T91+ImageNet & CNN & CVPR'17\\
    
    VDSR~\cite{kim2016accurate} & 665 & 31.35/0.8838 & 28.01/0.7674 & 27.29/0.7251 & 25.18/0.7524 & BSD+T91 & CNN & CVPR'17\\
    \rowcolor{light-gray}DRRN~\cite{tai2017image} & 207 & 31.68/0.8888 & 28.21/0.7720 & 27.38/0.7284 & 25.44/0.7638 & BSD+T91 & CNN & CVPR'17\\
    LapSRN~\cite{lai2017deep} & 812 & 31.54/0.8850 & 28.19/0.7720 & 27.32/0.7270 & 25.21/0.7560 & BSD+T91 & CNN & CVPR'17\\
    \rowcolor{light-gray}SRDenseNet~\cite{tong2017image} & 5452 & 32.02/0.7819 & 28.50/0.7782 & 27.53/0.7337 & 26.05/0.7819 & ImageNet & CNN & ICCV'17\\
    SRGAN~\cite{Ledig2017} & - & - & 26.02/0.7379 & 25.16/0.6688 & - & ImageNet & GAN & CVPR'17\\
    \rowcolor{light-gray}IDN~\cite{Hui2018} & 678 & 31.82/0.8903 & 28.25/0.7730 & 27.41/0.7297 & 25.41/0.7632 & BSD+T91 & CNN & CVPR'18\\
    DSRN~\cite{han2018image} & 1200 & 31.40/0.8830 & 28.07/0.7700 & 27.25/0.7240 & 25.08/0.7470 & T91 & CNN & CVPR'18\\
    \rowcolor{light-gray}ZSSR~\cite{shocher2018zero} & 225 & 31.13/0.8796 & 28.01/0.7651 & 27.12/0.7211 & - & - & CNN & CVPR'18\\
    ESRGAN~\cite{wang2018esrgan} & - & 30.44/0.8520 & - & 25.29/0.6500 & 24.37/0.7340 & DF2K+OST~\cite{wang2018recovering} & GAN & CVPRW'18\\
    
    \hline
    \hline
    
    \rowcolor{light-gray}EDSR~\cite{Lim2017} & 43000 & 32.46/0.8968 & 28.80/0.7876 & 27.71/0.7420 & 26.64/0.8033 & DIV2K & CNN & CVPRW'17\\
    RDN~\cite{Zhang2018} & 21900 & 32.47/0.8990 & 28.81/0.7871 & 27.72/0.7419 & 26.61/0.8028 & DIV2K & CNN & CVPR'18\\
    \rowcolor{light-gray}MSRN~\cite{li2018multi} & 6300 & 32.07/0.8903 & 28.60/0.7751 & 27.52/0.7273 & 26.04/0.7896 & DIV2K & CNN & ECCV'18\\
    RCAN~\cite{zhang2018image} & 16000 & 32.63/0.9002 & 28.87/0.7889 & 27.77/0.7436 & 26.82/0.8087 & DIV2K & CNN & ECCV'18\\
      \rowcolor{light-gray}SRFeat~\cite{park2018srfeat} & 6189 & 32.27/0.8938 & 28.71/0.7835 & 27.64/0.7378 & - & DIV2K & GAN & ECCV'18\\
      RDN-MetaSR~\cite{hu2019meta} & - & 32.38/- & 28.78/- & 27.71/- & 26.55/- & DIV2K & CNN & CVPR'19\\
      \rowcolor{light-gray}OISR~\cite{he2019ode} & 15592 & 32.53/0.8992 & 28.86/0.7878 & 27.75/0.7428 & 26.79/0.8068 & DIV2K & CNN & CVPR'19\\
      SAN~\cite{dai2019second} & 15700 & 32.64/0.9003 & 28.92/0.7888 & 27.78/0.7436 & 26.79/0.8068 & DIV2K & CNN & CVPR'19\\
      \rowcolor{light-gray}EBRN~\cite{qiu2019embedded} & 7900 & \textbf{\textcolor{blue}{32.79}}/\textbf{\textcolor{blue}{0.9032}} & \textbf{\textcolor{blue}{29.01}}/0.7903 & \textbf{\textcolor{blue}{27.85}}/\textbf{\textcolor{blue}{0.7464}} & 27.03/0.8114 & DIV2K & CNN & ICCV'19\\
      NatSR~\cite{soh2019natural} & 4800 & 30.98/0.8606 & 27.42/0.7329 & 26.44/0.6827 & 25.46/0.7602 & DIV2K & GAN & CVPR'19\\
      \rowcolor{light-gray}LatticeNet~\cite{luo2020latticenet} & 777 & 32.30/0.8962 & 28.68/0.7830 & 27.62/0.7367 & 26.25/0.7873 & DIV2K & CNN & ECCV'20\\
      HAN~\cite{niu2020single} & 64199 & 32.64/0.9002 & 28.90/0.7890 & 27.80/0.7442 & 26.85/0.8094 & DIV2K & CNN & ECCV'20\\
      \rowcolor{light-gray}RFANet~\cite{liu2020residual} & 11000 & 32.66/0.9004 & 28.88/0.7894 & 27.79/0.7442 & 26.92/0.8112 & DIV2K & CNN & CVPR'20\\
      CSNLN~\cite{mei2020image} & 30000 & 32.68/0.9004 & 28.95/0.7888 & 27.80/0.7439 & \underline{\textcolor{blue}{27.12}}/\textbf{\textcolor{blue}{0.8168}} & DIV2K & CNN & CVPR'20\\ 
      \rowcolor{light-gray}LAPAR~\cite{Li2020} & - & 32.57/0.8998 & \underline{\textcolor{blue}{28.96}}/0.7908 & \underline{\textcolor{blue}{27.84}}/0.7447 & 27.04/0.8128 & DIV2K & CNN & NeurIPS'20\\
      TPSR-D2~\cite{lee2020journey} & 610 & 29.60/- & 26.88/- & 26.23/- & 24.12/- & DIV2K & GAN & ECCV'20\\  
      \rowcolor{light-gray} AdderSR~\cite{song2021addersr} & 43000 & 32.13/0.8864 & 28.57/0.7800 & 27.58/0.7368 & 26.33/0.7874 & DIV2K & CNN & CVPR'21\\
      
      LUT~\cite{jo2021practical} & 1274 & 29.82/0.8478 & 27.01/0.7355 & 26.53/0.6953 & 24.02/0.6990 & DIV2K & CNN & CVPR'21\\
      
      \rowcolor{light-gray} SMSR~\cite{wang2021exploring} & 1006 & 32.12/0.8932 & 28.55/0.7808 & 27.55/0.7351 & 26.11/0.7868 & DIV2K & CNN & CVPR'21\\
      
      RDN-LIIF~\cite{chen2021learning} & - & 32.50/- & 28.80/- & 27.74/- & 26.68/- & DIV2K & CNN & CVPR'21\\

      \rowcolor{light-gray} NLSN~\cite{mei2021image} & - & 32.59/0.9000 & 28.87/0.7891 & 27.78/0.7444 & 26.96/0.8109 & DIV2K & CNN & CVPR'21\\
      
      NAPS~\cite{zhan2021achieving} & 125 & 31.93/0.8906 & 28.42/0.7763 & 27.44/0.7307 & 25.66/0.7715 & DIV2K & CNN & ICCV'21\\
      
      \rowcolor{light-gray} FAD-RCAN~\cite{xie2021learning} & - & 32.65/0.9007 & 28.88/0.7889 & 27.78/0.7437 & 26.86/0.8092 & DIV2K & CNN & ICCV'21\\
      
      ArbRCAN~\cite{wang2021learning} & 16600 & 32.55/- & 28.87/- & 27.76/- & 26.68/- & DIV2K & CNN & ICCV'21\\

      \rowcolor{light-gray}SPSR~\cite{ma2021structure} & - & 31.04/0.8772 & 27.07/\textcolor{blue}{\textbf{0.8076}} & 26.05/0.6818 & 25.23.0.9531 & DIV2K & GAN & TPAMI21\\
      
      EDSR-SLS~\cite{oh2022attentive} & 363 & - & 28.49/- & 27.51/- & 25.84/- & DIV2K & CNN & CVPR'22\\

      \rowcolor{light-gray}SPLUT~\cite{ma2022learning} & 18000 & 30.52/0.8630 & 27.54/0.7520 & 26.87/0.7090 & 24.46/0.7190 & DIV2K & CNN & ECCV'22\\

       RCLUT~\cite{liu2023reconstructed} & 1513 & 30.72/0.8677 & 27.67/0.7577 & 26.95/0.7145 & 24.57/0.7253 & DIV2K & CNN & ICCV'23\\
      
      \rowcolor{light-gray}ELAN~\cite{zhang2022efficient} & 8312 & \textcolor{blue}{\underline{32.75}}/\textcolor{blue}{\underline{0.9022}} & \underline{\textcolor{blue}{28.96}}/\underline{\textcolor{blue}{0.7914}} & 27.83/\underline{\textcolor{blue}{0.7459}} & \textbf{\textcolor{blue}{27.13}}/\underline{\textcolor{blue}{0.8167}} & DIV2K & Transformer & ECCV'22\\
      
      ESRT~\cite{lu2022transformer} & 751 & 32.19/0.8947 & 28.69/0.7833 & 27.69/0.7379 & 26.39/0.7962 & DIV2K & Transformer & CVPRW'22\\
      
      \rowcolor{light-gray}Omni-SR~\cite{wang2023omni} & 792 & 32.49/0.8988 & 28.78/0.7859 & 27.71/0.7415 & 26.64/0.8018 & DIV2K & Transformer & CVPR'23\\

      CRAFT~\cite{li2023feature} & 753 & 32.52/0.8989 & 28.85/0.7872 & 27.72/0.7418 & 26.56/0.7995 & DIV2K & Transformer & ICCV'23\\

      \rowcolor{light-gray}DLGSANet~\cite{li2023dlgsanet} & 761 & 32.54/0.8993 & 28.84/0.7871 & 27.73/0.7415 & 26.66/0.8033 & DIV2K & Transformer & ICCV'23\\

      SRFormer~\cite{zhou2023srformer} & 873 & 32.51/0.8988 & 28.82/0.7872 & 27.73/0.7422 & 26.67/0.8032 & DIV2K & Transformer & ICCV'23\\
      
      \rowcolor{light-gray}CAL-GAN~\cite{park2023content} & - & 31.18/0.8630 & - & 25.93/0.6760 & 25.29/0.7630 & DIV2K & GAN & ICCV'23\\
      
      WGSR~\cite{korkmaz2024training} & - & 31.51/0.8690 & 26.69/0.7160 & 26.37/0.6840 & 25.61/0.7770 & DIV2K & GAN & CVPR'24\\

        \hline
        \hline
       DBPN~\cite{Haris2018} & 10000 & 32.47/0.8980 & 28.82/0.7860 & 27.72/0.7400 & 26.38/0.7946 & DF2K & CNN & CVPR'18\\
      
      \rowcolor{light-gray}SRCliqueNet~\cite{Zhong2019} & 659 & 32.15/0.8944 & 28.61/0.7818 & 27.61/0.7366 & 26.14/0.7871 & DF2K & CNN & NeurIPS'19\\
      
       SRFBN~\cite{li2019feedback} & 3500 & 32.46/0.8968 & 28.80/0.7876 & 27.71/0.7420 & 26.64/0.8033 & DF2K & CNN & CVPR'19\\

      \rowcolor{light-gray}DRN~\cite{guo2020closed} & 9800 & 32.74/0.9020 & 28.98/0.7920 & 27.83/0.7450 & 27.03/0.8130 & DF2K & CNN & CVPR'20\\

      USRGAN~\cite{zhang2020deep} & - & 30.91/0.8660 & - & 25.97/0.6760 & 24.89/0.7500 & DF2K & GAN & CVPR20\\
      
      \rowcolor{light-gray}DFSA~\cite{magid2021dynamic} & - & 32.79/0.9019 & 29.06/0.7922 & 27.87/0.7458 & 27.17/0.8163 & DF2K & CNN & ICCV'21\\
      
      CRAN~\cite{zhang2021context} & 19940 & 32.72/0.9012 & 29.01/0.7918 & 27.86/0.7460 & 27.13/0.8167 & DF2K & CNN & ICCV'21\\

      \rowcolor{light-gray}ASSLN~\cite{Zhang2021} & 677 & 32.29/0.8964 & 28.69/0.7844 & 27.66/0.7384 & 26.27/0.7907 & DF2K & CNN & NeurIPS'21\\
      
      DCLS~\cite{luo2022deep} & - & 32.12/0.8890 & 28.54/0.7728 & 27.60/0.7285 & 26.15/0.7809 & DF2K & CNN & CVPR'22\\

      \rowcolor{light-gray}SwinIR~\cite{liang2021swinir} & 11900 & 32.92/0.9044 & 29.09/0.7950 & 27.92/0.7489 & 27.45/0.8254 & DF2K & Transformer & ICCVW'21\\
      
      CAT~\cite{zheng2022cross} & 16600 & 33.08/0.9052 & 29.18/0.7960 & 27.99/0.7510 & 27.89/0.8339 & DF2K & Transformer & NeurIPS'22\\

      \rowcolor{light-gray}SwinIR+LDL~\cite{liang2022details} & - & 31.03/0.8611 & 27.53/0.7478 & - & 26.23/0.7918 & DF2K & GAN & CVPR22\\
      
      HAT~\cite{chen2022activating} & 20800 & 33.04/\underline{\textcolor{red}{0.9056}} & \underline{\textcolor{red}{29.23}}/\underline{\textcolor{red}{0.7973}} & \underline{\textcolor{red}{28.00}}/\underline{\textcolor{red}{0.7517}} & \underline{\textcolor{red}{27.97}}/\underline{\textcolor{red}{0.8368}} & DF2K & Transformer & CVPR'23\\ 
      \rowcolor{light-gray}EQSR~\cite{wang2023deep} & - & 32.71/- & 29.12/- & 27.86/- & 27.30/- & DF2K & Transformer & CVPR'23\\
      
      DAT~\cite{chen2023dual} & 11212 & \underline{\textcolor{red}{33.08}}/0.9055 & \underline{\textcolor{red}{29.23}}/\underline{\textcolor{red}{0.7973}} & \underline{\textcolor{red}{28.00}}/0.7515 & 27.87/0.8343 & DF2K & Transformer & ICCV'23\\
      
      \rowcolor{light-gray}SAFMN~\cite{sun2023spatially} & 240 & 32.18/0.8948 & 28.60/0.7813 & 27.58/0.7359 & 25.97/0.7809 & DF2K & Transformer & ICCV'23\\
      CFAT~\cite{ray2024cfat} & 22070 &  \textbf{\textcolor{red}{33.19}}/ \textbf{\textcolor{red}{0.9068}} & \textbf{\textcolor{red}{29.30}}/\textbf{\textcolor{red}{0.7985}} & \textbf{\textcolor{red}{28.17}}/\textbf{\textcolor{red}{0.7524}} & \textbf{\textcolor{red}{28.11}}/\textbf{\textcolor{red}{0.8380}} & DF2K & Transformer & CVPR'24\\
      
      \hline
      \bottomrule[1.5pt]
    \end{tabular}}
    \label{tab:sisr_results}
  \end{table*}
\subsubsection{\textbf{Diffusion-based SISR}}
The inherent instability of generative adversarial networks (GANs) in super-resolution tasks - particularly their tendency to produce artifacts and training difficulties - has prompted a paradigm shift toward diffusion models. These models offer distinct advantages in training stability and generation quality, leading to two principal research directions: retraining mechanisms and diffusion priors utilization.

\textbf{Retraining Mechanisms:}
The evolution of diffusion-based super-resolution began with fundamental architectural innovations. SR3~\cite{saharia2022image} established the baseline approach by combining denoising diffusion probabilistic models with U-Net architectures, demonstrating superior performance over GANs through iterative multi-level refinement. Subsequent developments pursued three key improvements: (1) training efficiency, (2) computational optimization, and (3) architectural scalability. SR3+~\cite{sahak2023denoising} enhanced training robustness through noise-conditioning augmentation, while DiffIR~\cite{xia2023diffir} achieved computational efficiency via compact prior extraction. Resshift~\cite{yue2023resshift} further optimized the inference process through residual-based Markov chains, reducing required sampling steps to just 20. Parallel advancements in latent space operations emerged through LDM~\cite{rombach2022high}, which leveraged pre-trained autoencoders~\cite{esser2021taming} for efficient high-resolution synthesis. The field subsequently progressed to multi-scale architectures, with CDM~\cite{ho2022cascaded} employing cascaded models for error-resistant generation and IDM~\cite{gao2023implicit} introducing implicit representations for continuous-scale adaptation.

\textbf{Diffusion Priors:}
The second major direction exploits pre-trained diffusion models as powerful priors for super-resolution, addressing three core challenges: blind restoration, semantic consistency, and real-world degradation handling. StableSR~\cite{wang2024exploiting} pioneered this approach through adaptive knowledge transfer from text-to-image models, establishing key components like time-aware encoders. Subsequent works expanded these foundations along two dimensions: task generalization and semantic integration. DiffBIR~\cite{lin2024diffbir} developed a unified restoration pipeline with region-adaptive control, while PASD~\cite{yang2024pixel} demonstrated joint super-resolution and stylization capabilities. The most recent innovations focus on semantic-aware processing: SUPIR~\cite{yu2024scaling} introduced prompt-guided restoration, and SeeSR~\cite{wu2024seesr} developed degradation-aware semantic prompting, effectively bridging low-level reconstruction with high-level image understanding. These approaches collectively demonstrate how diffusion priors enable unified frameworks that surpass traditional methods in both perceptual quality and task versatility.
\begin{table}[!t]
    \caption{\resubmit{Performance comparison of regression-based and generation-based models for the $\times$4 task, with the best results for each category highlighted in \textcolor{red}{\textbf{red}} (regression) and \textcolor{blue}{\textbf{blue}} (generation).}}
    \centering
    \footnotesize
    \setlength{\tabcolsep}{4pt}{
    \begin{tabular}{c|cccc|c}
        \toprule[1.8pt]
        Model & PSNR$\uparrow$ & SSIM$\uparrow$ & LPIPS$\downarrow$ & FID$\downarrow$  & Venue\\
        \midrule[1pt]
        StableSR~\cite{wang2024exploiting} & 19.11 & 0.559 & 0.171 & 54.77 & IJCV'24\\
        BSRGAN~\cite{zhang2021designing} & 21.04 & 0.616 & 0.202 & 71.99 & ICCV'21\\
        Real-ESRGAN~\cite{wang2021real} & 20.46 & 0.612 & 0.190 & 66.45 & ICCVW'21\\
         SeeSR~\cite{wu2024seesr} & 20.55 & 0.596 & 0.168 & 51.38 & CVPR'24\\
         SUPIR~\cite{yu2024scaling} & 18.08 & 0.512 & 0.229 & 70.06 & CVPR'24\\
         RRDB+LDL~\cite{liang2022details} & \textbf{\textcolor{blue}{23.32}} & \textbf{\textcolor{blue}{0.717}} & \textbf{\textcolor{blue}{0.098}} & \textbf{\textcolor{blue}{23.05}} & CVPR'22\\
         \midrule[1pt]
         EDSR~\cite{Lim2017} & 23.90 & 0.737 & 0.204 & 46.60 & CVPR'17\\
         DRN~\cite{guo2020closed} & 24.80 & 0.768 & 0.171 & 40.14 & CVPR'20\\
         SwinIR~\cite{liang2021swinir} & 24.34 & 0.754 & 0.191 & 44.56 & ICCVW'21\\
         CAT~\cite{zheng2022cross} & 25.38 & 0.791 & 0.154 & 37.31 & NeurIPS'22\\
         SRFormer~\cite{zhou2023srformer} & 24.49 & 0.759 & 0.186 & 43.87 & ICCV'23\\
         CRAFT~\cite{li2023feature} & 24.41 & 0.757 & 0.186 & 43.96 & ICCV'23\\
         HAT~\cite{chen2022activating} & \textbf{\textcolor{red}{26.46}} & \textbf{\textcolor{red}{0.817}} & \textbf{\textcolor{red}{0.134}} & \textbf{\textcolor{red}{32.83}} & CVPR'23\\
      \bottomrule[1.5pt]
    \end{tabular}}
    \label{tab:sisr_generative_results}
  \end{table}
\begin{figure}[!t]
    \centering
    \includegraphics[width=\linewidth]{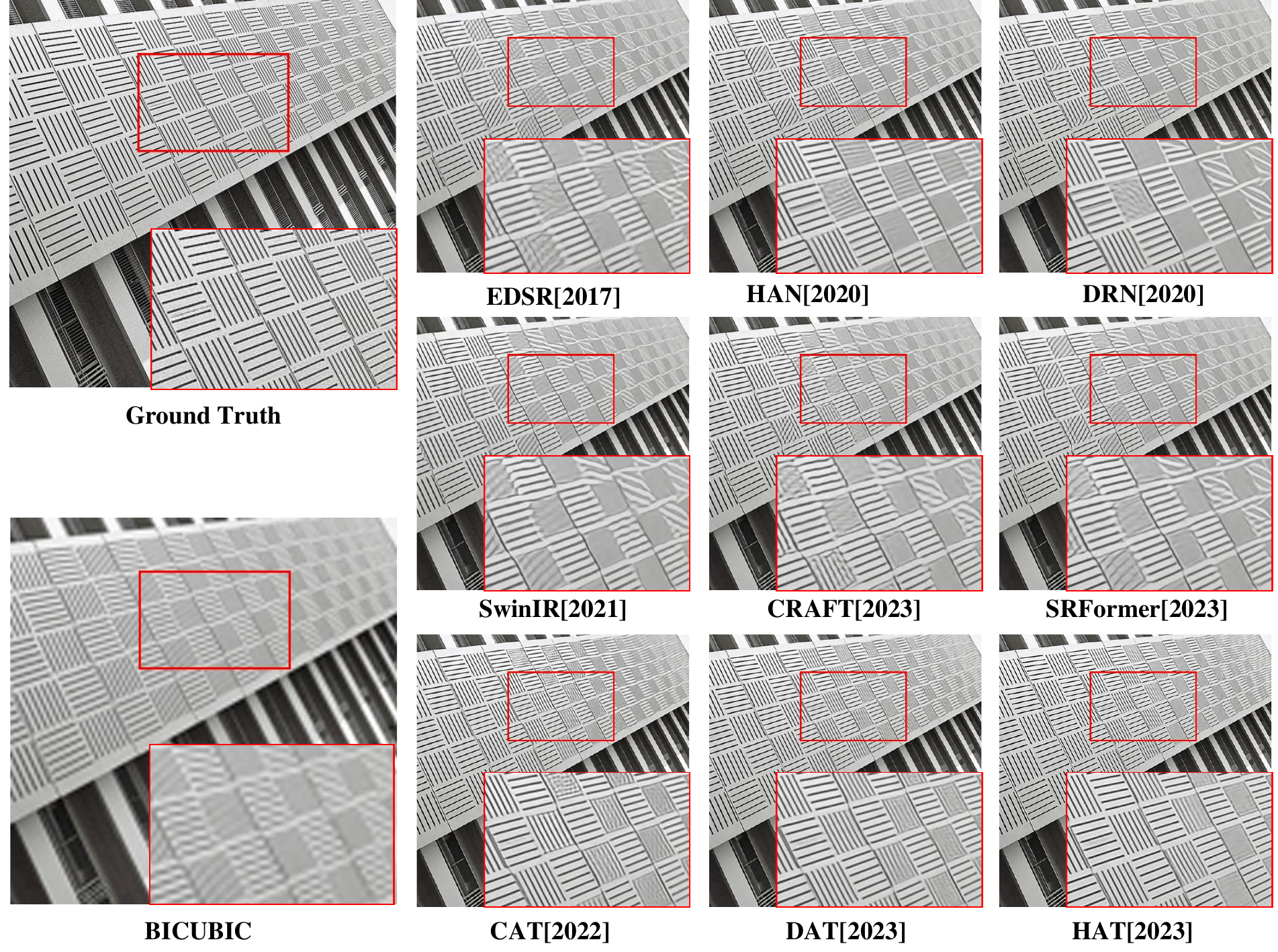}
    \caption{\resubmit{Visual comparison of regression-based super-resolution methods: CNN-based (top), lightweight Transformers (middle), and standard Transformers (bottom). Standard Transformers achieve the best reconstruction quality}}
    \label{fig:sisr_reg_vis}
\end{figure}
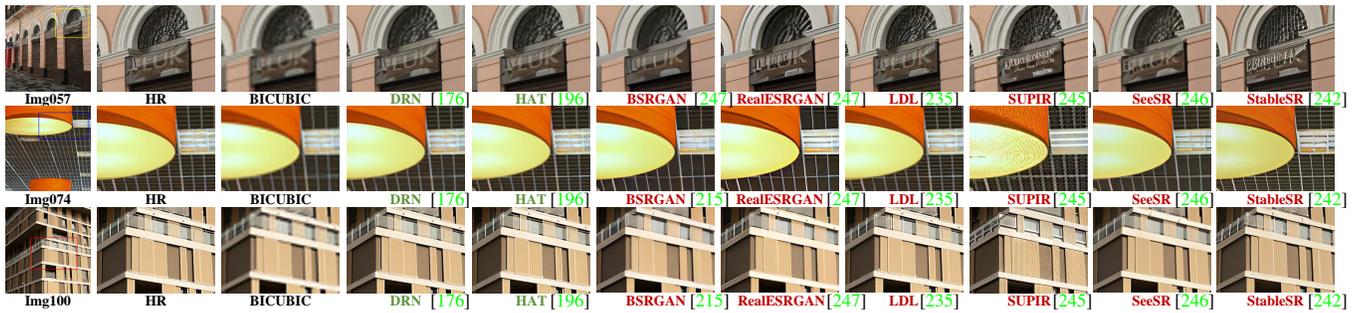
\begin{figure*}[!t]
    \centering
    \begin{overpic}[scale=.18]{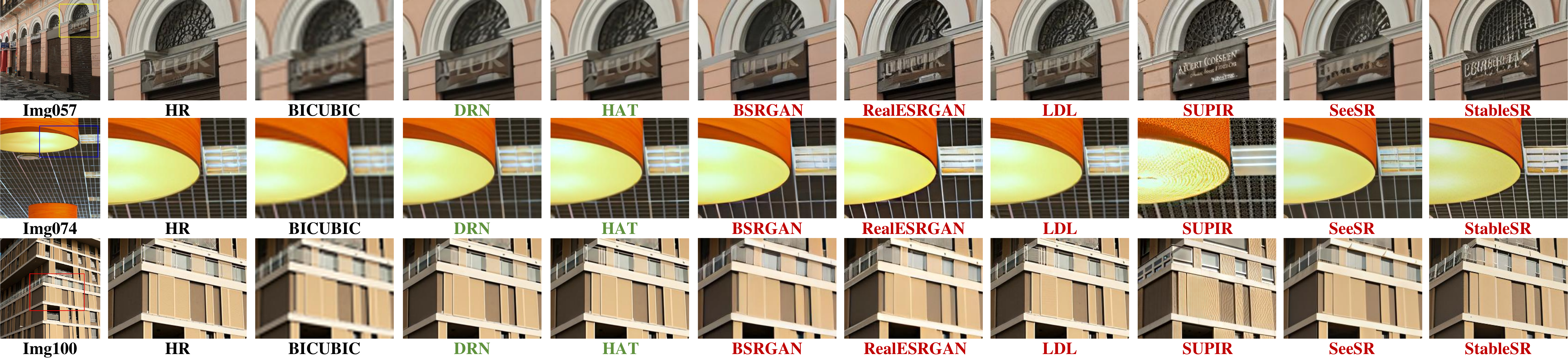}
        \put(31, 0.3){\scriptsize~~\cite{guo2020closed}}
        \put(31, 8){\scriptsize~~\cite{guo2020closed}}
        \put(31, 15.5){\scriptsize~~\cite{guo2020closed}}
        \put(40, 0.3){\scriptsize~~\cite{chen2022activating}}
        \put(40, 8){\scriptsize~~\cite{chen2022activating}}
        \put(40, 15.5){\scriptsize~~\cite{chen2022activating}}
        \put(50.5, 0.3){\scriptsize~~\cite{zhang2021designing}}
        \put(50.5, 8){\scriptsize~~\cite{zhang2021designing}}
        \put(50.8, 15.5){\scriptsize~~\cite{wang2021real}}
        \put(60.8, 0.3){\scriptsize~~\cite{wang2021real}}
        \put(60.8, 8){\scriptsize~~\cite{wang2021real}}
        \put(60.8, 15.5){\scriptsize~~\cite{wang2021real}}
        \put(67.8, 15.5){\scriptsize~~\cite{liang2022details}}
        \put(67.8, 0.3){\scriptsize~~\cite{liang2022details}}
        \put(67.8, 8){\scriptsize~~\cite{liang2022details}}
        \put(77.8, 15.5){\scriptsize~~\cite{yu2024scaling}}
        \put(77.8, 0.3){\scriptsize~~\cite{yu2024scaling}}
        \put(77.8, 8){\scriptsize~~\cite{yu2024scaling}}
        \put(87, 15.5){\scriptsize~~\cite{wu2024seesr}}
        \put(87, 0.3){\scriptsize~~\cite{wu2024seesr}}
        \put(87, 8){\scriptsize~~\cite{wu2024seesr}}
        \put(97, 15.5){\scriptsize~~\cite{wang2024exploiting}}
        \put(97, 0.3){\scriptsize~~\cite{wang2024exploiting}}
        \put(97, 8){\scriptsize~~\cite{wang2024exploiting}}
    \end{overpic}
    \caption{\resubmit{Visual comparison between \textcolor{regressive}{\textbf{regression-based}} and \textcolor{generative}{\textbf{generation-based}} super-resolution methods on sample images. Regression methods include CNN and Transformer approaches, while generation methods comprise GAN and diffusion-based techniques. While \textcolor{generative}{\textbf{generation-based}} methods can generate clearer and more detailed context, they may easily compromise fidelity. On the contrary, \textcolor{regressive}{\textbf{regression-based}} methods can produce more reliable output but are often blurrier. Choosing which type of method depends on the purpose.}}
    \label{fig:sisr_reg_gen_vis}
\end{figure*}
\subsection{\textbf{Training Details}}
\revision{\subsubsection{\textbf{Datasets}}
\
\newline
\indent There are five commonly used datasets for evaluating the performance of SISR methods: Set5~\cite{bevilacqua2012low}, Set14~\cite{zeyde2010single}, Urban100~\cite{huang2015single}, BSD100~\cite{martin2001database}, and Manga109~\cite{matsui2017sketch}. These datasets, widely adopted by the research community, serve as standard benchmarks in the field of super-resolution. The training and evaluation protocols referenced in this work follow the common practices outlined in previous literature, and the results discussed are based on established strategies rather than new, independent training experiments.}

\begin{itemize}
\item Set5~\cite{bevilacqua2012low}: This dataset consists of five classic images, including baby, bird, butterfly, head, and woman.
\item Set14~\cite{zeyde2010single}: Compared to Set5~\cite{bevilacqua2012low}, this dataset contains a wider range of categories. It includes 14 images, including both natural and manual images.
\item Urban100~\cite{huang2015single}: This dataset comprises 100 images that feature various urban buildings. It includes challenging city landscapes with different frequency details.
\item BSD100~\cite{martin2001database}: BSD100 consists of 100 images, including natural images as well as object-specific images such as plants, people, and food.
\item Manga109~\cite{matsui2017sketch}: Manga109 is a dataset containing 21,142 manga images drawn by 94 professional manga artists. The average size of the images is 833 $\times$ 1179 pixels. For SR, 109 images were selected from the dataset for testing purposes.
\end{itemize}

Regarding the training set, the main dataset used varies over time. In the early stages, methods used images extracted from ImageNet~\cite{deng2009imagenet}, T91~\cite{timofte2014a}, and General100~\cite{dong2016accelerating} for training the models. After the release of DIV2K~\cite{agustsson2017ntire}, almost all mainstream methods began using this dataset for training. To further improve model performance, Flicker2K~\cite{timofte2017ntire} was also included in the training process.
\begin{itemize}
\item ImageNet~\cite{deng2009imagenet}: A large image dataset with over a thousand categories, commonly used for various computer vision tasks.
\item T91~\cite{timofte2014a}: T91 consists of 91 common natural images.
\item General100~\cite{dong2016accelerating}: This dataset contains 100 HD lossless images.
\item DIV2K~\cite{agustsson2017ntire}: The NTIRE 2017 competition dataset, comprising 1000 HD images with a resolution of 2K. Among them, 800 are used for training, and 100 for validation, covering diverse scenes, categories, and frequency details.
\item Flicker2k~\cite{timofte2017ntire}: Another commonly used dataset, often combined with DIV2K to form DF2K, containing 2650 images with a resolution of 2K.
\end{itemize}

\subsubsection{\textbf{Training Settings}}
\
\newline
\indent\textbf{Patch-based Training:} Given resource limitations and computational efficiency, training patches are extracted from the available training samples to serve as input-output pairs during the training process. The size of LR patches typically ranges from $16\times 16$ to $92\times 92$, depending on the model architecture and available resources.

\textbf{Data Augmentation:} To enhance the diversity of training samples and improve model generalization, data augmentation techniques, such as random flipping, rotation, and color jittering, are applied to the extracted patches.

\textbf{Optimization Methods:} Stochastic Gradient Descent (SGD) with momentum or the Adam optimizer is commonly used for training deep learning models. The hyperparameters, including learning rate, weight decay, and momentum, require careful tuning. Typically, the learning rate is set between $10^{-3}$ and $10^{-4}$, depending on the network architecture and the specific method used.

\textbf{Learning Rate Schedule:} To facilitate efficient training, a learning rate schedule is employed to adjust the learning rate during the training process. \revision{Common schedules include step decay, learning rate warm-up~\cite{loshchilov2017sgdr}, or cyclic learning rate~\cite{smith2017cyclical}.}

\textbf{Batch Size:} The batch size determines the number of samples processed together during each iteration of the training process. In light of memory constraints and computational complexity, batch sizes are commonly set between $16$ and $32$, striking a balance between training efficiency and memory consumption.

\subsection{\textbf{Experimental Results}}
\label{sisr_results}
\resubmit{We comprehensively evaluate the performance of regression-based and generative SISR methods through both objective metrics and visual comparisons.}

\resubmit{\textbf{Quantitative Analysis.} Table~\ref{tab:sisr_results} summarizes the PSNR and SSIM scores of various methods on synthetic test datasets (Set5~\cite{bevilacqua2012low}, Set14~\cite{zeyde2010single}, BSD100~\cite{martin2001database}, Urban100~\cite{huang2015single}), alongside model complexity (number of parameters), training sets, architectures, venues, and publication years. The results are sourced from the respective papers. We also split the comparison within the same training set to ensure fair evaluation.

Due to the synthetic nature of these datasets, generative models are underrepresented; thus we include an additional Table~\ref{tab:sisr_generative_results} to explicitly compare regression and generative methods. Specifically, we randomly crop 512$\times$512 HR patches from Urban100 (100 samples total) and generate corresponding LR images using bicubic $\times$4 downsampling. We evaluate six generation-based methods~\cite{wang2024exploiting,zhang2021designing,wang2021real,wu2024seesr,yu2024scaling,liang2022details} and seven regression-based methods~\cite{Lim2017,liang2021swinir,zheng2022cross,zhou2023srformer,li2023feature,chen2022activating,guo2020closed} using both pixel-wise (PSNR, SSIM) and perceptual metrics (LPIPS, FID).

From the tables, several key observations emerge:~\circled{1} First, deep learning-based techniques consistently outperform the bicubic interpolation method, with a notable margin of at least 2dB in terms of PSNR. This highlights the effectiveness of deep learning in enhancing image resolution and generating higher-quality results.~\circled{2} The use of larger training datasets, such as DIV2K and Flickr2K, has played a significant role in improving the performance of SISR systems. The availability of more extensive and diverse datasets has allowed models to learn a wider range of image features and textures, resulting in better generalization and enhanced performance.~\circled{3} Analyzing the timeline of SISR developments, there has been a noticeable shift from CNN-based architectures to transformer-based solutions. Transformer architectures have gradually become dominant in SISR tasks due to their ability to effectively capture long-range dependencies. Specifically, they calculate attention scores based on the entire feature map, using dimensions $(HW)\times C$, where $H$ and $W$ represent the height and width of the feature map, and $C$ denotes the number of channels. However, it is important to note that achieving better performance often requires larger model parameters and increased computational resources (\eg, $\mathcal{O}((HW)^2C)$, where $(HW) \gg C$). Balancing the tradeoff between performance and complexity remains an ongoing challenge in SISR research.~\circled{4} Generation-based methods exhibit lower PSNR/SSIM but can produce visually clearer results, though with occasional fidelity loss. Interestingly, they don't always achieve better perceptual metrics (LPIPS/FID), likely due to hallucination effects visible in the super-resolved outputs.}

\resubmit{\textbf{Qualitative Analysis.} Visual comparisons (Fig.~\ref{fig:sisr_reg_vis} and Fig.~\ref{fig:sisr_reg_gen_vis}) further demonstrate these trends.

From Fig.~\ref{fig:sisr_reg_vis}, the visual comparison demonstrates the progressive improvements in regression-based super-resolution methods across different architectural paradigms. The first row (CNN-based methods: EDSR~\cite{Lim2017}, HAN~\cite{niu2020single}, DRN~\cite{guo2020closed}) shows competent reconstruction but with visible limitations in fine detail recovery and occasional blurring of complex textures. The second row (lightweight Transformers: SwinIR~\cite{liang2021swinir}, CRAFT~\cite{li2023feature}, SRFormer~\cite{zhou2023srformer}) exhibits noticeable improvements in texture preservation and edge sharpness, benefiting from more efficient attention mechanisms. The bottom row (standard Transformers: CAT~\cite{zheng2022cross}, DAT~\cite{chen2023dual}, HAT~\cite{chen2022activating}) achieves the best reconstruction quality, particularly in: (1) faithfully recovering high-frequency details and (2) producing the most natural-looking results with minimal artifacts. All methods significantly outperform the bicubic baseline, with standard Transformers demonstrating superior capability in handling challenging patterns while maintaining computational efficiency. The progression from CNNs to lightweight and then standard Transformers illustrates how architectural advances have addressed traditional super-resolution challenges, though with increasing model complexity.

From Fig.~\ref{fig:sisr_reg_gen_vis}, the comparison demonstrates distinct characteristics between regression-based (DRN~\cite{guo2020closed}, HAT~\cite{chen2022activating}) and generation-based (LDL~\cite{liang2022details}, SUPIR~\cite{yu2024scaling}, SeaSR~\cite{wu2024seesr}, StablaSR~\cite{wang2024exploiting}) approaches. Regression methods show more conservative reconstructions that maintain better fidelity to original structures, particularly in preserving geometric patterns. Generation-based methods produce visually sharper outputs but exhibit more variations in texture synthesis and occasional artifacts in complex areas. For example, Img057 shows generative methods (especially diffusion models) generating wrong characters, while Img074 reveals color shift problems in generative outputs.

Overall, CNN/Transformer regression methods appear more consistent across different image samples, while GAN/diffusion approaches show greater variability in their enhancement styles. The results validate Transformer-based approaches' effectiveness in capturing long-range dependencies for image restoration tasks.}

\revision{\section{Video Super-resolution}
\label{sec:vsr}
\subsection{\textbf{Regressive Models}}
\subsubsection{\textbf{CNN-based VSR}}
\
\newline
\textbf{Efficient Structural Designs:} \revision{Several studies have focused on lightweight network designs to reduce model complexity, as illustrated in Fig.~\ref{fig:vsr_efficient}. These approaches can be categorized into two primary types: recurrent modeling and convolution modeling.}
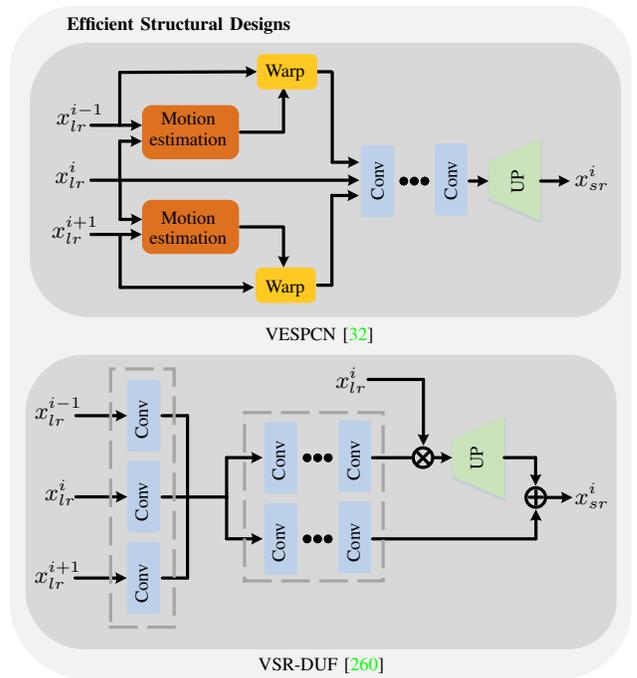
\begin{figure}[!th]
    \centering
    \begin{overpic}[scale=.4]{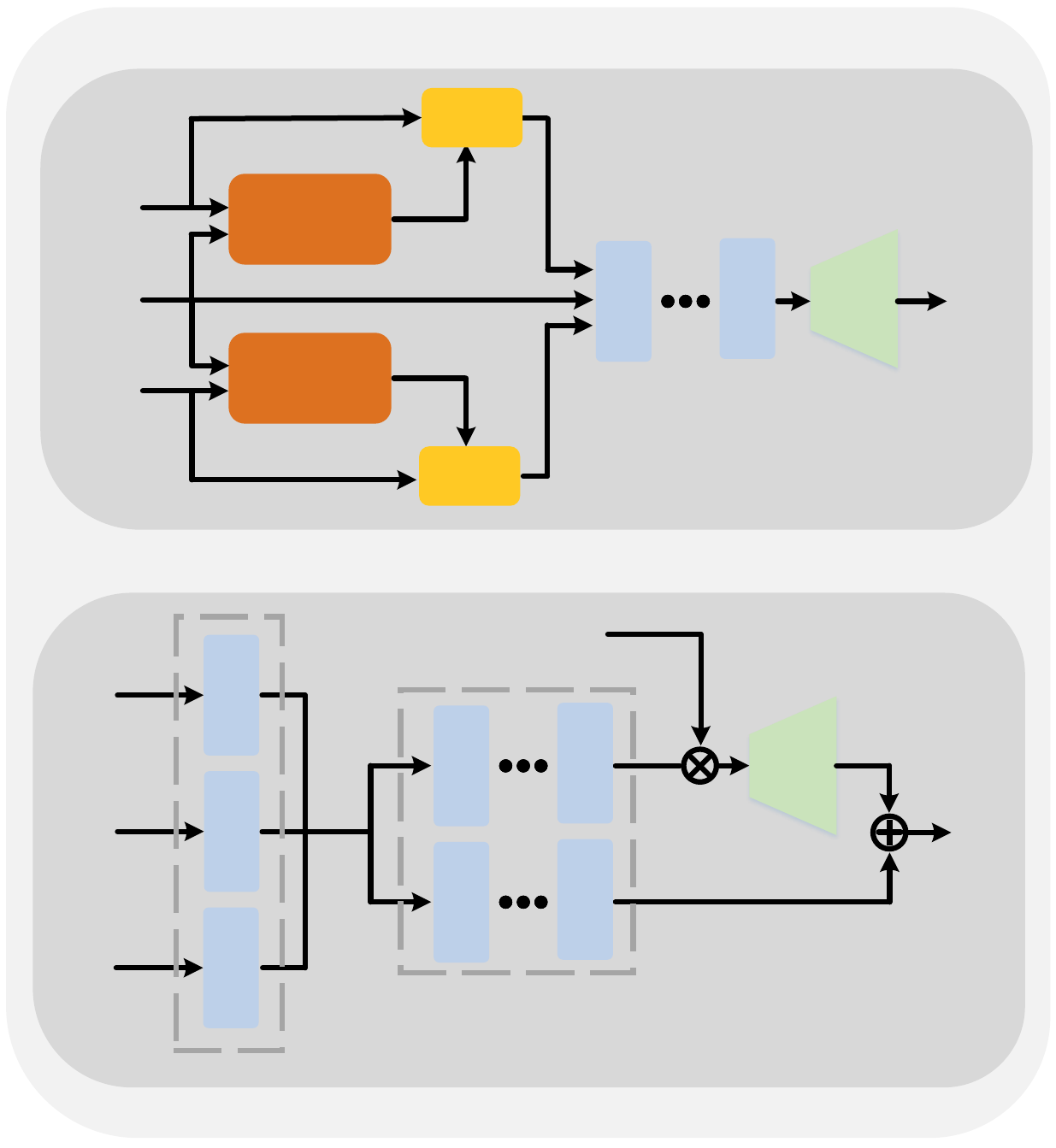}
        \put(8, 95.8){\scriptsize~\textbf{Efficient Structural Designs}}
        \put(37, 50.5){\scriptsize~VESPCN~\cite{Caballero2017}}
        \put(6.2, 82){\small~$x^{i-1}_{lr}$}
        \put(6, 73.5){\small~$x^{i}_{lr}$}
        \put(5.5, 65.5){\small~$x^{i+1}_{lr}$}
        \put(82, 73){\small~$x^{i}_{sr}$}
        \put(20.1, 79){\scriptsize~\shortstack{Motion\\estimation}}
        \put(20.1, 64.5){\scriptsize~\shortstack{Motion\\estimation}}
        \put(37, 89){\scriptsize~Warp}
        \put(37, 57.5){\scriptsize~Warp}
        \put(52.5, 71){\scriptsize~\rotatebox{90}{Conv}}
        \put(63.2, 71){\scriptsize~\rotatebox{90}{Conv}}
        \put(73, 72){\scriptsize~\rotatebox{90}{UP}}

        \put(36, 2.2){\scriptsize~VSR-DUF~\cite{Jo2018}}
        \put(3, 39){\small~$x^{i-1}_{lr}$}
        \put(4.5, 27){\small~$x^{i}_{lr}$}
        \put(47, 43.5){\small~$x^{i}_{lr}$}
        \put(3, 15){\small~$x^{i+1}_{lr}$}
        \put(82, 26.5){\small~$x^{i}_{sr}$}
        \put(18, 36){\scriptsize~\rotatebox{90}{Conv}}
        \put(18, 24){\scriptsize~\rotatebox{90}{Conv}}
        \put(18, 12){\scriptsize~\rotatebox{90}{Conv}}
        \put(38, 30){\scriptsize~\rotatebox{90}{Conv}}
        \put(49, 30){\scriptsize~\rotatebox{90}{Conv}}
        \put(38, 18){\scriptsize~\rotatebox{90}{Conv}}
        \put(49, 18){\scriptsize~\rotatebox{90}{Conv}}
        \put(67, 31.5){\scriptsize~\rotatebox{90}{UP}}
    \end{overpic}
    \caption{Illustration of efficient modification. Gray dash line indicate that all convolutional layers are shared weights. $\bm{\bigoplus}$ indicates element-wise addition. $\bm{\bigotimes}$ indicates matrix product.} 
    \label{fig:vsr_efficient}
\end{figure}
For recurrent modeling, BRCN~\cite{huang2015bidirectional} set a foundation by utilizing a bidirectional recurrent convolutional network for multi-frame video super-resolution, effectively modeling temporal dependencies without the high computational cost of traditional RNNs. It replaces recurrent full connections with weight-sharing convolutional connections, enabling it to handle complex motions with reduced computational overhead. Building on this, STCN~\cite{guo2017building} further enhanced the modeling by combining deep spatial encoding with multi-scale temporal components, integrating both intra-frame visual patterns and inter-frame dynamics for higher-quality reconstructions.

To improve efficiency in frame-based VSR, Sajjadi~\etal~\cite{Sajjadi2018} introduced a frame-recurrent approach, which focuses on reusing previously inferred high-resolution frames to super-resolve the next, significantly reducing redundancy in processing consecutive frames. Similarly, RISTN~\cite{zhu2019residual} combined spatial and temporal modeling with a novel architecture that incorporates a lightweight residual invertible block for minimizing information loss and a recurrent convolutional model with dense connections to prevent feature degradation. Haris~\etal~\cite{haris2019recurrent} took this further by using a recurrent encoder-decoder module to integrate spatial and temporal contexts across continuous frames, demonstrating better performance for complex video sequences.

Efforts to further improve computational efficiency led to RLSP~\cite{fuoli2019efficient}, which uses high-dimensional latent states to propagate temporal information without the need for heavy motion compensation, achieving significant speed-ups over traditional methods like DUF, with a 70x improvement in runtime. Isobe~\etal~\cite{isobe2020videoeccv} refined this by separating input frames into structure and detail components, processing them through a two-stream recurrent unit, and using a hidden state adaptation module to mitigate error accumulation, enhancing robustness against appearance changes. More recently, Xia~\etal~\cite{xia2023structured} explored structured sparsity, pruning key components such as residual and recurrent blocks, resulting in even more efficient models with minimal performance loss.

For convolution modeling, early work by Caballero~\etal~\cite{Caballero2017} introduced spatio-temporal sub-pixel convolution networks, which effectively leverage temporal redundancies to improve both speed and accuracy. Building on this, RTVSR~\cite{bare2019real} proposed a real-time VSR method that uses 1D motion convolution kernels and an enhanced gated residual unit (GEU), achieving significant gains over state-of-the-art methods while maintaining real-time performance. FFCVSR~\cite{yan2019frame} further advanced this by integrating a local network for processing consecutive frames with a context network that leverages previously estimated high-resolution frames, leading to temporally consistent, high-quality results.

In the realm of feature alignment, TDAN~\cite{tian2020tdan} introduced deformable convolutions to adaptively align frames at the feature level without relying on optical flow, enabling better super-resolution results with a more compact network architecture. Jo~\etal~\cite{Jo2018} built on this by using dynamic upsampling filters to compute residual images based on the local spatio-temporal neighborhood, eliminating the need for explicit motion compensation. Zhu~\etal~\cite{Zhu2019} added to these innovations by incorporating a residual invertible block for more efficient spatial information extraction, alongside a sparse fusion strategy for better integration of spatial-temporal features.

Recent work has also seen the exploration of more specialized architectures. Li~\etal~\cite{li2019fast} introduced a fast spatio-temporal residual network using 3D convolutions to improve VSR performance with low computational cost. Similarly, Yi~\etal~\cite{yi2019progressive} proposed a progressive fusion network to better utilize spatio-temporal information across consecutive frames. Isobe~\etal~\cite{isobe2020video} took a hierarchical approach by splitting input sequences into groups based on frame rates, enhancing temporal information processing. Meanwhile, Pan~\etal~\cite{pan2021deep} developed a blind VSR method that jointly estimates blur kernels, motion fields, and latent images, demonstrating its effectiveness in various real-world scenarios. Finally, Li~\etal~\cite{li2023towards} introduced a content-adaptive approach that applies different super-resolution methods to different video patterns, achieving a better balance between performance and efficiency.}
\begin{figure}
\centering
    \begin{overpic}[scale=.45, tics=5]{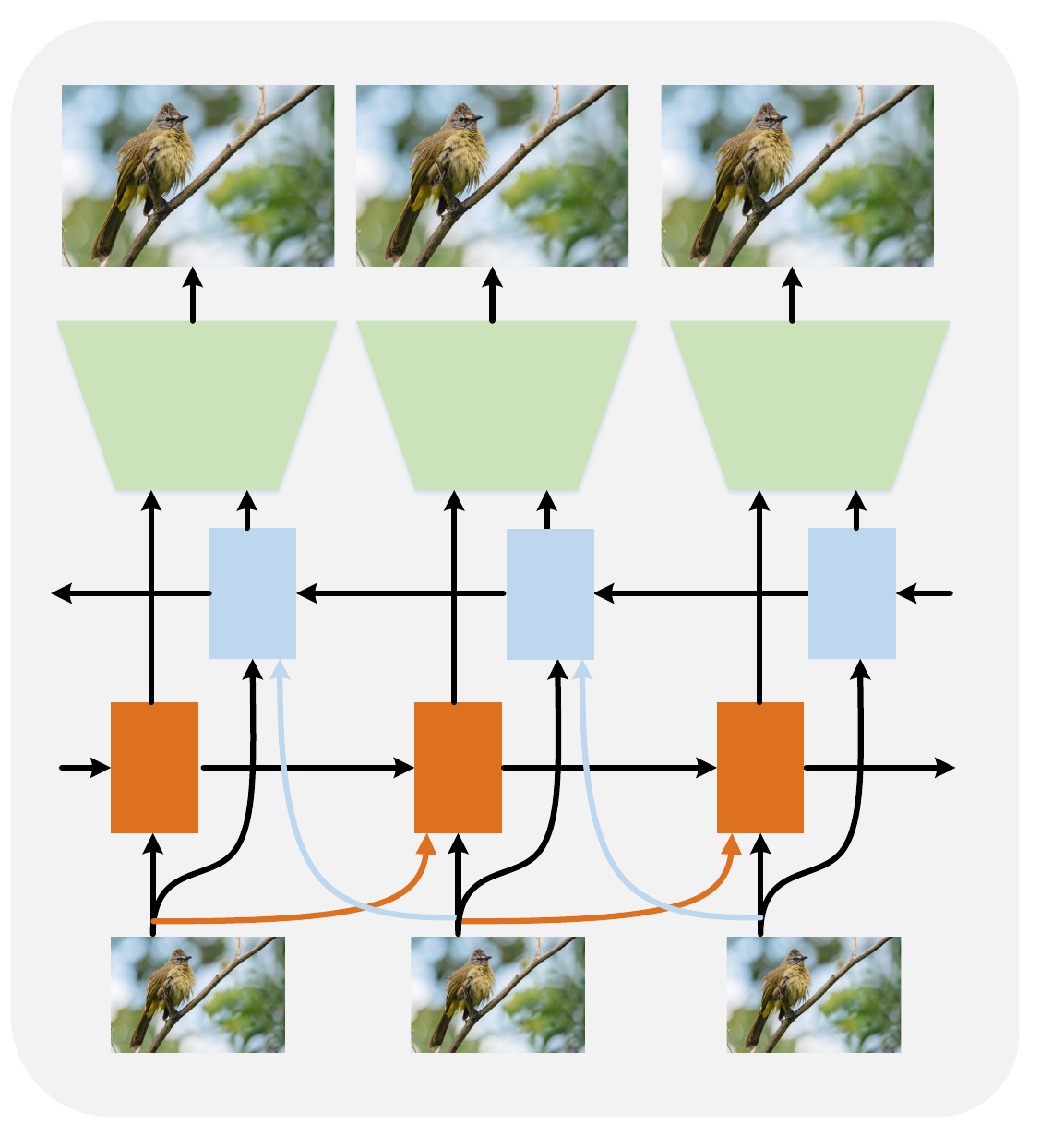}
        \put(13.5, 4.5){\small~$x^{i-1}_{lr}$}
        \put(40.5, 4.5){\small~$x^{i}_{lr}$}
        \put(66.2, 4.5){\small~$x^{i+1}_{lr}$}
        
        \put(13.5, 94.5){\small~$x^{i-1}_{sr}$}
        \put(40.5, 94.5){\small~$x^{i}_{sr}$}
        \put(66.2, 94.5){\small~$x^{i+1}_{sr}$}

        \put(19, 47){\normalsize~$F_b$}
        \put(45, 47){\normalsize~$F_b$}
        \put(72, 47){\normalsize~$F_b$}

        \put(11, 32){\normalsize~$F_f$}
        \put(37, 32){\normalsize~$F_f$}
        \put(64, 32){\normalsize~$F_f$}

        \put(13, 64){\normalsize~$UP$}
        \put(40, 64){\normalsize~$UP$}
        \put(69, 64){\normalsize~$UP$}
        
    \end{overpic}
    \caption{The BasicVSR~\cite{chan2021basicvsr} framework comprehensively examined the four fundamental components of VSR, namely Propagation, Alignment, Aggregation, and Upsampling, thereby establishing a robust and comprehensive foundation for future advancements in the field of VSR.}
\label{fig:basicsr}
\end{figure}
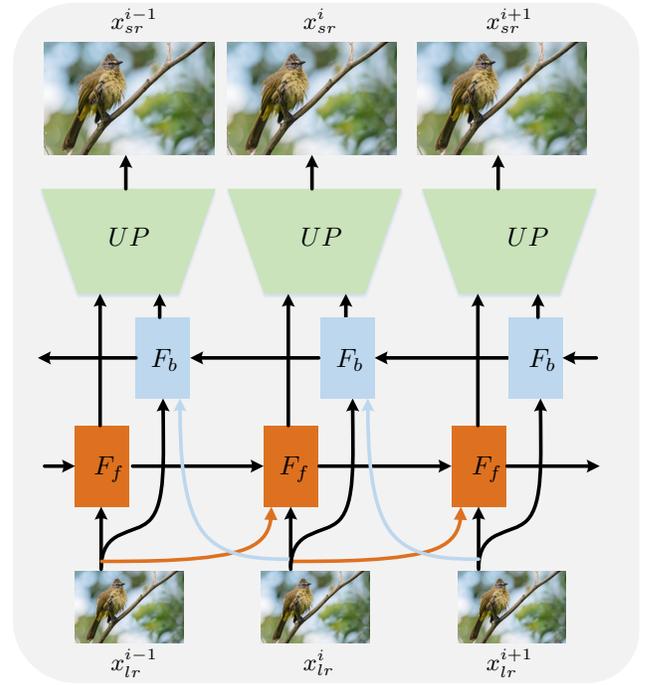

\textbf{Learning Better Representations:} \revision{VSRnet~\cite{kappeler2016video} introduces a CNN trained on both spatial and temporal dimensions of videos to enhance video super-resolution, using motion-compensated frames and exploring various frame-combining architectures, demonstrating that pretraining with images allows a smaller video database to effectively train the model and achieve superior performance compared to current video and image SR algorithms. SOFVSR~\cite{wang2019learningVSR} introduces an end-to-end trainable video super-resolution framework that enhances both image quality and optical flows, leveraging high-resolution optical flows for accurate motion compensation and improved correspondence, thereby achieving superior accuracy and consistency in super-resolved video sequences, as validated by extensive testing on benchmark datasets. MMCNN~\cite{wang2018multi} integrates an optical flow network and an image-reconstruction network with convolutional long short-term memory embedded in residual blocks to effectively utilize both spatial and temporal correlations between low-resolution frames, demonstrating superior performance in terms of PSNR and visual quality across various scaling factors in extensive experiments. RRCN~\cite{li2018video} leverages motion compensation, deep recurrent convolutional layers, and late fusion to effectively utilize temporal information, incorporates residual connections for enhanced accuracy, and employs a novel model ensemble strategy with a single-image SR method, demonstrating superior performance over existing state-of-the-art SR methods in quantitative visual quality assessments. MultiBoot VSR~\cite{kalarot2019multiboot} employs a unique structure comprising a motion compensation-based input subnetwork, a blending backbone, and a spatial upsampling subnetwork, applied recurrently to enhance image quality progressively by reshuffling high-resolution frames into multiple low-resolution images, demonstrating the ability to generate temporally consistent and high-quality results without artifacts. DNLN~\cite{wang2019deformable} introduced deformable convolution to adaptively align features and a non-local structure to capture global correlations, further enhanced by residual in residual dense blocks to fully utilize hierarchical features, achieving state-of-the-art performance on benchmark datasets. 3DSRnet~\cite{kim2019video} leverages spatio-temporal feature maps without the need for motion alignment preprocessing, maintaining temporal depth to capture the dynamic relationship between low and high-resolution frames, and incorporates residual learning with sub-pixel outputs, significantly outperforming state-of-the-art methods on the Vidset4 benchmark and addressing performance drops due to scene changes.} Tao~\etal~\cite{tao2017detail} highlighted the importance of proper frame alignment and motion compensation and proposed a sub-pixel motion compensation layer in a CNN framework. Xue~\etal~\cite{xue2019video} proposed a task-oriented framework to produce specific motion estimation and train with video processing simultaneously. Kim~\etal~\cite{Kim2020} proposed a multi-scale temporal loss to regularize spatio-temporal resolution. Chan \emph{et. al}~\cite{Chan2021aaai} introduced an offset-fidelity loss for deformable alignment in VSR to control offset learning. Isobe~\etal~\cite{Isobe2022} considered the temporal relation by computing the difference between frames and processing these pixels into two branches based on the level of difference. Chen~\etal~\cite{Chen2022cvpr} extended fixed scale to arbitrary scale VSR using implicit neural representation.  Chiche~\etal~\cite{Chiche2022cvpr} addressed high-frequency artifacts caused by recurrent VSR models using Lipschitz stability theory to generate stable and competitive results. Chan~\etal~\cite{Chan2022cvpr} designed a pre-cleaning stage to reduce noise and artifacts caused by long-term propagation cumulative error. They also proposed a stochastic degradation scheme for a better tradeoff between speed and performance and employed longer sequences instead of large batches to allow more effective use of temporal information. MFPI~\cite{li2023multi} introduces a novel multi-frequency representation enhancement module (MFE) that aggregates spatial-temporal information in the frequency domain and employs privilege training to efficiently utilize high-resolution video data, significantly surpassing existing state-of-the-art VSR models in both performance and efficiency across multiple datasets.

\textbf{Improving Information Interaction:} Compared to the previous approach of extracting features in the spatial domain and performing motion compensation in the temporal domain, which can lead to a lack of joint utilization of spatiotemporal information and thus affect super-resolution performance, D3DNet~\cite{ying2020deformable} leverages deformable 3D convolutions to fully exploit the spatiotemporal dependencies of video frames, thereby enhancing video super-resolution performance. Wang~\etal~\cite{wang2018learning} super-resolved optical flows and images simultaneously to obtain better performance. Yan~\etal~\cite{Yan2019} utilized inter LR frames and previous SR frames to produce temporally consistent high-quality results.  Li~\etal~\cite{li2020mucan} proposed a temporal multi-correspondence aggregation strategy to leverage similar patches across frames and a cross-scale non-local correspondence aggregation scheme to explore self-similarity of images across scales. Chan~\etal~\cite{chan2021basicvsr} reconsidered essential components for VSR and proposed a succinct pipeline as shown in Fig.~\ref{fig:basicsr}, BasicVSR, that achieved appealing improvements in speed and restoration quality compared to many state-of-the-art algorithms. Liu~\etal~\cite{liu2021large} proposed a deep neural network with dual subnet and multi-stage communicated up-sampling for VSR with large motion. Yi~\etal~\cite{yi2021omniscient} introduced a hybrid structure that utilized preceding, present, and future SR information. Li~\etal~\cite{Li2021} focused on compression VSR and proposed bi-directional recurrent warping, detail-preserving flow estimation, and Laplacian enhancement modules to handle compression properties. Yu~\etal~\cite{Yu2022} proposed cross-frame non-local attention to alleviate performance drops in large motion scenes and introduced a memory-augmented attention module to compensate for information loss caused by large motions. Cao~\etal~\cite{Cao2022eccv} formulated STVSR as a joint deblurring, interpolation, and super-resolution problem and introduced Fourier transform and recurrent enhancement layers to recover high-frequency details. Wang~\etal~\cite{wang2023compression} introduced metadata to predict the compression level combined with a RNN-style decoder to produce pleasing results.
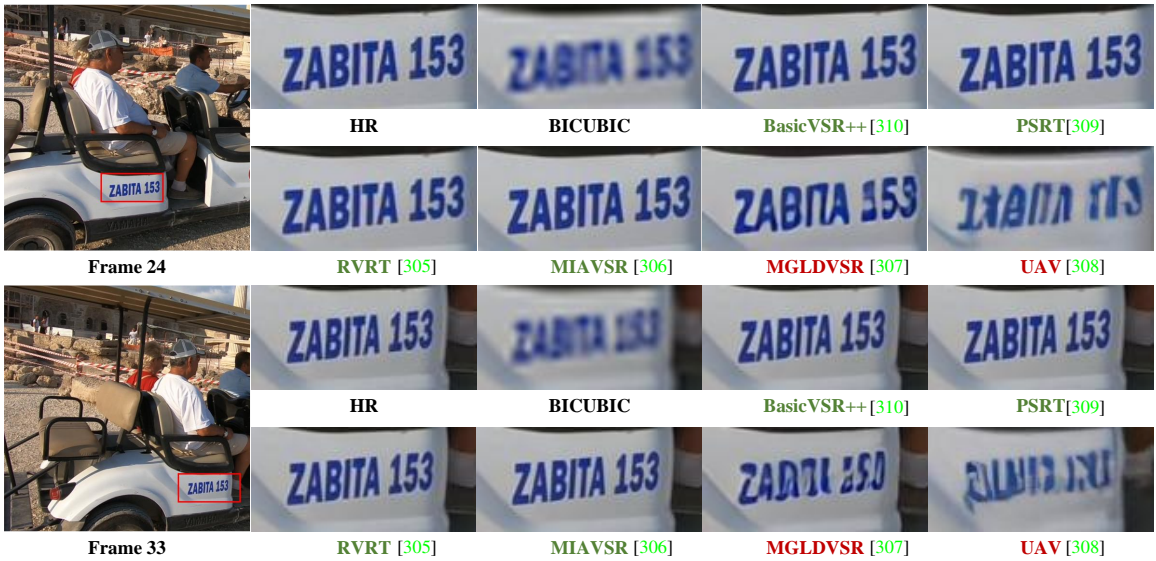
\begin{figure*}[!th]
    \centering
    \begin{overpic}[scale=.3]{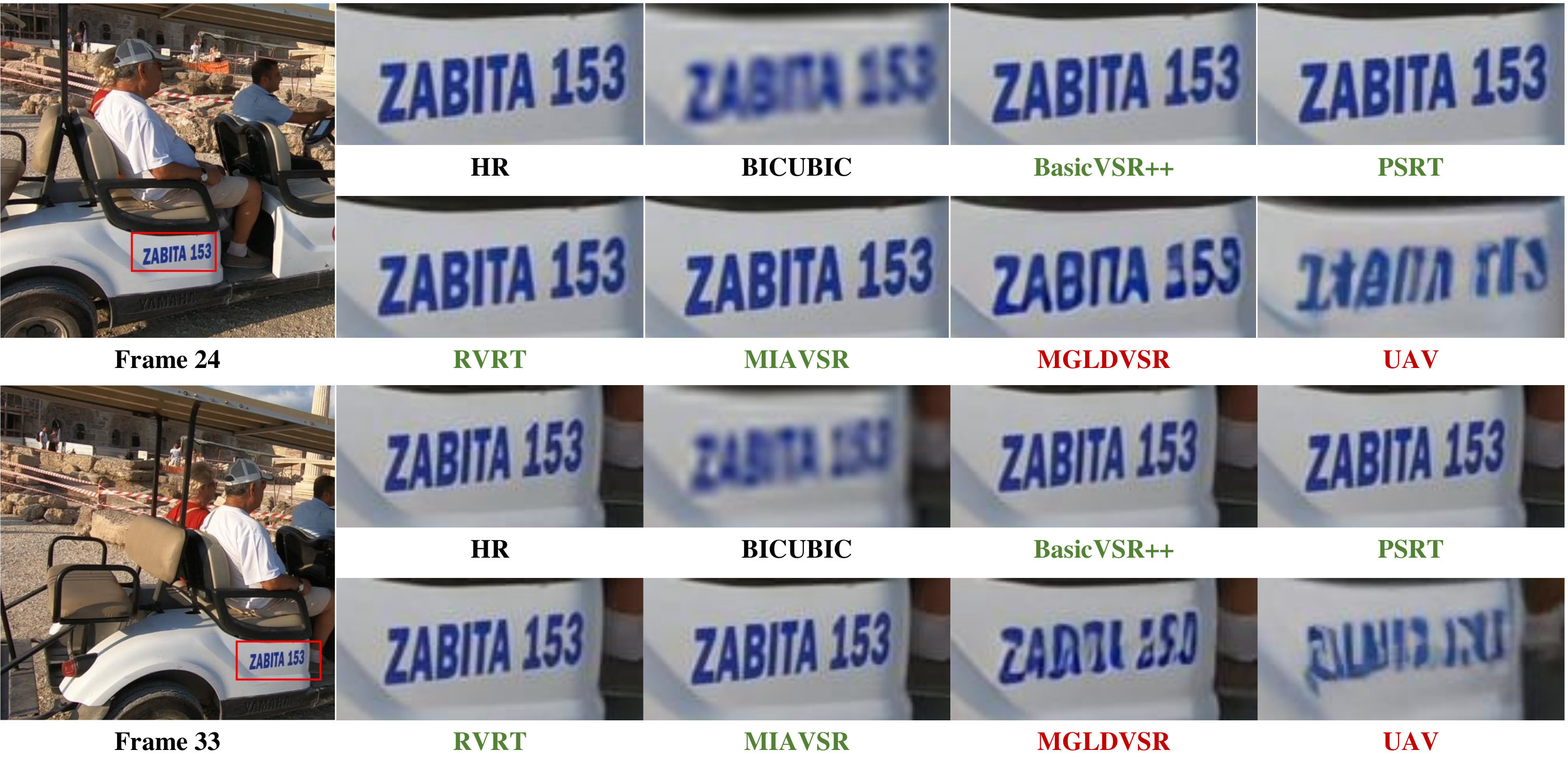}
        \put(33, 25.3){\scriptsize~~\cite{liang2022recurrent}}
        \put(33, 0.9){\scriptsize~~\cite{liang2022recurrent}}
        \put(53.5, 0.9){\scriptsize~~\cite{zhou2024video}}
        \put(74, 0.9){\scriptsize~~\cite{yang2023mgldvsr}}
        \put(91, 0.9){\scriptsize~~\cite{zhou2024upscale}}
        \put(91, 13.1){\scriptsize~~\cite{shi2022rethinking}}
        \put(74, 13.1){\scriptsize~~\cite{chan2022basicvsr++}}
        \put(53.5, 25.3){\scriptsize~~\cite{zhou2024video}}
        \put(74, 25.3){\scriptsize~~\cite{yang2023mgldvsr}}
        \put(91, 25.3){\scriptsize~~\cite{zhou2024upscale}}
        \put(91, 37.5){\scriptsize~~\cite{shi2022rethinking}}
        \put(74, 37.5){\scriptsize~~\cite{chan2022basicvsr++}}
    \end{overpic}
    
    \caption{\resubmit{Visual comparison of \textcolor{regressive}{\textbf{regression-based}} and \textcolor{generative}{\textbf{generation-based}} video super-resolution methods on selected frames from a video sequence. The bicubic upsampling result is included as baseline reference. Notably, \textcolor{regressive}{\textbf{regression-based}} methods like RVRT~\cite{liang2022recurrent} and MIAVSR~\cite{zhou2024video} tend to produce more faithful and recognizable text (``ZABITA 153"), even if slightly less sharp. In contrast, \textcolor{generative}{\textbf{generation-based}} methods such as MGLDVSR~\cite{yang2023mgldvsr} and UAV~\cite{zhou2024upscale}, while attempting to generate sharper details, often introduce significant artifacts or hallucinate incorrect content, as seen with the distorted or altered text in their outputs.}}
    \label{fig:vsr_reg_vis}
\end{figure*}

\subsubsection{\textbf{Transformer-based VSR}}
\
\newline
\indent Cao~\etal~\cite{cao2021video} made the first attempt to adopt a transformer for VSR. In addition, a spatial-temporal convolutional self-attention layer is proposed to exploit locality information. Liang~\etal~\cite{liang2022vrt} proposed a video restoration transformer with parallel frame prediction and long-range temporal dependency modeling abilities. Wang~\etal~\cite{wang2022stdan} designed a deformable attention network and proposed a long-short term feature interpolation module, where abundant information from more neighboring frames is explored for the interpolation process of missing frame features. Geng~\etal~\cite{geng2022rstt} addressed the slow inference speed issue by using a spatial-temporal transformer that naturally incorporates spatial and temporal super-resolution modules into a single model. \revision{PSRT~\cite{geng2022rstt} introduces an efficient patch alignment method that aligns image patches rather than pixels, enabling state-of-the-art performance across multiple benchmarks and offering new insights into the optimal use of multi-frame information and alignment techniques in VSR. FTVSR++~\cite{qiu2023learning} introduces a degradation-robust approach to video super-resolution by implementing self-attention within a combined space-time-frequency domain, featuring a novel dual frequency attention mechanism and divided attention strategy that effectively distinguish real textures from artifacts and address complex degradations, significantly enhancing video quality across various challenging conditions. RVRT~\cite{liang2022recurrent} proposed clip-based feature estimation and guided deformable attention for clip alignment, effectively balancing model size, effectiveness, and efficiency, and achieving state-of-the-art performance in video super-resolution, deblurring, and denoising on benchmark datasets. TTVSR~\cite{liu2022learning} leverages pre-aligned spatio-temporal trajectories and a cross-scale feature tokenization module to efficiently model long-range dependencies and address scale variations in video sequences, outperforming state-of-the-art models in extensive benchmarks through its novel approach to reducing computational costs and enhancing the utilization of temporal information.}
\subsection{\textbf{Generative Models}}
\subsubsection{\textbf{GAN-based VSR}}
\
\newline
\indent Lucas~\etal~\cite{lucas2019generative} designed a new discriminator architecture optimized for the VSR problem and introduced distance losses in feature-space and pixel-space to regularize GAN training. Thawakar~\etal~\cite{thawakar2019image} incorporated a recurrent mechanism into the GAN model to reduce the parameter count as the model depth increases. Yi~\etal~\cite{yi2020progressive} proposed a hybrid attention structure to capture local and non-local spatio-temporal correlations. They also employed generative adversarial training to mitigate temporal artifacts and generate more realistic results. Chadha~\etal~\cite{chadha2020iseebetter} proposed a recurrent back-projection network as a generator to extract spatial and temporal information, and they combined multiple losses to capture fine details. Cao~\etal~\cite{cao2021real} achieved a 7.92× performance speedup compared to state-of-the-art methods through BN layer fusion, convolutional layer acceleration, and efficient upsampling methods. Andrei~\etal~\cite{andrei2021supervegan} focused on low bitrate streams and proposed a two-stage VSR strategy. The first stage removed compression artifacts, and the second stage generated detailed information using a GAN model with perceptual metrics. Wang~\etal~\cite{wang2021video} introduced the Laplacian edge module to enhance intermediate results generated by the GAN model and added perceptual loss to obtain better visual quality. \revision{Chu~\etal~\cite{chu2020learning} introduced a temporal learning strategy to supervise both spatial and temporal relationships in VSR. They also proposed a Ping-Pong loss to alleviate temporal artifacts and obtain more natural super-resolved images. Wang~\etal~\cite{wang2019spatially} explored spatial information and introduced a spatially adaptive loss to enhance GAN model performance. López-Tapia~\etal~\cite{lopez2020single} introduced the Amortized MAP approximation to simulate multiple degradations and designed a GAN loss to eliminate artifacts. LongVideoGAN~\cite{brooks2022generating} enhances long-term temporal consistency and dynamic content creation by redesigning the temporal latent representation and utilizing a two-phase training strategy with varying video lengths and resolutions.}
\begin{table*}[!t]
  \caption{\revision{Performance comparison of different VSR models on the Vid~\cite{liu2013bayesian}, Vimeo-90K~\cite{2019Video}, and REDS~\cite{nah2019ntire} datasets for BI degradation, and on the Vid~\cite{liu2013bayesian}, Vimeo-90K~\cite{2019Video}, and UDM10~\cite{yi2019progressive} datasets for BD degradation, all under the $\times 4$ task. PSNR/SSIM values are reported for the Y channel across each dataset. ``Params" denotes the total number of network parameters. The \textbf{\textcolor{blue}{best}} and \textcolor{blue}{\underline{second best}} performances are highlighted for models trained on Vimeo-90K~\cite{2019Video}, while the \textbf{\textcolor{red}{best}} and \textcolor{red}{\underline{second best}} performances are highlighted for models trained on REDS~\cite{nah2019ntire}+Vimeo-90K~\cite{2019Video}. The results for each model are obtained from their respective original papers.}}
  \centering
  \scriptsize 
  \setlength{\tabcolsep}{1pt}{
  \begin{tabular}{c|c|ccc|ccc|c|c|c}
     \toprule[1.5pt]
     \multirow{2}*{Methods} &  \multirow{2}*{\makecell[c]{Params\\(M)}} & \multicolumn{3}{c|}{BI} & \multicolumn{3}{c|}{BD} & \multirow{2}*{Training Sets} & \multirow{2}*{Backbone} & \multirow{2}*{Venue} \\
     \cmidrule(lr){3-5}\cmidrule(lr){6-8}
     & & Vid4~\cite{liu2013bayesian} & Vimeo-90K~\cite{2019Video} & REDS~\cite{nah2019ntire} & Vid~\cite{liu2013bayesian} & Vimeo-90K~\cite{2019Video} & UDM10~\cite{yi2019progressive} & & \\
     \midrule
     \hline
     \rowcolor{light-gray} Bicubic\qquad  & - & 23.78/0.6347 & 31.32/0.8684 & 26.14/0.7292 & 21.80/0.5246 & 31.30/0.8687 & 28.47/0.8253 & - & - & -\\
     BRCN~\cite{huang2015bidirectional} & - & - & - & - & 24.43/0.6334 & - & - & YUV25 & CNN & NeurIPS'15\\
     \rowcolor{light-gray}VSRnet~\cite{kappeler2016video} & 0.27 & 24.84/0.7049 & - & - & - & - & - & Myanmar & CNN & TCI'16\\
     RRCN~\cite{li2018video} & - & 25.86/0.7591 & - & - & - & - & - & Myanmar & CNN & TIP'19\\
     \rowcolor{light-gray}VESPCN~\cite{Caballero2017} & 0.88 & 25.35/0.7557 & - & - & - & - & - & CDVL & CNN & CVPR'17\\
     SOFVSR~\cite{wang2018learning} & 1.71 & 26.01/0.7710 & - & - & 26.19/0.7850 & - & - & CDVL & CNN & ACCV'18\\
     \rowcolor{light-gray}RTVSR~\cite{bare2019real} & 15.00 & 26.36/0.7900 & - & - & - & - & - & harmonicinc.com & CNN & Neurocomp.'19\\
     MultiBoot VSR~\cite{kalarot2019multiboot} & 60.86 & - & - & 31.00/0.8822 & - & - & - & harmonicinc.com & CNN & CVPRW'19\\
     \rowcolor{light-gray}VSRResFeatGAN~\cite{lucas2019generative} & - & 25.51/0.7530 & - & - & - & - & - & Myanmar & GAN & TIP'19\\
     FFCVSR~\cite{yan2019frame} & - & 26.97/0.8300 & - & - & - & - & - & Venice+Myanmar & CNN & AAAI'19\\
     \rowcolor{light-gray}3DSRnet~\cite{kim2019video} & 0.11 & 25.71/0.7588 & - & - & - & - & - & largeSet & CNN & ICIP'19\\
     DSMC~\cite{liu2021large} & 11.58 & 27.29/0.8403 & - & 30.29/0.8381 & - & - & - & REDS & CNN & AAAI'21\\
     \rowcolor{light-gray}DRVSR~\cite{tao2017detail} & 2.17 & 25.52/0.7600 & - & - & - & - & - & - & CNN & ICCV'17\\
     DUF~\cite{jo2018deep} & - & 27.33/0.8251 & - & 28.63/0.8251 & 27.38/0.8392 & 36.87/0.9447 & 38.48/0.9605 & - & CNN & CVPR'18\\
     \rowcolor{light-gray}MMCNN~\cite{wang2018multi} & 10.58 & 26.28/0.7844 & - & - & - & - & - & - & CNN & TIP'19\\
     TecoGAN~\cite{chu2020learning} & 3.00 & - & - & - & 25.89/- & - & - & - & GAN & TOG'20\\
     \rowcolor{light-gray}GOVSR~\cite{yi2021omniscient} & - & - & - & - & 28.41/0.8724 & 37.63/0.9503 & 30.14/0.9713 & - & CNN & ICCV'21\\
     PSRT~\cite{shi2022rethinking} & 13.4 & 28.07/0.8485 & 38.27/0.9536 & 32.72/0.9106 & - & - & - & - & Transformer & NeurIPS22\\
     \rowcolor{light-gray}StableVSR~\cite{rota2023enhancing} & - & - & 31.97/0.8770 & 27.97/0.8000 & - & - & - & - & DDPM & ECCV'24\\
    \hline
    \hline
     FRVSR~\cite{Sajjadi2018} & 5.1 & - & - & - & 26.69/0.8220 & 35.64/0.9319 & 37.09/0.9522 & Vimeo-90K & CNN & CVPR'18\\
     \rowcolor{light-gray}DNLN~\cite{wang2019deformable} & 19.74 & 27.31/0.8257 & - & - & - & - & - & Vimeo-90K & CNN & ACCESS'19\\
     TOFlow~\cite{xue2019video} & 1.4 & 25.89/0.7651 & 33.08/0.9054 & 27.98/0.7990 & 25.85/0.7695 & 34.62/0.9212 & 36.26/0.9438 & Vimeo-90K & CNN & IJCV'19\\

     \rowcolor{light-gray} RLSP~\cite{fuoli2019efficient} & 4.21 & 27.55/0.8380 & - & - & 27.48/0.8388 & 36.49/0.9403 & - & Vimeo-90K & CNN & ICCVW'19\\
     PFNL~\cite{yi2019progressive} & 3.0 & 26.73/0.8029 & 36.14/0.9363 & 29.63/0.8502 & 27.16/0.8355 & - & 38.74/0.9627 & vimeo-90K & CNN & ICCV'19\\

     \rowcolor{light-gray} RBPN~\cite{haris2019recurrent} & 12.2 & 27.12/0.8180 & 37.07/0.9435 & 30.09/0.8590 & 27.17/0.8205 & 37.20/0.9458 & 38.66/0.9596 & Vimeo-90K & CNN & CVPR'19\\
     
     MEMC-Net~\cite{bao2019memc} & - & 24.37/0.8380 & 33.47/\textcolor{blue}{\textbf{0.9470}} & - & - & - & - & Vimeo-90K & CNN & TPAMI'19\\

     \rowcolor{light-gray}RISTN~\cite{zhu2019residual} & 3.67 & 26.13/0.7920 & - & - & - & - & - & Vimeo-90K & CNN & AAAI'19\\

     TGA~\cite{isobe2020video} & 5.8 & \textcolor{blue}{\underline{27.59}}/\textcolor{blue}{\underline{0.8419}} & - & - & 27.63/0.8423 & \textcolor{blue}{\textbf{37.59}}/\textcolor{blue}{\textbf{0.9516}} & 38.74/0.9627 & Vimeo-90K & CNN & CVPR'20\\

     \rowcolor{light-gray} TDAN~\cite{tian2020tdan} & 1.97 & 26.24/0.7800 & - & - & 26.58/0.8010 & - & - & Vimeo-90K & CNN & CVPR'20\\

     STARnet~\cite{haris2020space} & 111.61 & - & 30.83/0.9290 & - & - & - & - & Vimeo-90K & CNN & CVPR'20\\
     \rowcolor{light-gray}STVSR~\cite{xiang2020zooming} & 11.1 & 26.31/0.7976 & - & 25.27/0.8256 & - & - & - & Vimeo-90K & CNN & CVPR'20\\
     
     RSDN~\cite{isobe2020videoeccv} & 6.2 & \textcolor{blue}{\textbf{27.79}}/\textcolor{blue}{\textbf{0.8474}} & 37.05/0.9454 & - & \textcolor{blue}{\underline{27.92}}/\textcolor{blue}{\underline{0.8505}} & \textcolor{blue}{\underline{37.23}}/\textcolor{blue}{\underline{0.9471}} & \textcolor{blue}{\underline{39.35}}/\textcolor{blue}{\underline{0.9653}} & Vimeo-90K & CNN & ECCV'20\\
     
     \rowcolor{light-gray}MuCAN~\cite{li2020mucan} & 19.9 & - & \textcolor{blue}{\underline{37.32}}/0.9465 & \textcolor{blue}{\underline{30.88}}/\textcolor{blue}{\underline{0.8750}} & - & - & - & Vimeo-90K & CNN & ECCV'20\\ 
     D3Dnet~\cite{ying2020deformable} & 2.58 & 26.52/0.7990 & - & - & - & - & - & Vimeo-90K & CNN & SPL'20\\
     \rowcolor{light-gray}DNSTNet~\cite{sun2020video} & - & 27.21/0.8220 & 36.86/0.9387 & - & - & - & - & Vimeo-90K & CNN & Neurocomp.'20\\
     MSFFN~\cite{song2021multi} & 8.5 & 27.23/0.8218 & \textcolor{blue}{\textbf{37.33}}/\textcolor{blue}{\underline{0.9467}} & - & - & - & - & Vimeo-90K & CNN & TIP'21\\ 
     \rowcolor{light-gray}RSTT~\cite{geng2022rstt} & 7.67 & 26.43/0.7994 & - & - & - & - & - & Vimeo-90K & Transformer & CVPR22\\

     ETDM~\cite{Isobe2022} & 8.4 & - & - & \textcolor{blue}{\textbf{32.15}}/\textcolor{blue}{\textbf{0.9024}} & \textcolor{blue}{\textbf{28.81}}/\textcolor{blue}{\textbf{0.8725}} & -/- & \textcolor{blue}{\textbf{40.11}}/\textcolor{blue}{\textbf{0.9707}} & Vimeo-90K & CNN & CVPR'22\\

    \hline
    \hline
    \rowcolor{light-gray}EDVR~\cite{wang2019edvr} & 20.60 & 27.35/0.8264 & 37.61/0.9489 & 31.09/0.8800 & - & - & - & REDS+Vimeo-90K & CNN & CVPRW'19\\

    RRN~\cite{isobe2020revisiting} & 3.4 & - & - & - & 27.69/0.8488 & - & 38.96/0.9644 & REDS+Vimeo-90K & CNN & BMVC'20\\
    
     \rowcolor{light-gray} BasicVSR~\cite{chan2021basicvsr} & 6.3 & 27.24/0.8251 & 37.18/0.9450 & 31.42/0.8909 & 27.96/0.8553 & 37.53/0.9498 & 39.96/0.9694 & REDS+Vimeo-90K & CNN & CVPR'21\\
     
     IconVSR~\cite{chan2021basicvsr} & 8.7 & 27.39/0.8279 & 37.47/0.9476 & 31.67/0.8948 & 28.04/0.8570 & 37.84/0.9524 & 40.03/0.9694 & REDS+Vimeo-90K & CNN & CVPR'21\\

     \rowcolor{light-gray} BasicVSR++~\cite{chan2022basicvsr++} & 7.3 & 27.79/0.8400 & 37.79/0.9500 & 32.39/0.9069 & 29.04/0.8753 & 38.21/0.9550 & 40.72/0.9722 & REDS+Vimeo-90K & CNN & CVPR'22\\

     RVRT~\cite{liang2022recurrent} & 10.8 & 27.99/0.8462 & 38.15/0.9527 & 32.75/\textcolor{red}{\underline{0.9113}} & \textcolor{red}{\textbf{29.54}}/\textcolor{red}{\textbf{0.8810}} & \textcolor{red}{\underline{38.59}}/\textcolor{red}{\underline{0.9576}} & \textcolor{red}{\underline{40.90}}/\textcolor{red}{\underline{0.9729}} & REDS+Vimeo-90K & Transformer & NeurIPS'22\\
     
     \rowcolor{light-gray}TTVSR~\cite{liu2022learning} & 6.8 & - & - & 32.12/0.9021 & 28.40/0.8643 & 37.92/0.9526 & 40.41/0.9712 & REDS+Vimeo-90K & Transformer & CVPR'22\\
     
     SSL~\cite{xia2023structured} & 1.0 & 27.15/0.8208 & 36.82/0.9419 & 31.06/0.8933 & 27.56/0.8431 & 37.06/0.9458 & 39.35/0.9665 & REDS+Vimeo-90K & CNN & CVPR'23\\

     \rowcolor{light-gray} MFPI~\cite{li2023multi} & 7.3 & 28.11/0.8481 & \textcolor{red}{\textbf{38.28}}/\textcolor{red}{\textbf{0.9534}} & \textcolor{red}{\textbf{32.81}}/0.9106 &  \textcolor{red}{\underline{29.34}}/\textcolor{red}{\underline{0.8781}} & \textcolor{red}{\textbf{38.70}}/\textcolor{red}{\textbf{0.9579}} & \textcolor{red}{\textbf{41.08}}/\textcolor{red}{\textbf{0.9741}} & REDS+Vimeo-90K & CNN & ICCV'23\\

     FTVSR++~\cite{qiu2023learning} & 10.8 & \textcolor{red}{\textbf{28.80}}/\textcolor{red}{\textbf{0.8680}} & - & 32.42/0.9070 & - & - & - & REDS+Vimeo-90K & Transformer & TPAMI'23\\
     \rowcolor{light-gray}VideoGigaGAN~\cite{xu2024gigagan} & 369 & 26.78/- & 35.97/- & 30.46/- & 27.04/- & 35.30/- & 36.57/- & REDS+Vimeo-90K & GAN & ArXiv'24\\
     MIAVSR~\cite{zhou2024video} & 6.35 & \textcolor{red}{\underline{28.20}}/\textcolor{red}{\underline{0.8507}} & \textcolor{red}{\underline{38.22}}/\textcolor{red}{\underline{0.9532}} & \textcolor{red}{\underline{32.78}}/\textcolor{red}{\textbf{0.9220}} & - & - & - & REDS+Vimeo-90K & Transformer & CVPR'24\\
     \rowcolor{light-gray}SATeCo~\cite{chen2024learning} & - & 27.44/0.8420 & \textcolor{red}{\underline{38.22}}/\textcolor{red}{\underline{0.9532}} & 31.62/0.8932 & - & - & - & REDS+Vimeo-90K & DDPM & CVPR'24\\
     
     \hline
     \bottomrule
  \end{tabular}}
  \label{tab:vsr_results}
\end{table*}
\revision{\subsubsection{\textbf{Diffusion-based VSR}}
\
\newline
\indent To address the dual challenges of spatial fidelity and temporal consistency in video super-resolution. Zhou~\etal~\cite{zhou2024upscale} introduces a text-guided latent diffusion framework for video upscaling that ensures temporal coherence by integrating temporal layers into its architecture and employing a flow-guided recurrent latent propagation module, offering enhanced video stability and flexibility in texture creation through text prompts and adjustable noise levels, thereby achieving superior visual realism and consistency as demonstrated across various benchmarks. Yuan~\etal~\cite{yuan2024inflation} utilizes an efficient diffusion-based text-to-video super-resolution tuning approach that expands the capabilities of text-to-image SR models into video generation by inflating model weights and integrating a temporal adapter, enabling the creation of high-quality, temporally coherent videos with a balance between computational efficiency and super-resolution performance, as validated by thorough testing on the Shutterstock video dataset. StableVSR~\cite{rota2023enhancing} employs Diffusion Models enhanced with a Temporal Conditioning Module and Frame-wise Bidirectional Sampling to synthesize realistic and temporally-consistent details, achieving superior perceptual quality and temporal consistency in upscaled videos. MGLDVSR~\cite{yang2023mgldvsr} integrates pre-trained latent diffusion models with a motion-guided loss and a temporal module in the decoder, optimizing the latent sampling path to ensure content consistency and temporal coherence in high-resolution video generation from low-resolution inputs, significantly surpassing existing methods in perceptual quality on real-world VSR benchmarks. Imagen Video~\cite{ho2022imagen} uses a cascade of video diffusion models, including interleaved spatial and temporal super-resolution, to produce high-definition videos with high fidelity, controllability, and diverse stylistic and 3D capabilities. SATeCo~\cite{chen2024learning} proposes lightweight Spatial Feature Adaptation (SFA) and Temporal Feature Alignment (TFA) modules. These leverage pixel-wise affine modulation and tubelet-based cross-attention, respectively, to guide denoising and reconstruction—while keeping the pretrained diffusion model frozen for efficiency.}
\subsection{\textbf{Training Details}}
\subsubsection{\textbf{Datasets}}
\
\newline
\indent The evaluation of VSR methods is conducted using the following datasets:
\begin{itemize}
\item REDS~\cite{nah2019ntire}: The NTIRE19 challenge dataset, consisting of sequences with 100 frames each, and a resolution of $720 \times 1280$.
\item Vimeo-90K~\cite{2019Video}: A large-scale, high-quality video dataset comprising 89,800 video clips with a resolution of $448 \times 256$.
\item Vid4~\cite{liu2013bayesian}: This dataset includes four categories: walk, city, calendar, and foliage. The resolutions are $720 \times 480$ for Foliage and Walk, $720 \times 576$ for Calendar, and $704 \times 576$ for City.
\item UDM10~\cite{yi2019progressive}: A video dataset containing 522 video sequences for training, collected from 10 videos.
\item UDF~\cite{Jo2018}: This dataset comprises 351 Internet videos with diverse content, including wildlife, activity, and landscape.
\item SPMC-11~\cite{tao2017detail}: It consists of 975 sequences from high-quality 1080p HD video clips, mostly shot with high-end cameras and containing rich details in natural-world and urban scenes.
\end{itemize}

\subsubsection{\textbf{Training Settings}}
\
\newline
\indent\textbf{Patch Extraction:} LR video frames are partitioned into fixed-size patches, such as $64\times64$, either non-overlapping or with overlap. Overlapping patches are favored to encompass information near boundaries, enhancing the super-resolution outcome.

\textbf{Data Augmentation:} VSR employs spatial and temporal transformations on training data to boost diversity and generalization. Spatial transformations encompass random cropping, flipping, and rotation, while temporal transformations involve jittering, frame shuffling, and skipping.

\textbf{Batch Size and Temporal Span:} The batch size typically ranges from $4$ to $16$ video sequences based on available GPU memory and computational resources. The temporal span, ranging from $5$ to $15$ frames, captures temporal coherence, ensuring sharper and temporally consistent high-resolution video frames.

\textbf{Learning Schedule:} A step-wise learning rate decay is often used, reducing the learning rate by a factor of 10 after a set number of epochs. For instance, training for 200 epochs with an initial learning rate of $10^{-3}$ and decaying at the $100$th epoch to $10^{-4}$.

\resubmit{\subsection{\textbf{Experimental Results}}
\textbf{Quantitative Analysis.} The performance of various VSR methods is systematically evaluated in Table~\ref{tab:vsr_results} using PSNR and SSIM metrics under two degradation scenarios: Bicubic (BI) and Blur Downsampling (BD). The results reveal several important trends. First, VSR models consistently demonstrate higher parameter counts than their SISR counterparts, reflecting the inherent complexity of temporal modeling. Second, model performance shows strong dependence on training data scale, with methods trained on larger datasets like REDS and Vimeo-90K achieving superior metrics. Most notably, transformer-based architectures have established dominance in recent years, with their long-range attention mechanisms proving particularly effective for temporal modeling. However, this advantage comes with substantial computational costs that create deployment challenges for resource-constrained applications.

\textbf{Qualitative Analysis.} Visual comparisons in Fig.~\ref{fig:vsr_reg_vis} provide complementary insights beyond numerical metrics. Regression-based approaches such as BasicVSR++~\cite{chan2022basicvsr++} and PSRT~\cite{shi2022rethinking} exhibit excellent temporal consistency and faithfully reconstruct fine details like the ``ZABITA" text sequences. In contrast, generation-based methods like MGLDVSR~\cite{yang2023mgldvsr} and UAV~\cite{zhou2024upscale}, while producing perceptually sharper results, frequently fail to accurately reconstruct critical information (\eg, the frame numbers ``153"/``155") and exhibit visible temporal instability. Transformer-based solutions, particularly RVRT~\cite{liang2022recurrent}, stand out by maintaining structural integrity across frames while achieving superior detail preservation compared to both CNN-based and generative approaches. All modern methods significantly outperform the bicubic baseline, though with distinct characteristic artifacts.

\textbf{Future Prospects.} The experimental results collectively suggest that VSR technology remains far from saturation, with multiple promising research directions. Improving computational efficiency represents a critical challenge, particularly for transformer architectures where memory and processing demands limit practical deployment. Alternative training paradigms that reduce dependence on massive labeled datasets could help democratize VSR development. Furthermore, hybrid approaches that combine the fidelity of regression methods with the perceptual strengths of generative techniques may help bridge the current quality gap. These advancements could potentially address the fundamental trade-offs between accuracy, efficiency, and visual quality that currently define the VSR landscape.}

\section{Stereo Super-resolution}
\label{sec:ssr}
\subsection{\textbf{Regressive Models}}
\subsubsection{\textbf{CNN-based SSR}}
\
\newline
\indent \textbf{Learning Better Representations:} Jeon~\etal~\cite{jeon2018enhancing} learned a parallax prior from stereo image datasets by jointly training two-stage networks for luminance and chrominance, respectively, and enhanced the spatial resolution of stereo images significantly more than SISR methods. Song~\etal~\cite{song2020stereoscopic} imposed stereo-consistency constraints on integrating stereo information. 
\begin{table*}[!t]
  \caption{\revision{Performance comparison of different SSR models on three benchmarks for the $\times 4$ task. PSNR/SSIM values on the Y channel are reported for each dataset. ``Params" represents the total number of network parameters. The \textbf{\textcolor{blue}{best}} performance is highlighted for models trained on Flickr1024~\cite{wang2019flickr1024}, while the \textbf{\textcolor{red}{best}} performance is highlighted for models trained on Middlebury~\cite{scharstein2014high} + Flickr1024~\cite{wang2019flickr1024}. The results for each model are obtained from their respective original papers.}}
  \centering
  \setlength{\tabcolsep}{3pt}{
  \begin{tabular}{c|c|ccc|c|c|c}
     \toprule[1.5pt]
     Model & \makecell[c]{Params\\(M)} & \makecell[c]{KITTI2015~\cite{mayer2016large}\\(PSNR/SSIM)} & \makecell[c]{KITTI2012~\cite{geiger2012we}\\(PSNR/SSIM)} &
     \makecell[c]{Middlebury~\cite{scharstein2014high}\\(PSNR/SSIM)} &  Training Sets & Backbone & Venue\\
     \midrule
     \hline
     \rowcolor{light-gray} Bicubic\qquad & - & 23.90/0.7100 & 24.64/0.7334 & 26.39/0.7564 &  - & - & -\\
     
     StereoSR~\cite{jeon2018enhancing} & 1.06 & 25.12/0.7679 & 25.94/0.7839 & 28.24/0.8133 & Middlebury+KITTI+Tsukuba & CNN & CVPR'18\\
     \hline
     \hline
     \rowcolor{light-gray} 
     DASSR~\cite{yan2020disparity} & 1.1 & 25.35/0.8740 & \textcolor{blue}{\textbf{26.96}}/\textcolor{blue}{\textbf{0.8820}} & \textcolor{blue}{\textbf{29.83}}/\textcolor{blue}{\textbf{0.9090}} & Flickr1024 & CNN & CVPR'20\\

    SRRes+SAM~\cite{ying2020stereo} & 1.73 & 25.55/0.7825 & 26.35/0.7957 & 28.76/0.8287 & Flickr1024 & CNN & SPL'20\\
     
     \rowcolor{light-gray} CVCNet~\cite{zhu2021cross} & 0.99 & 25.55/0.7801 & 26.35/0.7935 & 28.65/0.8231 & Flickr1024 & CNN & TMM'21\\
     
     SSRDE-FNet~\cite{dai2021feedback} & 2.24 & 25.74/0.7884 & 26.61/0.8028 & 29.29/0.8407 & Flickr1024 & CNN & ACMMM'21\\
     
     \rowcolor{light-gray} NAFSSR~\cite{chu2022nafssr} & 1.53 & \textcolor{blue}{\textbf{25.78}}/\textcolor{blue}{\textbf{0.7927}} & 26.62/0.8051 & 29.27/0.8447 & Flickr1024 & CNN & CVPRW'22\\
     
     Steformer~\cite{lin2023steformer} & 1.34 & 25.74/0.7906 & 26.61/0.8037 & 29.29/0.8424 & Flickr1024 & Transformer & TMM'23\\
     \hline
     \hline
     \rowcolor{light-gray} EDSR~\cite{Ledig2017} & 38.9 & 25.38/0.7811 & 26.26/0.7954 & 29.15/0.8383 & Middlebury+Flickr1024 & CNN & CVPR'17\\

     RCAN~\cite{zhang2018image} & 15.4 & 25.53/0.7836 & 26.36/0.7968 & 29.20/0.8381 & Middlebury+Flickr1024 & CNN & ECCV'18\\
     
     \rowcolor{light-gray} PASSRnet~\cite{wang2019learning} & 1.35 & 25.34/0.7722 & 26.18/0.7874 & 28.36/0.8153 & Middlebury+Flickr1024 & CNN & CVPR'19\\
     
     IMSSRnet~\cite{lei2020deep} & 6.89 & 25.59/- & 26.44/- & 29.02/- & Middlebury+Flickr1024 & CNN & TCSVT'20\\
     
     \rowcolor{light-gray} iPASSR~\cite{wang2021symmetric} & 1.42 & 25.61/0.7850 & 26.47/0.7993 & 29.07/0.8363 & Middlebury+Flickr1024 & CNN & CVPRW'21\\
     
     SIRFormer~\cite{yang2022sir} & 1.48 & 25.75/0.7882 & 26.53/0.7998 &  29.23/0.8396 & Middlebury+Flickr1024 & Transformer & ACMMM'22\\
     
     \rowcolor{light-gray} SwinFIRSSR~\cite{zhang2022swinfir} & 24.09 & 26.15/0.8062 & 27.06/0.8175 &  30.33/0.8676 & Middlebury+Flickr1024 & Transformer & ArXiv'22\\

     MSSFNet~\cite{gao2024learning} & 1.82 & 26.07/0.7990 & 26.88/0.8098 &  29.67/0.8498 & Middlebury+Flickr1024 & Transformer & ArXiv'24\\
     
     \rowcolor{light-gray} LSSR~\cite{gao2024learningMM} & 1.11 & 26.12/0.7997 & 26.93/0.8097 &  29.86/0.8489 & Middlebury+Flickr1024 & CNN & ACMMM'24\\

     SCGLANet~\cite{zhou2024toward} & 25.29 & \textcolor{red}{\textbf{26.94}}/\textcolor{red}{\textbf{0.8268}} & \textcolor{red}{\textbf{27.10}}/\textcolor{red}{\textbf{0.8204}} & \textcolor{red}{\textbf{30.23}}/\textcolor{red}{\textbf{0.8628}} & Middlebury+Flickr1024 & GAN & ESWA'24\\
     \hline
     \bottomrule
  \end{tabular}}
  \label{tab:ssr_results}
\end{table*}

\begin{figure}
\centering
    \begin{overpic}[scale=.25]{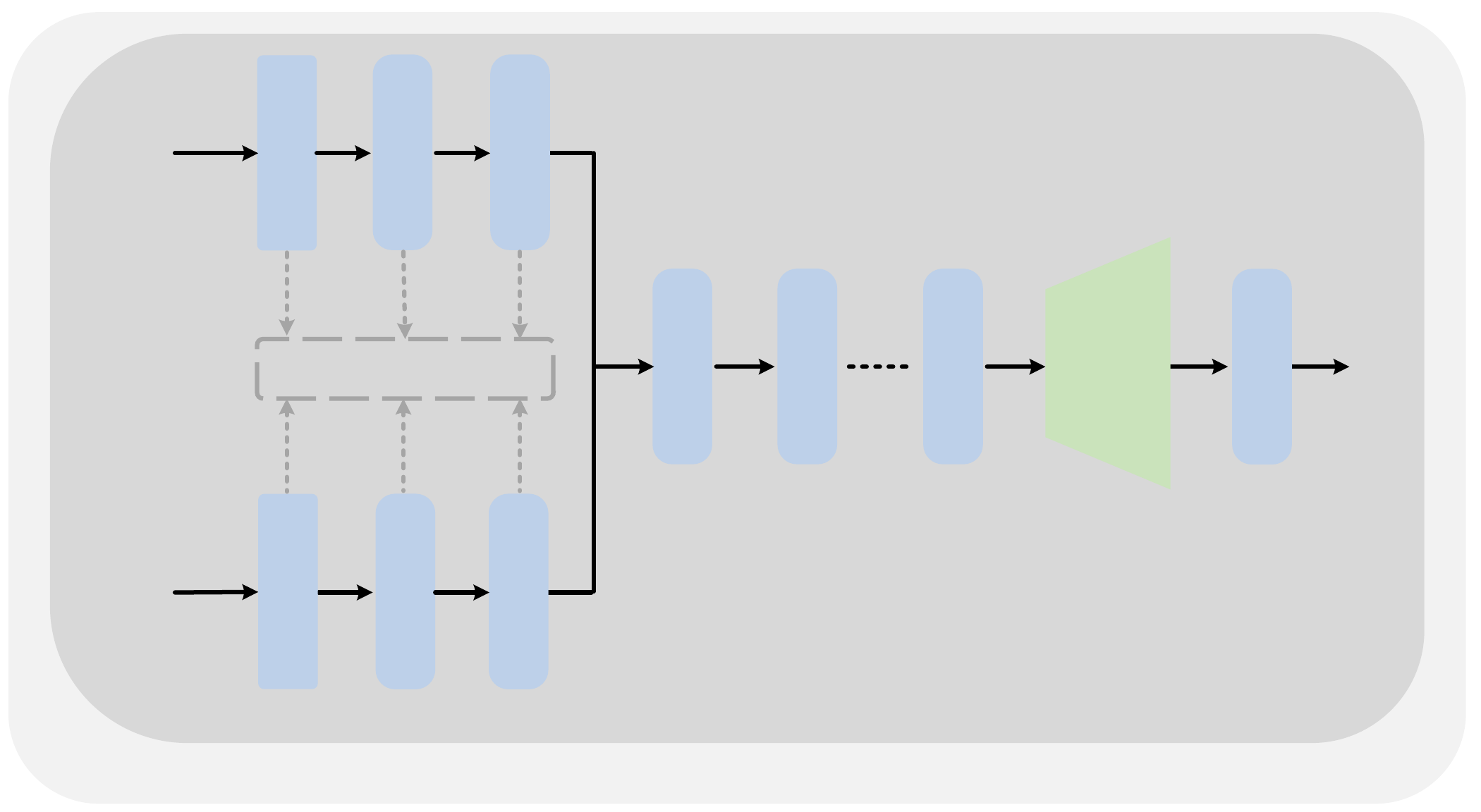}
        \put(4, 43.8){\footnotesize~$x_{left}$}
        \put(3, 14){\footnotesize~$x_{right}$}
        \put(90.5, 29.7){\footnotesize~$x_{sr}$}
        \put(16.8, 29){\scriptsize~Shared Weights}
        \put(40.5, 1.8){\scriptsize~PASSRnet~\cite{wang2019learning}}

        \put(17.3, 41.2){\scriptsize~\rotatebox{90}{Conv}}
        \put(25.3, 41.2){\scriptsize~\rotatebox{90}{Conv}}
        \put(33, 41.2){\scriptsize~\rotatebox{90}{Conv}}
        \put(17.3, 11.5){\scriptsize~\rotatebox{90}{Conv}}
        \put(25.3, 11.5){\scriptsize~\rotatebox{90}{Conv}}
        \put(33, 11.5){\scriptsize~\rotatebox{90}{Conv}}
        
        \put(44, 27){\scriptsize~\rotatebox{90}{Conv}}
        \put(52.3, 27){\scriptsize~\rotatebox{90}{Conv}}
        \put(62.3, 27){\scriptsize~\rotatebox{90}{Conv}}
        \put(83, 27){\scriptsize~\rotatebox{90}{Conv}}
        \put(73, 28.5){\scriptsize~\rotatebox{90}{UP}}
    \end{overpic}
    \caption{The framework of PASSRnet~\cite{wang2019learning}. It introduced a compelling and efficient parallax-attention mechanism, along with the creation of the extensive dataset Flickr1024, both of which have had a profound impact on subsequent SSR research. The results for each model are obtained from their respective original papers.}
\label{fig:stereosr}
\end{figure}
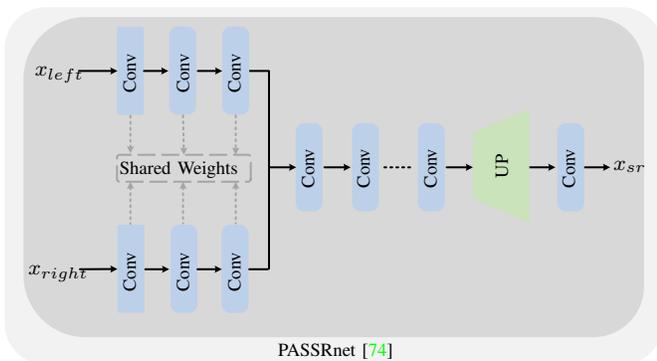

\textbf{Improving Information Interaction:} Wang~\etal~\cite{wang2019learning} introduced a generic parallax attention mechanism with a global receptive field along the epipolar line to handle different stereo images with large disparity variations, as shown in Fig.~\ref{fig:stereosr}. Ying~\etal~\cite{ying2020stereo} proposed a generic stereo attention module to extend arbitrary SISR networks for SSR. Zhang~\etal~\cite{zhang2020stereo} incorporated a disparity-based constraint mechanism into the generation of SR images with an additional atrous parallax-attention module. Yan~\etal~\cite{yan2020disparity} proposed a unified stereo image restoration framework that explicitly learns the inherent pixel correspondence between stereo views and restores stereo images by leveraging cross-view information at the image and feature level. Lei~\etal~\cite{lei2020deep} proposed an interaction module-based stereoscopic network to effectively utilize the correlation information in stereoscopic images. Wang~\etal~\cite{wang2021symmetric} proposed a symmetric bi-directional parallax attention module and an inline occlusion handling scheme to effectively interact with cross-view information. Chen~\etal~\cite{chen2021cross} proposed a cross parallax attention stereo network that can perform stereo SR for multiple scale factors and both views. Zhu~\etal~\cite{zhu2021cross} proposed a cross view capture network for stereo image super-resolution by using both global contextual and local features extracted from both views. Xu~\etal~\cite{xu2021deep} incorporated bilateral grid processing in a CNN framework and proposed a bilateral stereo network. Dan~\etal~\cite{dan2021disparity} proposed a disparity feature alignment module to exploit the disparity information for feature alignment and fusion. Dai~\etal~\cite{dai2021feedback} proposed a disparity estimation feedback network, which simultaneously handles stereo image super-resolution and disparity estimation in a unified framework and leverages their interaction to further improve performance. Chu~\etal~\cite{chu2022nafssr} inherited a strong and simple image restoration model, NAFNet, for single-view feature extraction and extended it by adding cross-attention modules to fuse features between views to adapt to binocular scenarios.
\subsubsection{\textbf{Transformer-based SSR}}
\
\newline
\indent Imani~\etal~\cite{imani2022new} designed a self-attention and optical flow-based feed-forward layer to make the transformer suitable for stereo video SR. Jin~\etal~\cite{jin2022swinipassr} adopted the swin transformer as the backbone and incorporated it with the bi-directional parallax attention module to maximize auxiliary information given the binocular mechanism. Lin~\etal~\cite{lin2023steformer} resorted to the transformer to capture reliable stereo information and proposed a cross-to-intra attention mechanism to further extract information. Yang~\etal~\cite{yang2022sir} leveraged long-range context dependencies and proposed the stereo alignment transformer to align the views, and the stereo fusion transformer to aggregate cross-view information. PFT-SSR~\cite{PFT-SSR-ICASSP} proposed a fusion strategy based on the transformer structures and designed two brances to explore the inter-view and the intra-view information. \revision{SIR-Former~\cite{yang2022sir} leverages a stereo alignment transformer (SAT) for global correspondence alignment and a stereo fusion transformer (SFT) for cross-view information aggregation, enhancing performance across various image restoration tasks.}
\subsection{\textbf{Generative Models}}
\
\newline
\revision{\indent Ma~\etal~\cite{ma2021perception} proposed a perception-oriented StereoSR framework to restore stereo images with better subjective quality by employing feedback from the quality assessment of the StereoSR results. Dinh~\etal~\cite{DSR2020} proposed a model to simultaneously solve denoising and SR problems and introduced matchability attention to improve the interaction between generated stereo images. RealSCGLAGAN~\cite{zhou2024toward} is a novel stereo image super-resolution method that integrates an implicit stereo information discriminator and a hybrid degradation model to enhance visual quality while preserving disparity consistency. DiffSteISR~\cite{zhou2024diffsteisr} is a pioneering stereo image reconstruction framework that integrates techniques like time-aware stereo cross attention, omni attention control, and stereo semantic extraction to enhance texture and semantic consistency in super-resolved stereo images.}
\subsection{\textbf{Training Details}}
\subsubsection{\textbf{Datasets}}
\
\newline
\indent There are five commonly used datasets for evaluating the performance of SSR methods: Middlebury~\cite{scharstein2014high}, Tsukuba~\cite{peris2012towards}, KITTI~2012~\cite{geiger2012we}, KITTI 2015~\cite{mayer2016large}, and Flickr1024~\cite{wang2019flickr1024}.
\begin{figure*}[!th]
    \centering
    \begin{overpic}[scale=.2]{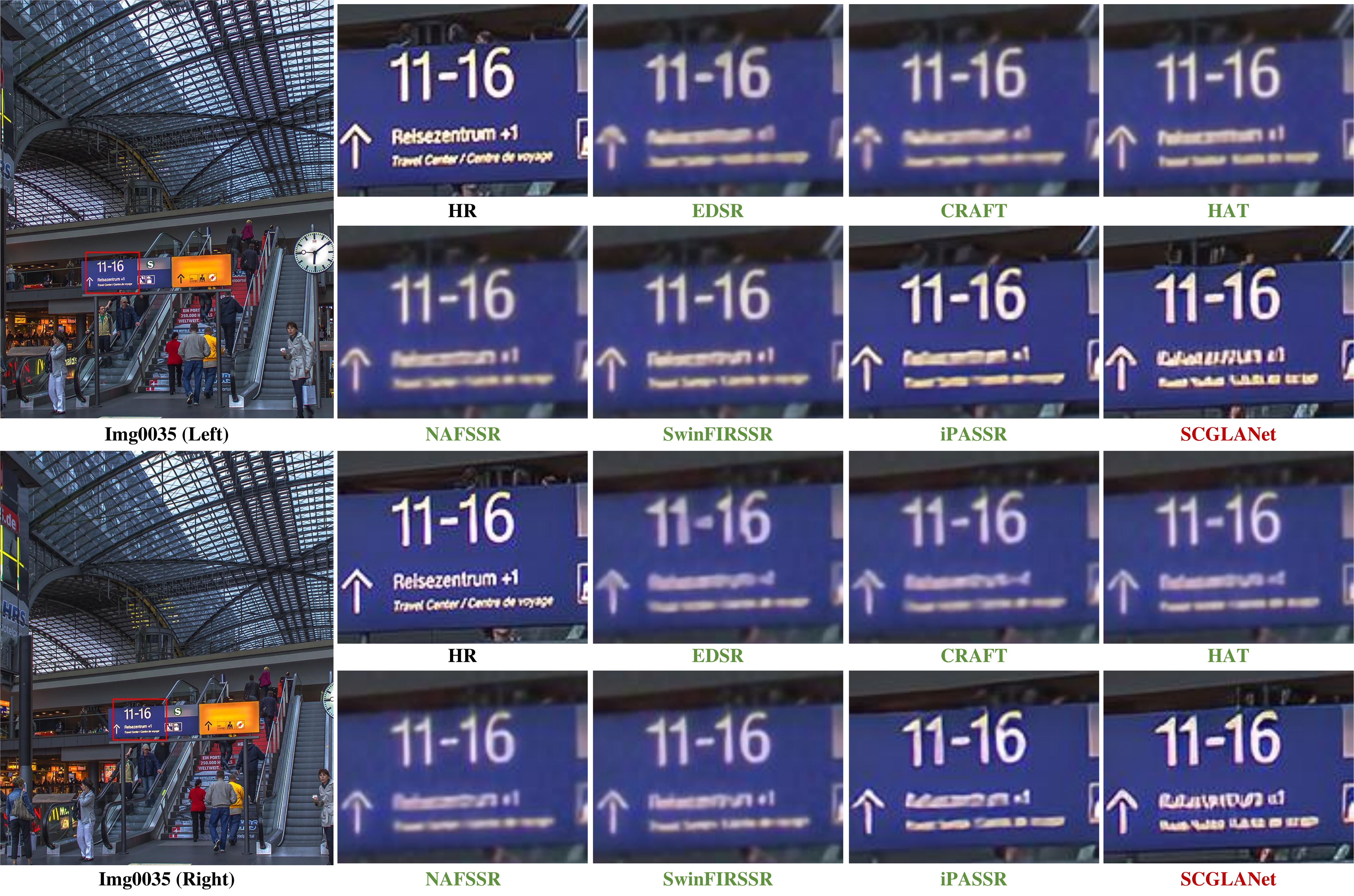}
        \put(36.2, 0.8){\scriptsize~~\cite{chu2022nafssr}}
        \put(56.2, 0.8){\scriptsize~~\cite{zhang2022swinfir}}
        \put(73.7, 0.8){\scriptsize~~\cite{wang2021symmetric}}
        \put(93.6, 0.8){\scriptsize~~\cite{zhou2024toward}}

        \put(54.2, 17.3){\scriptsize~~\cite{Ledig2017}}
        \put(73.7, 17.3){\scriptsize~~\cite{li2023feature}}
        \put(91.5, 17.3){\scriptsize~~\cite{chen2022activating}}

        \put(36.2, 33.7){\scriptsize~~\cite{chu2022nafssr}}
        \put(56.2, 33.7){\scriptsize~~\cite{zhang2022swinfir}}
        \put(73.7, 33.7){\scriptsize~~\cite{wang2021symmetric}}
        \put(93.6, 33.7){\scriptsize~~\cite{zhou2024toward}}

        \put(54.2, 50.1){\scriptsize~~\cite{Ledig2017}}
        \put(73.7, 50.1){\scriptsize~~\cite{li2023feature}}
        \put(91.5, 50.1){\scriptsize~~\cite{chen2022activating}}
    \end{overpic}
    
    \caption{\resubmit{Visual comparison of \textcolor{regressive}{\textbf{regression-based}} and \textcolor{generative}{\textbf{generation-based}} stereo super-resolution methods. \textcolor{regressive}{\textbf{Regression-based}} methods such as EDSR~\cite{Ledig2017}, CRAFT~\cite{li2023feature}, and NAFSSR~\cite{chu2022nafssr} generally succeed in making the primary text ``11-16" legible and attempt to reconstruct the finer``Reisezentrum +1" text, albeit with noticeable blur. Conversely, the \textcolor{generative}{\textbf{generation-based}} method SCGLANet~\cite{zhou2024toward}, while also clearly rendering ``11-16", tends to introduce significant artifacts and unfaithful, sharp textures in place of the smaller textual details, often rendering them illegible and unnatural compared to the HR or even the \textcolor{regressive}{\textbf{regression-based}} outputs.}}
    \label{fig:ssr_reg_vis}
\end{figure*}
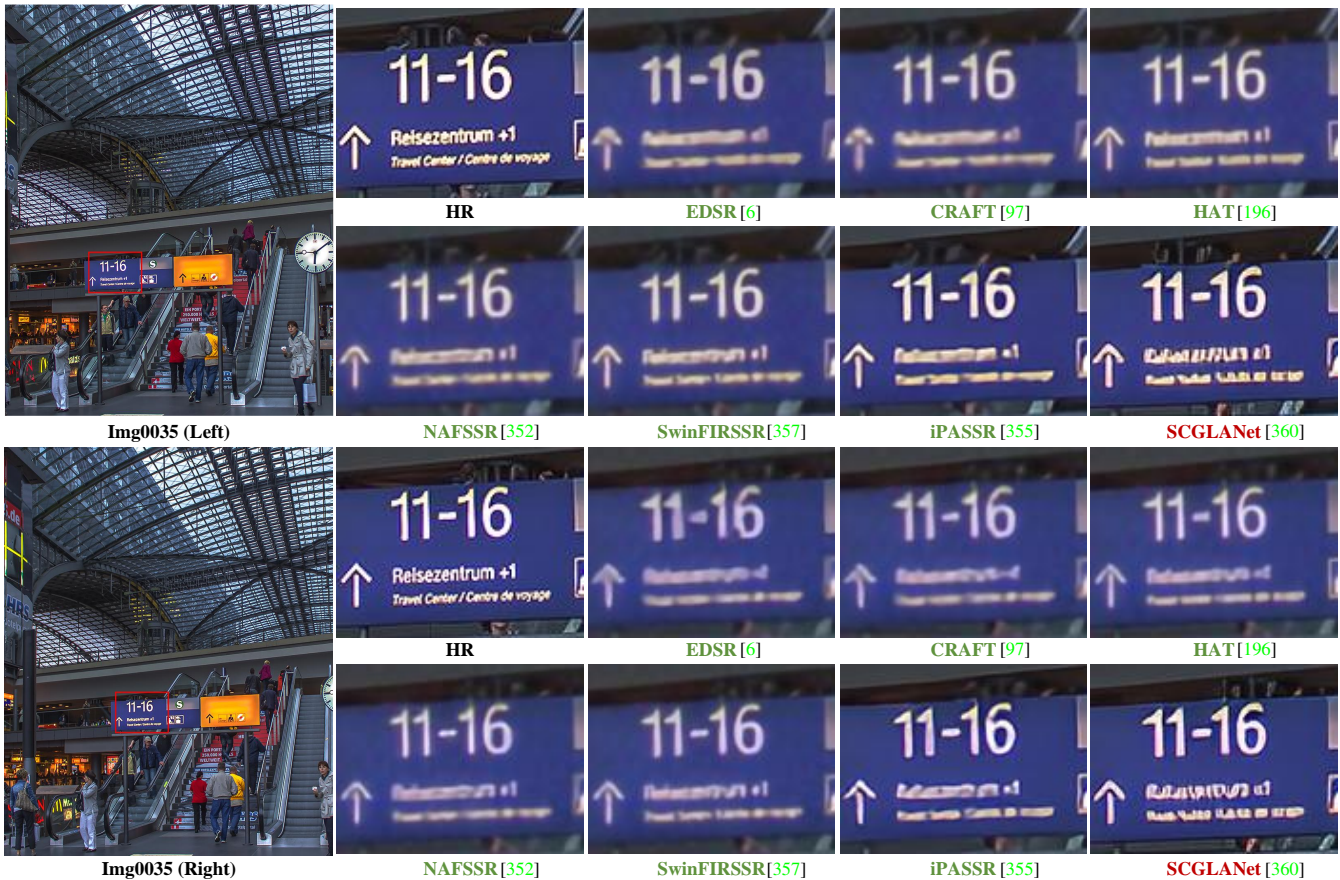
\begin{itemize}
\item Middlebury~\cite{scharstein2014high}: This dataset consists of input images captured under multiple exposures and ambient illuminations, with and without a mirror sphere present to capture the lighting conditions.
\item Tsukuba~\cite{peris2012towards}: The dataset contains 1800 full-color stereo pairs per illumination condition, along with ground truth disparity maps. It was created using an animated stereo camera and includes 1-minute videos at 30FPS.
\item KITTI 2012~\cite{geiger2012we}: This dataset selects a subset of sequences where the environment is static. It includes 194 training and 195 test image pairs.
\item KITTI 2015~\cite{mayer2016large}: It consists of stereo videos of road scenes captured by a calibrated pair of cameras mounted on a car. Ground truth for optical flow and disparity is obtained from a 3D laser scanner combined with the egomotion data of the car.
\item Flickr1024~\cite{wang2019flickr1024}: This is the largest dataset for stereo SR, containing 1024 high-quality image pairs that cover diverse scenarios. The dataset provides realistic cases that align with daily photography scenarios.
\end{itemize}
\begin{figure*}
    \centering
    \begin{overpic}[scale=.25, tics=5]{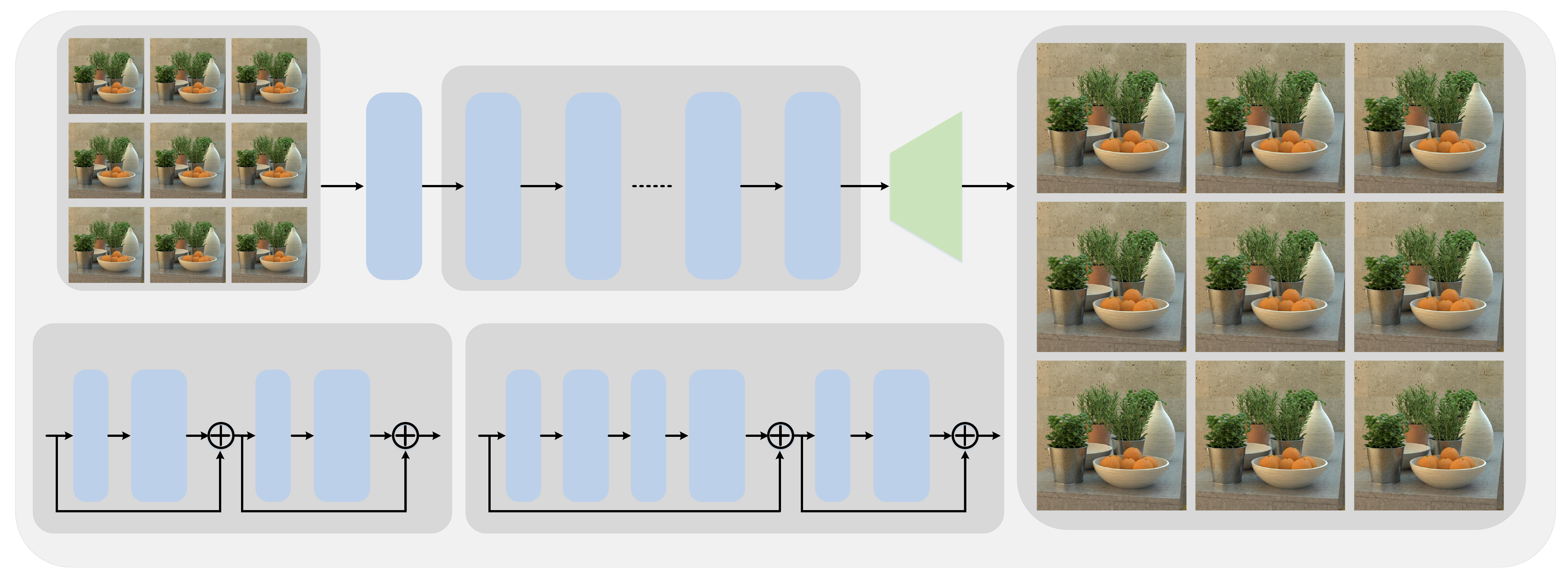}
        \put(4.8, 16.8){\scriptsize~LR Light Field Images}
        \put(3, 14.6){\scriptsize~Angular Transformer}
        \put(31, 14.6){\scriptsize~Spatial Transformer}
        \put(70.6, 1.4){\scriptsize~Super-resolved Light Field Images}

        \put(24.1, 23){\scriptsize~\rotatebox{90}{Conv}}
        \put(30.6, 19.2){\fontsize{6pt}{14pt}\selectfont~\rotatebox{90}{Angular Transformer}}
        \put(37.2, 19.5){\fontsize{6pt}{14pt}\selectfont~\rotatebox{90}{Spatial Transformer}}
        \put(44.7, 19.2){\fontsize{6pt}{14pt}\selectfont~\rotatebox{90}{Angular Transformer}}
        \put(50.9, 19.5){\fontsize{6pt}{14pt}\selectfont~\rotatebox{90}{Spatial Transformer}}
        \put(58.5, 23.8){\scriptsize~\rotatebox{90}{UP}}

        \put(5, 5.7){\fontsize{6pt}{14pt}\selectfont~\rotatebox{90}{Layer Norm}}
        \put(8.5, 5.2){\fontsize{6pt}{12pt}\selectfont~\rotatebox{90}{\shortstack{Multi-head\\Self-Attention}}}
        \put(16.7, 5.7){\fontsize{6pt}{14pt}\selectfont~\rotatebox{90}{Layer Norm}}
        \put(20.2, 5.2){\fontsize{6pt}{14pt}\selectfont~\rotatebox{90}{\shortstack{Pointwise\\Feedforward}}}

        \put(32.6, 6){\fontsize{6pt}{14pt}\selectfont~\rotatebox{90}{Unfolding}}
        \put(35.7, 6.2){\fontsize{5.5pt}{10pt}\selectfont~\rotatebox{90}{\shortstack{Multi-layer\\Perception}}}
        \put(40.5, 5.7){\fontsize{6pt}{14pt}\selectfont~\rotatebox{90}{Layer Norm}}
        \put(44, 6.2){\fontsize{6pt}{14pt}\selectfont~\rotatebox{90}{\shortstack{window\\Attention}}}
        \put(52.2, 5.7){\fontsize{6pt}{14pt}\selectfont~\rotatebox{90}{Layer Norm}}
        \put(56, 5.2){\fontsize{6pt}{14pt}\selectfont~\rotatebox{90}{\shortstack{Pointwise\\Feedforward}}}
        
    \end{overpic}
    \caption{The framework of LFT~\cite{liang2022light}. It introduced a straightforward yet highly efficient transformer-based model that incorporated an angular transformer block and a spatial transformer block. This design enabled the extraction of abundant non-local information, ultimately resulting in exceptional performance and establishing a new state-of-the-art.}
    \label{fig:lfsr}
\end{figure*}
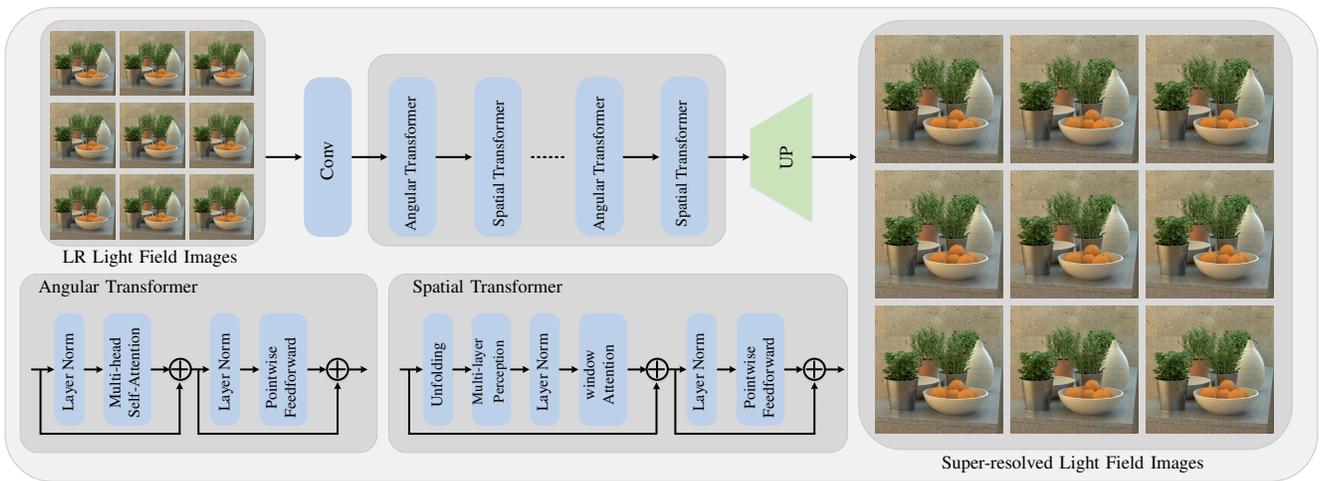
\subsubsection{\textbf{Training Settings}}
\
\newline
\indent\textbf{Patch Extraction:} In stereo super-resolution (SSR), patch extraction involves dividing low-resolution stereo image pairs into smaller fixed-size patches, usually ranging from $32\times32$ to $64\times64$ pixels. For instance, the StereoLR dataset divides each LR stereo image pair into $30K$ patches of size $41\times41$ pixels, creating training samples for the super-resolution model.

\textbf{Data Augmentation:} To increase the diversity of training samples, data augmentation is applied to both the left and right images in stereo pairs. Common techniques, such as random rotations, flips, color transformations, and additional strategies like random horizontal shifting, Mixup, CutMix, and CutMixup, are employed to further enhance the model's performance.

\textbf{Optimization:} Standard optimization algorithms like SGD, Adam, and RMSprop are commonly used to iteratively update the model weights and minimize the disparity error. For instance, a recent stereo super-resolution method used the Adam optimizer with a learning rate of $10^{-4}$ for $100$ epochs, leading to an average disparity error reduction of $25\%$.

\textbf{Learning Schedule:} A typical learning rate schedule involves starting with an initial learning rate of $10^{-3}$ and then reducing it by a factor of $10^{-1}$ after every $10K$ training iterations. This facilitates effective model convergence and improves accuracy and efficiency.

\textbf{Batch Size:} Batch sizes in SSR typically range from $4$ to $32$, with smaller batch sizes employed in memory-constrained scenarios or when dealing with large-resolution stereo images. A common choice is a batch size of $8$, which strikes a balance between computational efficiency and memory requirements. These training settings are crucial for achieving high-quality results in this field.

\resubmit{\subsection{\textbf{Experimental Results}}
We conduct comprehensive experiments to evaluate the performance of SSR methods in comparison with traditional SISR approaches. The quantitative results are summarized in Table~\ref{tab:ssr_results}, which includes standard evaluation metrics (PSNR, SSIM), model complexity (number of parameters), training datasets, network architectures, and publication venues.

The experimental results demonstrate clear advantages of dedicated SSR methods over traditional SISR approaches when processing stereo imagery. As shown in Table~\ref{tab:ssr_results}, while both types of methods can achieve comparable performance in terms of standard metrics like PSNR and SSIM, the stereo-specific designs demonstrate superior capability in maintaining view consistency and handling disparity variations. Traditional SISR methods such as EDSR~\cite{Ledig2017} and RCAN~\cite{zhang2018image}, despite their good performance on single-image tasks, exhibit significant limitations when directly applied to stereo pairs. These approaches process left and right views independently, failing to exploit the inherent correlation between stereo images. In contrast, modern SSR methods like NAFSSR~\cite{chu2022nafssr} and SwinFIRSSR~\cite{zhang2022swinfir} explicitly address these challenges through specialized architectures. They incorporate stereo matching components and disparity-aware processing modules that maintain view consistency during super-resolution. Visualization results, as shown in Fig.~\ref{fig:ssr_reg_vis}, particularly highlight this advantage. Specifically, we visualize the outputs of CNN and Transformer-based SISR methods. From these results, SISR methods (EDSR~\cite{Ledig2017} and HAT~\cite{chen2022activating}) produce more artifacts and blurriness. Although Transformer-based methods like HAT possess strong long-term dependency capture capabilities, their outputs still exhibit some artifacts. In contrast, SSR methods produce more coherent and geometrically consistent high-resolution outputs compared to their SISR counterparts. In addition, it can be observed that generative models (\eg, SCGLANet~\cite{zhou2024toward}) can produce more perceptually pleasant results than regression-based models.}

\section{Light Field Image Super-resolution}
\label{sec:lfsr}
Light-field images capture both the spatial and angular information of light rays, providing a richer representation of the scene compared to traditional 2D images. This additional angular information enables post-capture image manipulation such as refocusing, depth estimation, and perspective changes. However, the resolution of light-field images is typically lower due to hardware limitations, necessitating the use of super-resolution techniques to enhance both spatial and angular resolutions simultaneously.
\subsection{\textbf{Regressive Models}}
\subsubsection{\textbf{CNN-based LFSR}}
\
\newline
\indent \textbf{Efficient Structural Designs:} LFCNN~\cite{yoon2015learning} enhances both spatial and angular resolutions of Light-Field images using a deep convolutional neural network, by first applying spatial super-resolution to sub-aperture images and then generating novel views with angular super-resolution, culminating in end-to-end fine-tuning. Yoon~\etal~\cite{yoon2017light} introduced a method that uses CNNs to simultaneously up-sample the spatial and angular resolutions of light field images. Yuan~\etal~\cite{yuan2018light} presented a combined deep CNN framework that incorporates a SISR module and an epipolar plane image (EPI) enhancement module to achieve geometrically consistent super-resolved light field images. Wing~\etal~\cite{yeung2018light} proposed an hourglass-shaped model that performs feature extraction at the low-resolution level to reduce computation and memory costs. They utilized spatial-angular separable convolutions for efficient extraction of joint spatial-angular features. Zhu~\etal~\cite{zhu2019revisiting} developed an EPI-based CNN-LSTM network specifically designed for simultaneous spatial and angular super-resolution of light fields. Wang~\etal~\cite{wang2020spatial} introduced an approach to extract and decouple spatial and angular features. Meng~\etal~\cite{meng2019high} formulated LFSR as tensor restoration and developed a two-stage restoration framework based on 4-dimensional convolution. Cheng~\etal~\cite{cheng2021light} introduced a zero-shot learning framework for LFSR, where the reference view is super-resolved solely based on examples extracted from the input low-resolution light field. Wang~\etal~\cite{wang2022disentangling} developed a mechanism to disentangle coupled information in light field image processing, leading to the design of three networks (DistgSSR, DistgASR, and DistgDisp) for spatial super-resolution, angular super-resolution, and disparity estimation, respectively. These networks leverage the disentangling mechanism for improved performance in LFSR tasks. Jin~\etal~\cite{jin2020light} individually super-resolved each view of a light field image by leveraging complementary information among views through combinatorial geometry embedding. Jin~\etal~\cite{Jin2020} utilized depth estimation and light field blending modules to leverage the intrinsic geometry information. They also introduced a loss based on the gradient of epipolar-plane images to preserve the light field parallax structure. Zhang~\etal~\cite{zhang2021end} proposed an end-to-end LFSR framework called MEG-Net, which utilizes multiple epipolar geometries to simultaneously super-resolve all views in a light field by exploring sub-pixel information in different spatial and angular directions.

\textbf{Improving Information Interaction:} Wang~\etal~\cite{wang2018lfnet} proposed an implicitly multi-scale fusion scheme for light field super-resolution by accumulating contextual information from multiple scales. Zhang~\etal~\cite{zhang2019residual_lfsr} introduced residual convolutional networks for reconstructing high-resolution light fields. Cheng~\etal~\cite{cheng2019light} proposed a framework for LFSR that exploits both internal and external similarities, demonstrating their complementary nature. Liu~\etal~\cite{liu2021intra} proposed LF-IINet, an intra-inter view interaction network consisting of two parallel branches that model global inter-view information and correlations among intra-view features, respectively. Wang~\etal~\cite{wang2020light} designed an angular deformable alignment module for feature-level alignment and proposed a collect-and-distribute approach for bidirectional alignment between the center-view feature and each side-view feature. \revision{Meng~\etal~\cite{Meng2020} explored the geometric structural properties in light field data and proposed a second-order residual network to learn geometric features for reconstruction. LFSR-AFR~\cite{ko2021light} improves Light-Field image super-resolution by utilizing spatial and angular SR networks with a trainable disparity estimator and an adaptive feature remixing (AFR) module to enhance multi-view features, achieving superior performance on various datasets. DDAN~\cite{mo2021dense} enhances Light-Field image super-resolution by employing a dense dual-attention network that adaptively captures and fuses discriminative features across different views and channels, leveraging a chain structure and dense connections to improve SR performance.} Xiao~\etal~\cite{xiao2023cutmib} conducted a series of pre-processing steps involving blending crop patches from each view of LR/HR to produce a blended patch and pasting it to the corresponding position of HR/LR. This approach achieves improved SR results while keeping the model unchanged.

\subsubsection{\textbf{Transformer-based LFSR}}
\
\newline
\indent Liang~\etal~\cite{liang2022light} designed an angular transformer and a spatial transformer to model the relationship among different views and capture local and non-local context information, s, as shown in Fig.~\ref{fig:lfsr}. Wang~\etal~\cite{wang2022detail} treated LFSR as a sequence learning problem and proposed a detail-preserving transformer that uses gradient maps of light fields to guide the sequence learning. Shabbir~\etal~\cite{shabbir2022learning} introduced a transformer module to extract texture features from the reference image in a four-step reconstruction process. Guo~\etal~\cite{guo2022light} focused on data generation and proposed a new raw data generation method, followed by the use of transformer models to aggregate angular and non-local cross-view information. Wang~\etal~\cite{wang2022multi} proposed intro- and inter-transformers to dynamically exploit information between sub-aperture images. EPIT~\cite{liang2023learning} takes into account the spatial and angular correlation within the EPI and introduces a transformer-based structure to capture this relationship.
\begin{table*}[!t]
  \caption{\revision{Performance comparison of different LFSR models on five benchmarks for the $\times 4$ task. PSNR/SSIM values on the Y channel are reported for each dataset. ``Params" represents the total number of network parameters. The \textbf{\textcolor{red}{best}} and \textcolor{red}{\underline{second best}} performances are indicated in bold and underlined.}}
  \centering
  \setlength{\tabcolsep}{5pt}{
  \begin{tabular}{c|c|ccccc|c|c}
     \toprule[1.5pt]
     Model & \makecell[c]{Params\\(M)} & \makecell[c]{EPFL\\(PSNR/SSIM)} & \makecell[c]{HCInew\\(PSNR/SSIM)} & \makecell[c]{HCIold\\(PSNR/SSIM)} & \makecell[c]{INRIA\\(PSNR/SSIM)} & \makecell[c]{STFgantry\\(PSNR/SSIM)} & Backbone & Venue\\
     \midrule
     \hline
     \rowcolor{light-gray} Bicubic\qquad   & - & 25.14/0.8324 & 27.61/0.8517 & 32.42/0.9344 & 26.82/0.8867 & 25.93/0.8452 & - & -\\
     VDSR~\cite{kim2016accurate} &  0.66 & 27.25/0.8777 & 29.31/0.8823 & 34.81/0.9515 & 29.19/0.9204 & 28.51/0.9009 & CNN & CVPR’16\\
     \rowcolor{light-gray} EDSR~\cite{Lim2017} &  38.9 & 27.84/0.8854 & 29.60/0.8869 & 35.18/0.9536 & 29.66/0.9257 & 28.70/0.9072 & CNN & CVPR’17\\
     RCAN~\cite{zhang2018image} &  15.4 & 27.88/0.8863 & 29.63/0.8886 & 35.20/0.9548 & 29.76/0.9276 & 28.90/0.9131 & CNN & ECCV’18\\
     \rowcolor{light-gray} DDAN~\cite{mo2021dense} &  0.51 & 29.19/0.9178 & 31.60/0.9211 & 37.36/0.9722 & 31.44/0.9513 & 30.72/0.9419 & CNN & TCSVT’18\\

     LFSSR~\cite{yeung2018light} & 1.77 & 28.27/0.9118 & 30.72/0.9145 & 36.70/0.9696 & 30.31/0.9467 & 30.15/0.9426 & CNN & TIP'18\\
     
     \rowcolor{light-gray}resLF~\cite{zhang2019residual_lfsr} & 8.64 & 28.27/0.9035 & 30.73/0.9107 & 36.71/0.9682 & 30.34/0.9412 & 30.19/0.9372 & CNN & CVPR'19\\
     
     LF-ATO~\cite{jin2020light} & 1.36 & 28.25/0.9115 & 30.88/0.9135 & 37.00/0.9699 & 30.71/0.9484 & 30.61/0.9430 & CNN & CVPR'20\\
     \rowcolor{light-gray}LF-InterNet~\cite{wang2020spatial} & 5.48 & 28.67/0.9162 & 30.98/0.9161 & 37.11/0.9716 & 30.64/0.9491 & 30.53/0.9409 & CNN & ECCV'20\\
     
     LF-DFnet~\cite{wang2020light} & 3.99 & 28.77/0.9165 & 31.23/0.9196 & 37.32/0.9718 & 30.83/0.9503 & 31.15/0.9494 & CNN & TIP'20\\
     
     \rowcolor{light-gray}LF-IINet~\cite{liu2021intra} & 4.89 & 29.11/0.9188 & 31.36/0.9208 & 37.62/0.9734 & 31.08/0.9515 & 31.21/0.9502 & CNN & TMM'21\\
     
     MEGNet~\cite{zhang2021end} & 1.77 & 28.74/0.9160 & 31.10/0.9177 & 37.28/0.9716 & 30.66/0.9490 & 30.77/0.9453 & CNN & TIP'21\\
     
     \rowcolor{light-gray}DistgSSR~\cite{wang2022disentangling} & 3.58 & 28.99/0.9195 & 31.38/0.9217 & 37.56/0.9732 & 30.99/0.9519 & 31.65/0.9535 & CNN & TPAMI'22\\
     
     DPT~\cite{wang2022detail} & 3.78 & 28.93/0.9170 & 31.19/0.9188 & 37.39/0.9721 & 30.96/0.9503 & 31.14/0.9488 & Transformer & AAAI'22\\
     \rowcolor{light-gray}LFT~\cite{liang2022light} & 1.16 & 29.25/\textcolor{red}{\underline{0.9210}} & \textcolor{red}{\underline{31.46}}/\textcolor{red}{\underline{0.9218}} & \textcolor{red}{\underline{37.63}}/\textcolor{red}{\underline{0.9735}} & 31.20/0.9524 & \textcolor{red}{\underline{31.86}}/\textcolor{red}{\underline{0.9548}} & Transformer & SPL'22\\
     LFSAV~\cite{cheng2022spatial} & 1.54 & \textcolor{red}{\textbf{29.37}}/\textcolor{red}{\textbf{0.9223}} & 31.45/0.9217 & 37.50/0.9721 & \textcolor{red}{\underline{31.27}}/\textcolor{red}{\textbf{0.9531}} & 31.36/0.9505 & Transformer & TCI'22\\
     \rowcolor{light-gray}EPIT~\cite{liang2023learning} & 1.47 & \textcolor{red}{\underline{29.34}}/0.9197 & \textcolor{red}{\textbf{31.51}}/\textcolor{red}{\textbf{0.9231}} & \textcolor{red}{\textbf{37.68}}/\textcolor{red}{\textbf{0.9737}} & \textcolor{red}{\textbf{31.37}}/\textcolor{red}{\underline{0.9526}} & \textcolor{red}{\textbf{32.18}}/\textcolor{red}{\textbf{0.9571}} & Transformer & ICCV'23\\
     \hline
     \bottomrule
  \end{tabular}}
  \label{tab:lfsr_results}
\end{table*}
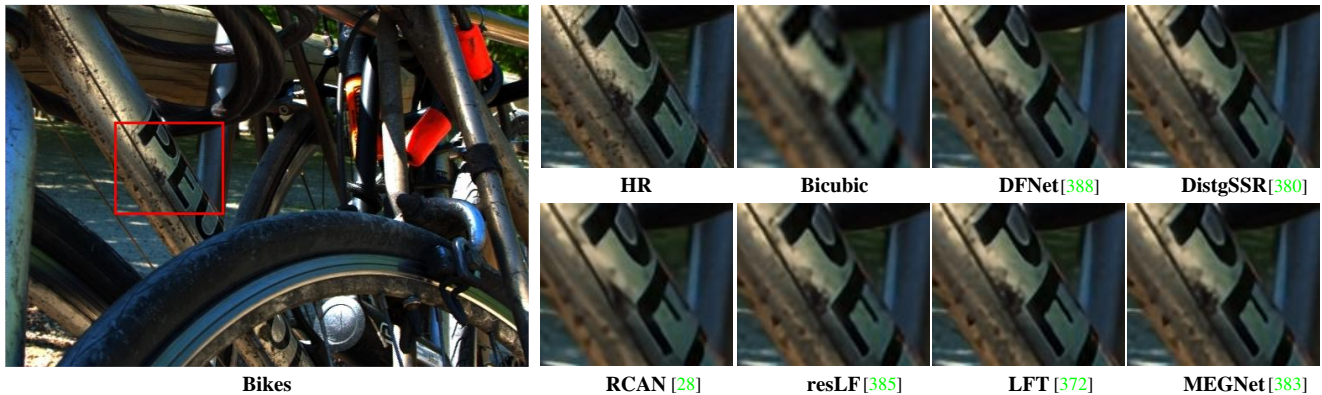
\begin{figure*}[!th]
    \centering
    \begin{overpic}[scale=.22]{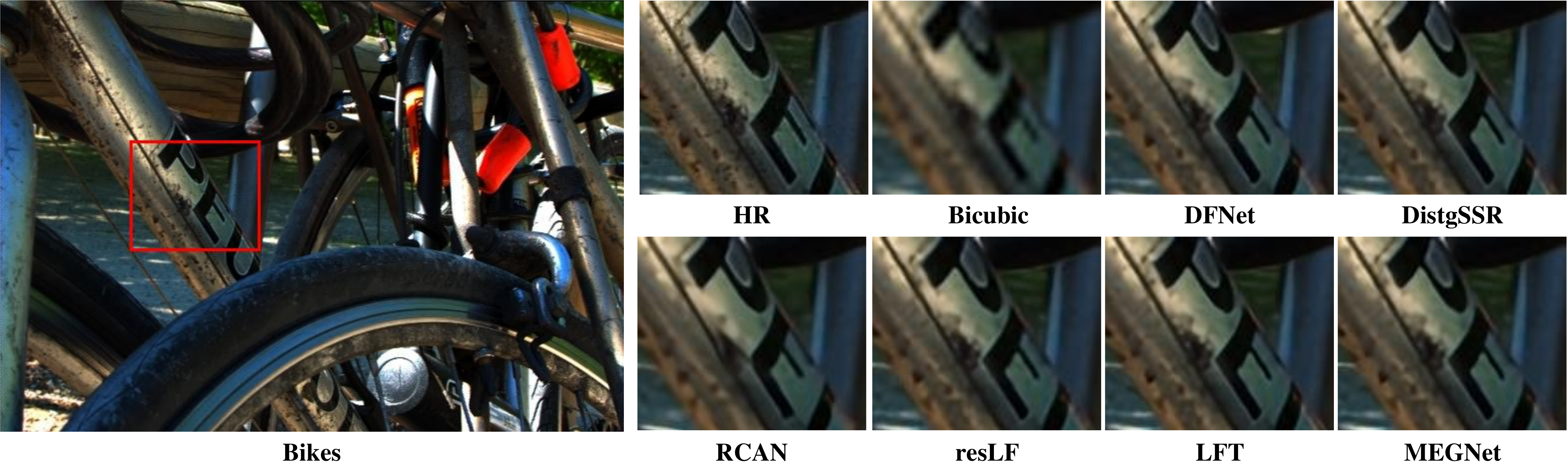}
        \put(49.6, 0.9){\scriptsize~~\cite{zhang2018image}}
        \put(64.2, 0.9){\scriptsize~~\cite{zhang2019residual_lfsr}}
        \put(78.7, 0.9){\scriptsize~~\cite{liang2022light}}
        \put(95, 0.9){\scriptsize~~\cite{zhang2021end}}

        \put(79.2, 16){\scriptsize~~\cite{wang2020light}}
        \put(95, 16){\scriptsize~~\cite{wang2022disentangling}}

    \end{overpic}
    \caption{\resubmit{Visual comparison of light field super-resolution methods. Dedicated light field super-resolution methods (like LFT~\cite{liang2022light} or MEGNet~\cite{zhang2021end}) generally achieve sharper results and more accurate texture reconstruction by leveraging the multi-view information inherent in the light field data. In contrast, traditional single-image methods like RCAN~\cite{zhang2018image}, while an improvement over bicubic, cannot fully exploit these angular cues, resulting in comparatively less detailed and often blurrier outputs in this context. }}
    \label{fig:lfsr_reg_vis}
\end{figure*}
\subsection{\textbf{Generative Models}}
Ko~\etal~\cite{ko2018gan} employed a GAN-based structure to generate finer texture details in skin for high up-scaling factors. They introduced content and adversarial loss to enhance image quality. Ruan~\etal~\cite{ruan2018light} proposed a wasserstein generative adversarial network with gradient penalty to further improve generative qualities. Additionally, they introduced perceptual loss on the epipolar plane to enhance depth information generation. Zhu~\etal~\cite{Zhu2019lfsr} proposed an auto-encoder structure to extract neighboring view information and employed a discriminator model with WGAN loss to enhance sharpness in reconstructed light fields. Chen~\etal~\cite{Chen2020lfgan} generated a 4D LF from a single image using GAN and devised a multi-stage training strategy combined with multiple losses to constrain the training process. Meng~\etal~\cite{meng2019spatial} introduced a generative model called LFGAN, which incorporated high dimensional convolution layers into a GAN framework to learn high correlations among neighboring LF views. Guo~\etal~\cite{guo2021real} designed a strategy for selecting foreground Elemental Images (EIs) to improve rendering performance and utilized a GAN network to individually super-resolve the EIs for better image quality. Wafa~\etal~\cite{wafa2022light} proposed a residual dense structure as the generator and introduced a patch discriminator to enhance visual quality. \revision{LFSRDiff~\cite{chao2023lfsrdiff} leverages a disentangled UNet to effectively capture and fuse spatial and angular information, producing diverse and realistic SR outcomes.}
\subsection{\textbf{Training Details}}
\subsubsection{\textbf{Datasets}}
\
\newline
\indent Five commonly used datasets are employed for training and evaluating the performance of LFSR methods: EPFL~\cite{rerabek2016new}, HCInew~\cite{honauer2016dataset}, HCIold~\cite{wanner2013datasets}, INRIA~\cite{le2018light}, and STFgantry~\cite{vaish2008new}.

\begin{itemize}
\item EPFL~\cite{rerabek2016new}: This dataset contains images in the LFR (Light Field Raw) file format captured by the Lytro Illum camera. The LFR files store raw, uncompressed lenslet images before the demosaicing process, with 10 bits per pixel precision in little-endian format.
\item HCInew~\cite{honauer2016dataset}: It consists of 24 carefully designed synthetic densely sampled 4D light fields with highly accurate disparity ground truth.
\item HCIold~\cite{wanner2013datasets}: This dataset is characterized by a dense sampling of light fields, which is suitable for current plenoptic cameras.
\item INRIA~\cite{le2018light}: The synthetic light fields ``butterfly" and ``stillife" are taken from the HCI database. The others were captured with a Lytro camera, and the sub-aperture images were extracted using the toolbox of Dansereau~\etal~\cite{dansereau2013decoding}.
\item STFgantry~\cite{vaish2008new}: The Computer Graphics Laboratory at Stanford University has acquired several light fields for research in computer graphics and vision.
\end{itemize}

\subsubsection{\textbf{Training Settings}}
\
\newline
\indent\textbf{Patch Extraction:} In LFSR, patch extraction involves dividing the 4D light field data into smaller patches to create training samples. Spatial patches of size, \eg, $32\times32$, are commonly extracted from each view, and angular patches of size, \eg, $9\times9$ views, are sampled to account for the multi-view nature. This process results in a large number of training samples, contributing to the effectiveness of training light field super-resolution models.

\textbf{Data Augmentation:} Commonly used transformations include random shifts, rotations, and flips applied to the $4D$ light field data. For instance, on the EPFL dataset, data augmentation can increase the number of training samples from $500$ to over $5,000$ by applying random shifts and flips in both spatial and angular dimensions.

\textbf{Optimization:} In LFSR, popular optimizers, such as Adam and SGD, are commonly used for parameter updates during training. An average learning rate of Adam is set to $10^{-3}$ to $10^{-4}$, while SGD is used with an average learning rate of $10^{-2}$ to $10^{-3}$.

\textbf{Learning Schedule:} A common learning rate schedule for LFSR models starts with an initial learning rate of $10^{-3}$ and uses a step decay strategy, reducing the learning rate by a factor of $10^{-1}$ after every $50,000$ iterations. This approach helps achieve convergence and avoid overfitting while training LFSR models on datasets like EPFL with an average training time of $150,000$ iterations.

\textbf{Batch Size:} Due to the high dimensionality of light field data, memory limitations, and computational complexity, the batch size is typically smaller compared to single image super-resolution models. Common batch sizes for light field super-resolution range from $2$ to $8$ light field samples per iteration.
\resubmit{\subsection{\textbf{Experimental Results and Analysis}}
We evaluate the competing methods on five widely-used benchmark datasets, assessing their performance using PSNR and SSIM metrics while analyzing model complexity through parameter counts. The implementation details, including backbone architectures and source publications, are systematically documented for reference.

As quantitatively demonstrated in Table \ref{tab:lfsr_results}, LFSR models consistently outperform direct SISR approaches across all metrics. Notably, current LFSR methods exhibit suboptimal performance on the EPFL dataset (PSNR $<$ 30 dB), revealing significant room for improvement in this domain.

A detailed comparison reveals that transformer-based architectures achieve state-of-the-art performance by effectively exploiting global attention mechanisms. However, this superior performance comes at the cost of substantially higher computational complexity, potentially limiting their deployment in resource-constrained scenarios. Qualitative results in Fig~\ref{fig:lfsr_reg_vis} demonstrate that deep learning-based approaches, particularly LFSR methods, generate significantly sharper textures compared to traditional interpolation techniques. While conventional SISR methods like RCAN show competitive results, all specialized LFSR implementations deliver visually satisfactory reconstruction quality.

Compared to traditional SISR, LFSR remains a relatively understudied domain, with only a limited body of dedicated research to date. This presents compelling opportunities for future exploration, particularly in two key directions: (1) the development of computationally efficient architectures optimized for edge-device deployment, where resource constraints demand lightweight yet high-performance solutions; and (2) the investigation of advanced angular-spatial feature fusion strategies to better harness the unique 4D structure of light field data, which remains underexploited in current methods. }

\section{Remaining Issues and Future Directions}
\label{future}
\revision{\subsection{\textbf{Unlocking the Power of SR-Specific Priors}}
While many existing methods focus on network architecture—such as model depth and spatial or channel correlations—to improve performance, they often overlook the importance of domain-specific priors. As a result, these models tend to be more generic and less tailored for super-resolution (SR) tasks. There is a clear need to explore SR-specific priors to enhance both quantitative and qualitative performance~\cite{mei2020image,kong2021classsr,Nazeri2019,Cheng2020}, as shown in Fig.~\ref{fig:priors}.

Incorporating edge priors, for example, can significantly improve the restoration of sharp, high-quality images~\cite{Gu2022eccv,Nazeri2019}. Specifically, using edge loss helps the network focus on edges and textures, yielding more visually pleasing results. Additionally, modules designed to extract and integrate edge information into later stages allow the network to learn more useful features~\cite{fang2020soft}.

Similarly, exploiting the inherent sparsity in SR, a sparse mask can be predicted from feature maps, enabling sparse convolution to skip unnecessary positions and reduce model complexity~\cite{wang2021exploring}. Integrating recurrence priors~\cite{shocher2018zero} can further enhance SR by considering near/far distances through patch cropping and applying self-supervised methods to improve performance.}

\subsection{\textbf{Advancements in Architectural Structures for Super-Resolution}}
The field of SR has witnessed the emergence of numerous architectural structures, each contributing to the progress of the field. Initially, basic convolutional layers were employed, and they were later enhanced by incorporating residual connections and attention mechanisms. More recently, transformer-based architectures have emerged as the leading models, exhibiting remarkable performance in SR tasks. Alongside, diffusion models~\cite{sr3_pami,sahak2023denoising,gao2023implicit} have introduced an iterative diffusion process to enhance low-resolution images, gradually refining image details by propagating information between adjacent pixels. This iterative approach results in higher levels of image clarity. \resubmit{Furthermore, the exploration of State Space Models (SSMs), particularly the Mamba architecture, has demonstrated significant potential in SR tasks by efficiently modeling long-range dependencies with linear computational complexity~\cite{di2024qmambabsr,lei2024dvmsr,huang2024irsrmamba,ren2024mambacsr}.} Additionally, Graphic Neural Networks (GNNs) have yielded promising results in SR tasks, while NAS techniques have been employed to discover more efficient and effective structures, despite the time-consuming nature of the process. Although the design of new structures poses challenges, it represents a crucial direction for advancing the SR community, pushing the boundaries of performance and fostering innovation.

\subsection{\textbf{Towards Lightweight Super-Resolution: Balancing Model Complexity and Performance}}
Despite the significant performance advancements achieved in SR, the practical implementation of these models is hindered by their large parameter sizes and high computational costs. It is crucial to develop lightweight SR models that can be deployed on resource-constrained devices. Several methods have been explored to reduce model complexity, including knowledge distillation, re-parameterization, NAS, and quantization~\cite{li2020pams,hong2022daq,hong2022cadyq}. However, these methods often suffer from performance degradation due to information loss during the simplification process. Designing lightweight models that can maintain high performance is still a challenging task that necessitates further research and development.

\subsection{\textbf{Beyond Singular Degradation: Modeling Diverse Processes for Enhanced Super-Resolution}}
Current SR methods commonly focus on modeling a single degradation process, such as Gaussian blur or bicubic blur. However, real-world data presents a broad spectrum of factors that contribute to image degradation, including camera aging, diverse optical settings, and combinations of multiple degradation processes. Some approaches have attempted to tackle this issue by enumerating various degradation processes and integrating them into the training process. Others have redefined the degradation process and employed generative models to estimate the degradation kernel. Nevertheless, comprehensive research in this area is still lacking, and finding suitable solutions for modeling diverse degradation processes remains a challenge that needs to be addressed.

\subsection{\textbf{Bridging the Gap: Advancing Arbitrary Super-Resolution for Diverse Scaling Factors}}
While numerous SR methods primarily concentrate on fixed-scale SR tasks, there is an increasing demand for arbitrary SR, where the scaling factor is not predefined. \revision{Recent works have made strides in exploring arbitrary SR and have achieved promising results~\cite{chen2021learning,lee2022local,hu2019meta}.} These approaches involve modifying the weight kernel to adaptively handle diverse scales or utilizing implicit neural expressions to regress pixel values for arbitrary scaling factors. However, there still exists a performance gap between fixed-scale and arbitrary SR models. Further exploration and research are necessary to enhance the performance of arbitrary SR and bridge this gap effectively.

\subsection{\textbf{Exploring the Synergy Between Low-Level and High-Level Tasks in Super-Resolution}}
Low-level tasks in SR have a dual objective: to restore images and to provide robust inputs for high-level tasks. Regrettably, there is a paucity of research that explores the relationship and interaction between low-level and high-level tasks. Establishing connections between these two tasks can enable seamless information exchange across different levels and potentially foster the fusion of different models. This direction holds promise and bears significant value for future implementations in SR.

\subsection{\textbf{Addressing Data Challenges in Super-Resolution: Towards Dedicated Large-Scale Low-Level Datasets}}
In contrast to high-level tasks that often benefit from access to large-scale datasets, the field of SR lacks such abundant resources, which presents a challenge to its development. Although some approaches have suggested leveraging datasets like ImageNet for pre-training models, the image quality in such datasets may not be adequate for low-level tasks, where image quality plays a crucial role. Constructing dedicated large-scale low-level datasets would serve as a major boost to SR performance.

\section{Conclusions}
\label{conclusion}

\revision{We present a comprehensive and in-depth review of Super-Resolution (SR) methods, covering a wide range of areas including single image super-resolution (SISR), video super-resolution (VSR), stereo super-resolution (SSR), and light field super-resolution (LFSR). Our survey encompasses over 150 SISR methods, nearly 70 VSR approaches, and approximately 30 techniques for SSR and LFSR. We categorize SR methods into two primary groups: regression-based and generative models, and further subdivide them according to their backbone architectures. For regression-based methods, we discuss convolutional neural network (CNN)-based and transformer-based approaches, while for generative models, we explore those using generative adversarial networks (GANs) and diffusion models. Each category is further refined based on specific architectural innovations and techniques. Our review also delves into critical topics such as datasets, evaluation protocols, empirical performance, and implementation challenges. Moreover, we highlight key open issues that require further exploration, aiming to inspire continued research and innovation in the SR domain. We hope this review will serve as a valuable resource for researchers, providing insights and guidance for future advancements in super-resolution technologies.}

\ifCLASSOPTIONcaptionsoff
  \newpage
\fi

\bibliographystyle{IEEEtran}
\bibliography{IEEEabrv,./bib/Survey.bib}

\end{document}